\documentclass[11pt]{article}

\usepackage[final]{acl}

\usepackage{graphicx}
\usepackage{caption} 
\usepackage{kotex}   
\usepackage{amsmath} 
\usepackage{listings}

\usepackage{times}
\usepackage{kotex}
\usepackage{latexsym}
\usepackage{booktabs}
\usepackage{multirow}
\usepackage[T1]{fontenc}
\usepackage{amsmath}
\usepackage[htt]{hyphenat}
\usepackage[utf8]{inputenc}
\usepackage[table]{xcolor}

\newcommand{\tightcell}[2]{%
  {\begingroup\setlength{\fboxsep}{1pt}\colorbox{#1}{#2}\endgroup}%
}
\usepackage{microtype}
\usepackage{hyperref}

\usepackage{inconsolata}

\usepackage{graphicx}
\usepackage{pifont} 
\usepackage{xcolor} 
\usepackage{tcolorbox}
\usepackage{amssymb}
\usepackage{booktabs}

\usepackage{caption}
\definecolor{expertgreen}{RGB}{0, 150, 0}
\definecolor{expertred}{RGB}{200, 0, 0}
\newcommand{\pmstd}[2]{#1\textsubscript{\scriptsize#2}}
\newcommand{\cmark}{\textcolor{expertgreen}{\ding{51}}} 
\newcommand{\xmark}{\textcolor{expertred}{\ding{55}}}  
\newcommand{\pmark}{$\triangle$}

\usepackage{tcolorbox}
\tcbuselibrary{listings,breakable}
\definecolor{promptframe}{RGB}{70,90,120}
\definecolor{promptback}{RGB}{245,247,250}
\newtcblisting{promptbox}[1]{%
  listing only,
  breakable,
  colback=promptback,
  colframe=promptframe,
  coltitle=white,
  fonttitle=\bfseries\small,
  title=#1,
  boxrule=0.6pt,
  arc=2pt,
  left=4pt, right=4pt, top=3pt, bottom=3pt,
  listing options={
    basicstyle=\ttfamily\footnotesize,
    breaklines=true,
    breakatwhitespace=true,
    columns=fullflexible,
    keepspaces=true,
    showstringspaces=false
  }
}

\title{
KMMMU: Evaluation of \underline{M}assive \underline{M}ulti-discipline \underline{M}ultimodal \underline{U}nderstanding in \underline{K}orean Language and Context
}

\author{
\textbf{Nahyun Lee\textsuperscript{1,4,5}\thanks{Equal contribution.}} \quad
\textbf{Guijin Son\textsuperscript{2,4,5}\footnotemark[1]} \quad
\textbf{Hyunwoo Ko\textsuperscript{4,5}} \quad
\textbf{Chanyoung Kim\textsuperscript{3,5}} \\
\textbf{Junyoung An\textsuperscript{2}} \quad
\textbf{Kyubeen Han\textsuperscript{5}} \quad
\textbf{Youngjae Yu\textsuperscript{2}} \quad
\textbf{Il-Youp Kwak\textsuperscript{1}\thanks{Corresponding author: \href{mailto:ikwak2@cau.ac.kr}{ikwak2@cau.ac.kr}}} \\[0.5em]
{\small
\textsuperscript{1}Chung-Ang University
\quad
\textsuperscript{2}Seoul National University
\quad
\textsuperscript{3}SK AX
\quad
\textsuperscript{4}OnelineAI
\quad
\textsuperscript{5}HAE-RAE
}}

\begin{document}
\maketitle

% It also includes a 300-item Korean-specific diagnostic subset and a 627-item hard subset, enabling analysis of Korean language understanding, structured visual reasoning, domain knowledge, and localized conventions.

\begin{abstract}
Multimodal evaluation is strengthened when benchmarks reflect the languages, institutions, and disciplinary conventions that shape expert visual reasoning.
We introduce \textbf{KMMMU}, a native Korean benchmark built from Korean-native examination materials.
KMMMU contains 3,466 questions spanning nine disciplines and nine visual modality categories. For targeted evaluation on challenging questions and Korean-specific knowledge, we additionally provide KMMMU-Adv and KMMMU-Local, respectively.
Experiments show that the strongest evaluated open-weight model reaches 42.05\% on the KMMMU, while the best evaluated proprietary model achieves 52.42\% on the KMMMU-Adv.
Performance varies substantially across disciplines and answer formats, with Korean-specific questions showing gaps of up to 13.43\%.
Error analysis suggests that inspected failures often involve convention-to-label mapping, few-shot symbolic induction, localized knowledge recall, and domain-specific standards, not reasoning depth alone.
KMMMU provides a testbed for native Korean expert-level multimodal understanding beyond English-centric or translation-based evaluation.
\footnote{Dataset is available at \url{https://huggingface.co/datasets/HAERAE-HUB/KMMMU}}

\end{abstract}

\section{Introduction}
\begin{figure}[t]
    \centering
    \includegraphics[width=1.05\linewidth]{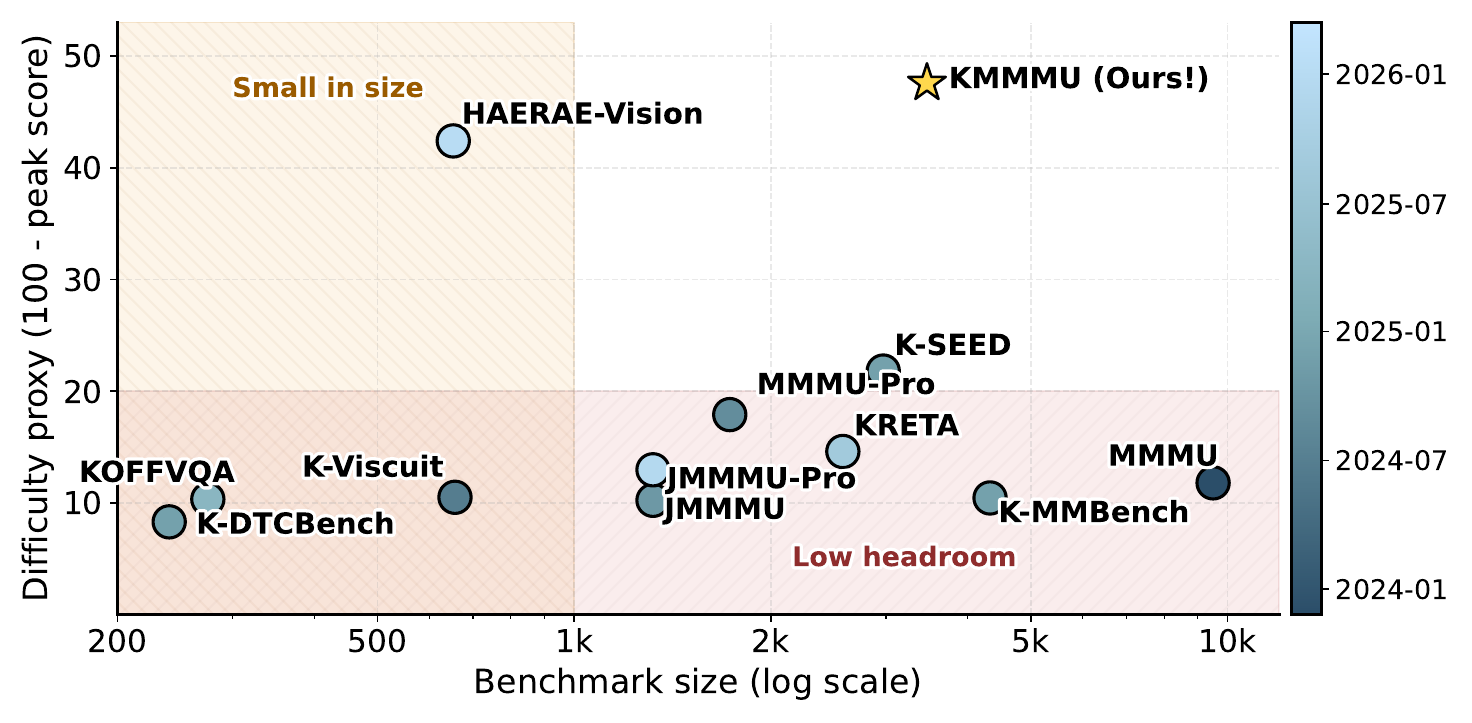}
    \caption{\footnotesize
    \textbf{Comparison of English (MMMU, MMMU-Pro), Japanese (JMMMU, JMMMU-Pro), and Korean (others) multimodal benchmarks.} Each point is positioned by benchmark size (x-axis, log scale) and difficulty proxy (100 $ - $ peak public score), with lighter colors indicating more recent releases. Shaded regions mark two common limitations: small size (left) and high peak model performance (bottom). 
    }
    \label{fig:benchmark_landscape}
% \vspace{-3mm}
\end{figure}

Multimodal Large Language Models (MLLMs) have shown strong performance on a range of vision--language tasks, including visual recognition, document understanding, and multimodal question answering~\citep{alayrac2022flamingo, li2023blip2, liu2023llava, team2023gemini, bai2025qwen3vltechnicalreport}. 
As MLLMs are deployed across a wider range of languages and domains, evaluation must account for tasks where linguistic context, visual structure, domain knowledge, official standards, and specialized terminology jointly shape model performance.
However, most widely used multimodal benchmarks remain English-centric or are derived from translated datasets~\citep{li2023evaluating, yue2024mmmu}. Korean multimodal benchmarks have made important progress, but many either target narrower settings or have reported peak scores high enough to leave limited separation among current strong models (Figure~\ref{fig:benchmark_landscape}).
% existing Korean multimodal benchmarks are either small or already saturated (Figure~\ref{fig:benchmark_landscape}), motivating the need for a more reliable evaluation tool.

To bridge this gap, we introduce \textbf{KMMMU}, a collection of \textbf{3,466} questions drawn from Korean assessment sources, spanning nine disciplines, nine visual modality categories, and both \textit{multiple-choice} and \textit{open-form} question formats. To support practitioners with lightweight evaluation, we further provide two subsets: \textbf{KMMMU-Local}, which targets Korean-specific localized knowledge, and \textbf{KMMMU-Adv}, a challenging subset consisting of questions selected models fail to answer.

\begin{figure*}[t]
    \centering
    \includegraphics[width=0.9\linewidth]{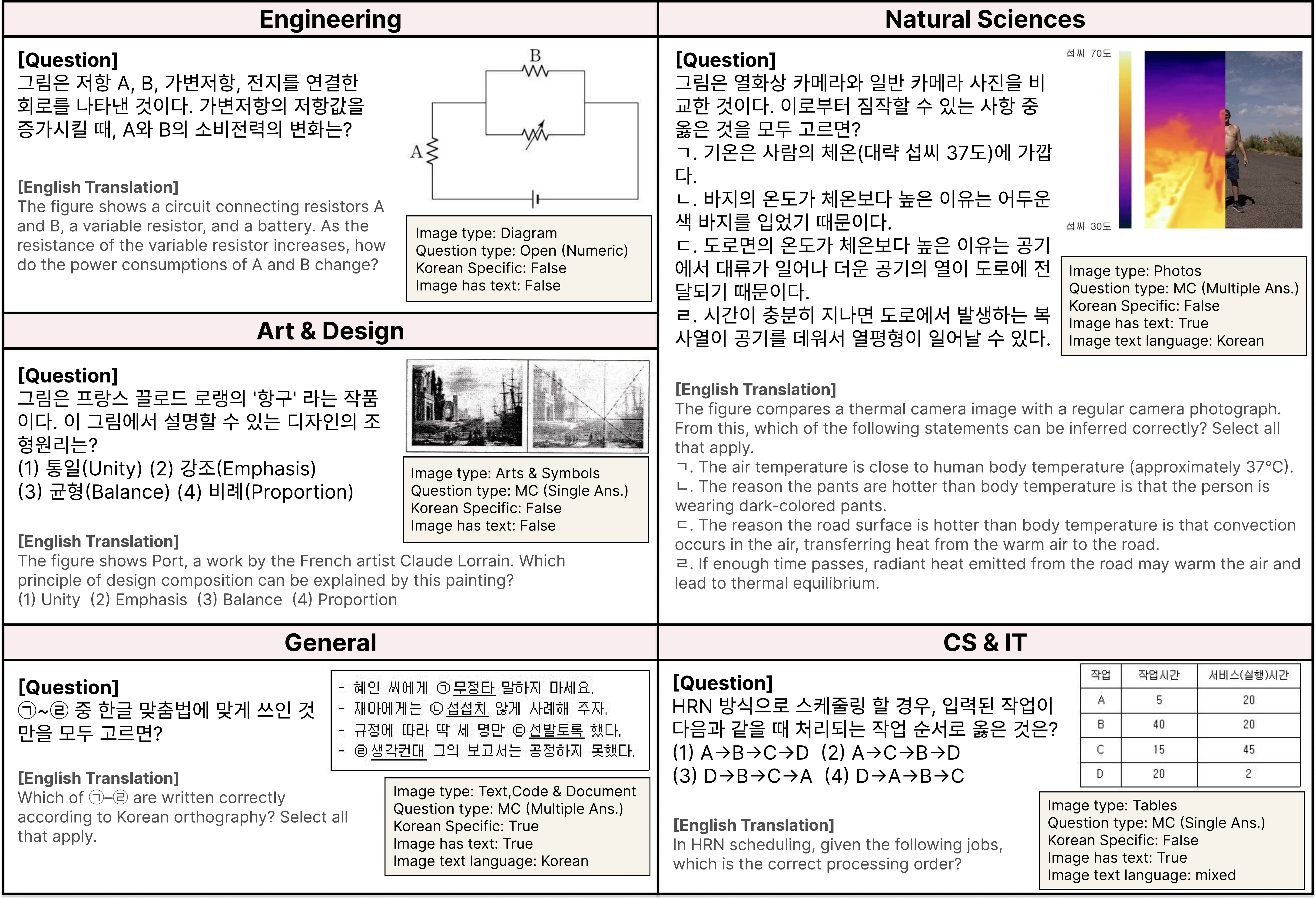}
    \caption{\footnotesize
    \textbf{Examples of KMMMU questions.}
    Examples include the original questions, associated images, English translations, and metadata such as visual modality, question format, and Korean-specific labels.
    }
    \label{fig:question_example}
\vspace{-2mm}
\end{figure*}

Experiments on KMMMU reveal several consistent findings.
We evaluate 26 multimodal models spanning open-weight and proprietary systems, and among them the strongest evaluated open-weight model reaches 42.05\% on the full set, while the best proprietary model reaches 52.42\% on KMMMU-Adv.
Accuracy on Korean-specific questions drops by up to 13.43\% relative to non-Korean-specific items, indicating that they remain challenging even for strong multilingual models.
These results suggest that high overall multimodal performance does not necessarily translate into reliable performance on native Korean expert-level assessment tasks.

\section{Related Works}

\begin{table}[t]
\centering
\small
\setlength{\tabcolsep}{3pt}
\resizebox{\columnwidth}{!}{%
\begin{tabular}{lcccccc}
\toprule
\textbf{Benchmark} & \textbf{Size} & \textbf{Native} & \textbf{Expert} &
\textbf{Multi-disc.} & \textbf{Diverse vis.} & \textbf{Local subset} \\
\midrule
K-MMBench     & 4.3k & \xmark & \xmark & \xmark & \xmark & \xmark \\
K-SEED        & 3.0k & \xmark & \xmark & \xmark & \xmark & \xmark \\
K-MMStar      & 1.5k & \xmark & \xmark & \xmark & \xmark & \xmark \\
K-DTCBench    & 0.2k & \cmark & \xmark & \xmark & \pmark & \xmark \\
K-Viscuit     & 0.7k & \cmark & \xmark & \xmark & \xmark & \pmark \\
KoFFVQA       & 0.3k & \cmark & \xmark & \xmark & \xmark & \pmark \\
KoNET         & 2.4k & \cmark & \xmark & \cmark & \pmark & \xmark \\
KRETA         & 2.6k & \cmark & \xmark & \cmark & \cmark & \xmark \\
KorMedMCQA-V  & 1.5k & \cmark & \cmark & \xmark & \pmark & \xmark \\
HAERAE-Vision & 0.7k & \cmark & \xmark & \pmark & \xmark & \cmark \\
\midrule
\textbf{KMMMU (ours)} & 3.5k & \cmark & \cmark & \cmark & \cmark & \cmark \\
\bottomrule
\end{tabular}}
\caption{\textbf{Comparison with Korean multimodal benchmarks.} Native denotes items sourced
without translation, and Local subset denotes a dedicated subset for Korean-localized
knowledge (KMMMU-Local in our case). \pmark\ denotes partial coverage.}
\label{tab:kor_bench_compare}
\end{table}

Several Korean multimodal benchmarks have been proposed, targeting text-rich VQA~\citep{hwang2025kreta}, educational exams~\citep{park2025konet}, medical reasoning~\citep{choi2026kormedmcqa}, free-form VQA~\citep{kim2025koffvqa}, cultural understanding~\citep{park2025evaluating}, under-specified queries~\citep{choi2026users}, translated variants, and document reasoning~\citep{ju2024varcovisionexpandingfrontierskorean}.
However, as shown in Figure~\ref{fig:benchmark_landscape}, these efforts often compromise on breadth, realism, and evaluation headroom (the performance gap remaining before saturation). Consequently, most existing Korean benchmarks already lie in the low-headroom region.
Table~\ref{tab:kor_bench_compare} further shows that translated benchmarks inherit English source tasks, while native benchmarks each miss at least one of expert-level content, disciplinary breadth, visual diversity, or a localized-knowledge subset.

Exam-derived MMMU-style benchmarks have become a standard tool for measuring broad knowledge and reasoning in MLLMs, offering scale, disciplinary breadth, and an interpretable link to human expertise~\citep{hendrycks2020measuring, yue2024mmmu, zhang2023m3exam, zhong2024agieval}. This paradigm has been extended to other languages such as Japanese (JMMMU) and Chinese (CMMMU)~\citep{onohara2025jmmmu, zhang2024cmmmu}, but Korean still lacks a comparable resource.
KMMLU is constructed from original Korean exams rather than translations and captures linguistic and cultural factors that translated benchmarks often miss~\citep{son2025kmmlu}, and KMMLU-Pro~\citep{hong2025kmmlu} reports that the gap between translated MMMLU~\citep{openai2024mmmlu} and locally authored Korean professional exams is small in medicine but substantially larger in law-related domains, where country-specific knowledge is indispensable.
These gaps have so far been examined only in text, however; how Korean characters embedded in images or culturally grounded visual content affect multimodal model performance remains largely unexplored.
KMMMU follows the MMMU paradigm in the same sense as JMMMU and CMMMU. The name refers to this evaluation paradigm, namely broad expert-level multimodal understanding measured with examination questions in multiple-choice and open-form formats. KMMMU does not reuse or adapt any MMMU item, and every question comes from a pre-existing Korean examination.

\begin{table}[htbp]
\centering
\setlength{\abovecaptionskip}{4pt}
\scriptsize
\setlength{\tabcolsep}{2pt}
\renewcommand{\arraystretch}{1.15}
\begin{tabular}{p{0.35\linewidth} p{0.10\linewidth} p{0.12\linewidth} p{0.38\linewidth}}
\toprule
\textbf{Benchmark} & \textbf{Year} & \textbf{Size} & \textbf{Filtering Method} \\
\midrule

\textbf{HellaSwag}
\newline \textit{Commonsense NLI}
& 2019
& 70K
& Adversarially selected hard generated distractors \\

\textbf{MMLU-Pro}
\newline \textit{Text-only exams}
& 2024
& 12K
& Remove items solved by $>4$ of 8 LLMs \\

\textbf{SimpleQA}
\newline \textit{Short-form factuality}
& 2024
& 4.3K
& Collect questions adversarially against GPT-4 responses \\

\textbf{MMMU-Pro}
\newline \textit{Multimodal exams}
& 2025
& 3.5K
& Remove items solved by $\geq 3$ of 4 text-only LLMs \\

\textbf{KMMLU-Pro}
\newline \textit{Text-only Korean exams}
& 2025
& 2.8K
& Filter items solved by $\geq 4$ of 7 small LLMs \\

\textbf{HLE}
\newline \textit{Expert-level knowledge}
& 2025
& 2.5K
& Frontier-LLM difficulty check before expert review \\

\textcolor{red}{\textbf{KMMMU}}  (ours)
\newline \textit{Multimodal Korean exams}
& 2026
& 3.4K
& Remove items solved by four MLLMs \\

\bottomrule
\end{tabular}
\caption{
\footnotesize
\textbf{Comparison of benchmark construction strategies that preserve evaluation difficulty through adversarial selection, model-aware filtering, or frontier-model difficulty checks.}
}
\label{tab:model_aware_filtering_main}
\vspace{-2mm}
\end{table}

\section{The KMMMU Benchmark}

\subsection{Data Collection and Annotation}
KMMMU is constructed from Korean-native official examinations and competitions.
These sources include the civil service recruitment (PSAT), National Technical Qualifications (NTQ), National Competency Standards exam (NCS), and academic Olympiads (see Appendix~\ref{sec:appendix_sources} for details).
We initially collect approximately 68k raw instances.

We process the collected exam materials into structured multimodal instances using automated extraction followed by source-grounded manual verification.
Technical qualification data are collected through web crawling, while other sources are digitized using the \textit{MinerU-2.5} OCR system~\citep{niu2025mineru25decoupledvisionlanguagemodel}.
Five native Korean annotators review the items, and to support verification, we built a custom interface to compare parsed instances with the original source materials.
The verification covered question text, answer choices, LaTeX formulas, image crops, image references, and answer keys (see Appendix~\ref{sec:appendix_annotation} for detailed procedure).
Because many source materials are digitized from PDF documents, the benchmark may be less exposed to direct web-scale crawl contamination than datasets collected from public web QA repositories. 
We therefore expect this process to reduce contamination risk.

\begin{figure}[t]
    \centering
    \includegraphics[width=0.6\linewidth]{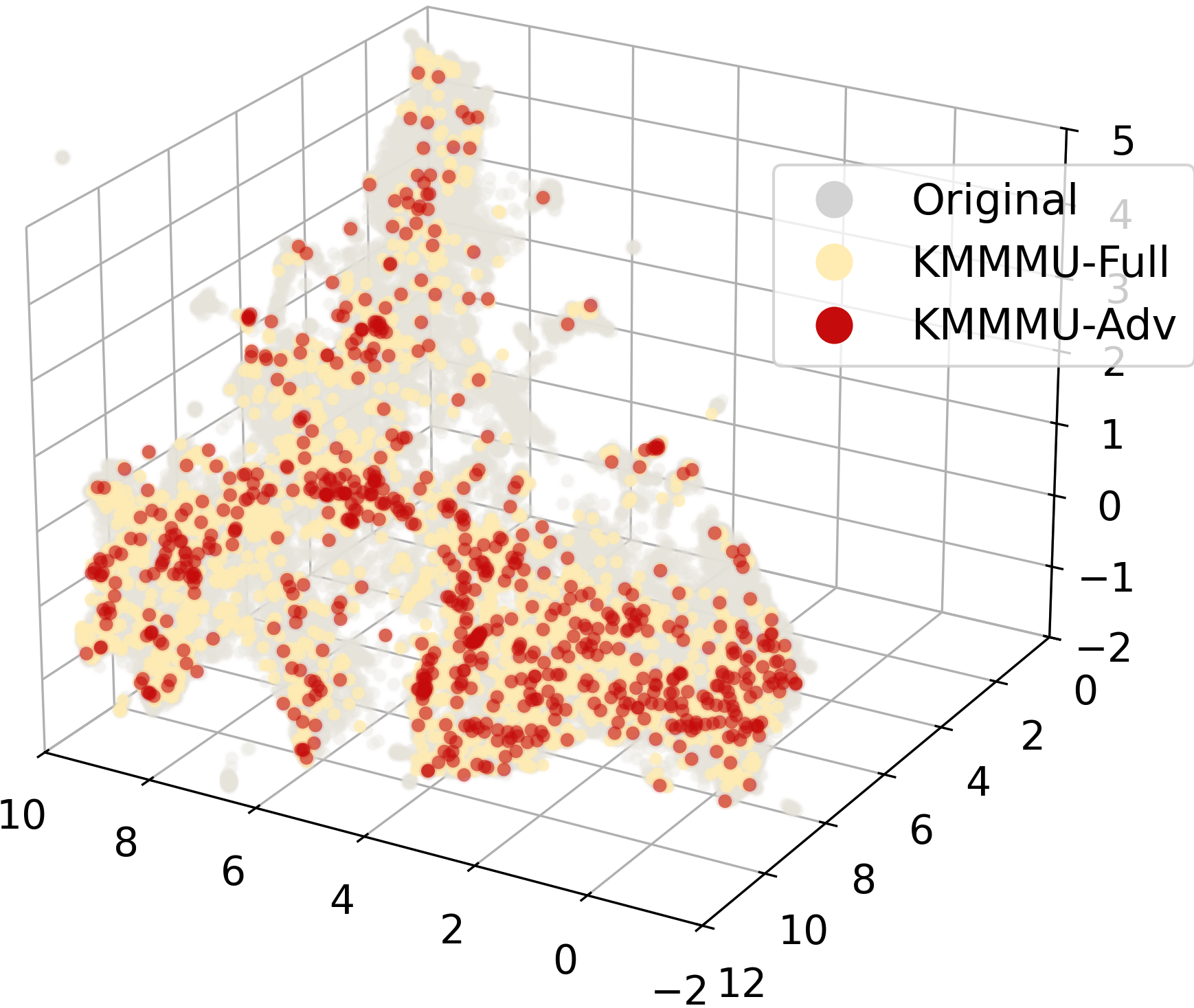}
    \caption{\footnotesize \textbf{Embedding-based distributional comparison after selection.}
    Question text embeddings from the original 68k corpus, the \textit{KMMMU}, and the \textit{KMMMU-Adv} are projected using PCA followed by 3D UMAP.}
    % Both retained subsets largely preserve the global structure of the original distribution.}
    \label{fig:embedding_umap}
% \vspace{-3mm}
\end{figure}

\subsection{KMMMU Dataset Construction}

Starting from approximately 68k manually verified questions, we apply multi-stage filtering to retain challenging instances. 
We evaluate the candidate pool with four MLLMs in sequence: \texttt{Phi-3.5-Vision-Instruct}~\citep{abdin2024phi}, \texttt{InternVL-3.5-38B}~\citep{wang2025internvl3_5}, \texttt{Gemini-2.5-Flash-Lite}, and \texttt{Gemini-2.5-Flash}~\citep{comanici2025gemini}. 
At each stage, questions answered correctly by the corresponding model are removed from the candidate pool (see Appendix~\ref{app:adversarial_filtering} for retention statistics).

This filtering step follows a common practice in recent benchmark construction~\citep{zellers2019hellaswag,wang2024mmlu,yue2025mmmu,hong2025kmmlu,wei2024measuring,phan2025humanity}.
As summarized in Table~\ref{tab:model_aware_filtering_main}, model-based filtering is critical for producing challenging, high-quality benchmarks that preserve evaluation headroom against frontier models.
At the same time, aggressive filtering can distort the original distribution, so it is equally important to ensure that the filtered set retains broad coverage of the source corpus.

Therefore, we additionally assess whether filtering substantially changes the original distribution.
Using \texttt{multilingual-e5-large} embeddings projected with PCA ($n=50$) and 3D UMAP, we find that both \textit{KMMMU} and \textit{KMMMU-Adv} largely preserve the global structure of the original 68k-sample pool, as shown in Figure~\ref{fig:embedding_umap}.
Dimension-wise KL divergence also remains low for \textit{KMMMU} ($D_{KL}=0.11$--$0.15$), suggesting that filtering increases difficulty without strongly distorting the original corpus structure.
Appendix~\ref{sec:appendix_distribution} provides additional density comparisons and KL analyses.
% while \textit{KMMMU-Adv} deviates mainly along one dimension ($D_{KL}=0.3747$), => 이건 appendix에서 언급하면 될 듯

In short, we build KMMMU, a collection of \textbf{3,466} questions sourced entirely from native Korean materials without translation, the first MMMU-style benchmark of its kind for the Korean cultural sphere.
Figure~\ref{fig:question_example} shows representative instances across disciplines.

% while item correctness and data fidelity are ensured through source-grounded answer keys and human verification. 

% Model-based filtering may indirectly reduce the chance that easily exposed or memorized items remain in the final benchmark.
% However, model correctness is an ambiguous signal, since it may reflect genuine capability as well as possible memorization.
% Given the difficulty of direct training-set verification under limited data transparency~\citep{bommasani2023foundation,jacovi2023stop}, we report contamination analysis separately as a post-hoc black-box behavioral check.
% ~\citep{golchin2023time}.

\begin{figure}[t]
    \centering
    \includegraphics[width=0.95\linewidth]{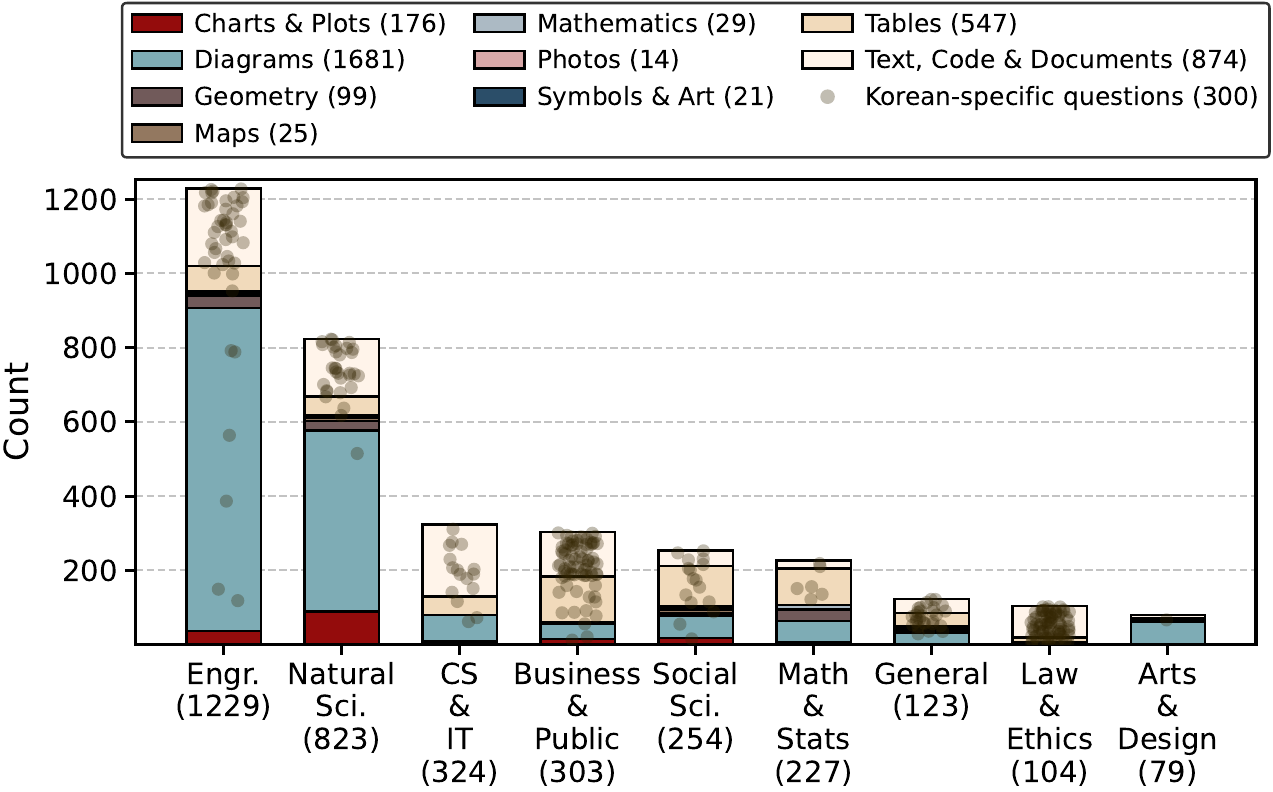}
    \caption{\footnotesize \textbf{Discipline-wise visual modality composition of KMMMU.} Stacked bars show the number of questions for each visual modality in each discipline, with total counts shown beneath the labels. Scatter points indicate \textit{Korean-specific} items overlaid on the corresponding discipline--modality segments, and jittered randomly.}
    \label{fig:kmmmu_dataset_taxonomy}
\vspace{-3mm}
\end{figure}

\subsection{Taxonomy and Dataset Composition}
\label{subsec:taxonomy}

KMMMU is designed to evaluate expert-level multimodal understanding across diverse domains.
Each instance is annotated along four axes: discipline, visual modality, question format, and Korean-specificity (detailed in Section~\ref{subsec:adv_local}).
Initial labels are assigned by \texttt{Gemini-2.5-Flash}~\citep{comanici2025gemini} and validated through human audit and consolidation.

Figure~\ref{fig:kmmmu_dataset_taxonomy} presents the discipline-wise distribution of KMMMU.
Overall, \textit{Engineering} forms the largest portion of the dataset, followed by \textit{Natural Science} and \textit{CS\&IT}.
Meanwhile, \textit{Diagrams} and \textit{Text, Code \&Documents} constitute the most frequent visual modalities (see Appendix~\ref{app:visual_modality_taxonomy} for detailed examples).

\subsection{Construction of KMMMU-Adv and KMMMU-Local}
\label{subsec:adv_local}
To further analyze model limitations, we construct \textit{KMMMU-Adv}, a challenging subset of \textbf{627} questions answered incorrectly by all three baseline models: \texttt{Gemma-3-27B}~\citep{team2025gemma}, \texttt{Qwen3-VL-235B-Thinking}~\citep{bai2025qwen3vltechnicalreport}, and \texttt{GPT-5-nano}~\citep{openai_introducing_gpt5_2025}.
This subset corresponds to $18\%$ of the full KMMMU dataset and provides a baseline-relative stress test for instances that remain difficult.
Additional composition details are provided in Appendix Figure~\ref{fig:discipline_hard_visual_cross}.

We also provide \textit{KMMMU-Local}, a subset of \textbf{300} Korean-specific items designed to assess whether strong multimodal performance transfers to localized institutional, cultural, and terminological knowledge.
The Korean-specificity label identifies items whose solutions depend on Korean-localized knowledge, including Korean laws, standards, institutions, geography, history, culture, or official terminology beyond general world knowledge.
Following the labeling pipeline described in Section~\ref{subsec:taxonomy}, we manually audit 300 random instances and verify all items labeled as \textit{KMMMU-Local}.
As shown by the overlaid scatter points in Figure~\ref{fig:kmmmu_dataset_taxonomy}, \textit{KMMMU-Local} items are concentrated in institutionally grounded disciplines such as \textit{Business \& Public} and \textit{Law \& Ethics}, where solving questions often requires knowledge of Korean administrative categories, regulations, or official standards. By contrast, many STEM and technical items, even when drawn from Korean exams, can still be solved using general domain knowledge.

\begin{table*}[h]
\centering
\setlength{\abovecaptionskip}{4pt}
\small
\resizebox{\textwidth}{!}{%
\begin{tabular}{lcccccccccc}
\toprule
\multicolumn{1}{l}{\textbf{Model}} &
\multicolumn{1}{c}{\shortstack{\textbf{Arts} \\ \textbf{\& Design}}} &
\multicolumn{1}{c}{\shortstack{\textbf{Business} \\ \textbf{\& Public}}} &
\multicolumn{1}{c}{\shortstack{\textbf{CS} \\ \textbf{\& IT}}} &
\multicolumn{1}{c}{\textbf{Engineering}} &
\multicolumn{1}{c}{\shortstack{\textbf{General}}} &
\multicolumn{1}{c}{\shortstack{\textbf{Law} \\ \textbf{\& Ethics}}} &
\multicolumn{1}{c}{\shortstack{\textbf{Math} \\ \textbf{\& Stats}}} &
\multicolumn{1}{c}{\shortstack{\textbf{Natural} \\ \textbf{Sciences}}} &
\multicolumn{1}{c}{\shortstack{\textbf{Social} \\ \textbf{Sciences}}} &
\multicolumn{1}{c}{\shortstack{\textbf{Overall} \\ \textbf{Acc.}}} \\
\midrule
\multicolumn{11}{l}{\textbf{Open-Weight Multilingual Non-Reasoning Models}} \\
\midrule
Gemma-3-4B-IT & \textbf{29.96$_{1.58}$} & 26.51$_{1.58}$ & 18.83$_{2.40}$ & 23.19$_{0.66}$ & 19.24$_{1.67}$ & 22.76$_{2.27}$ & 17.03$_{1.81}$ & 15.25$_{1.13}$ & 19.29$_{2.32}$ & 20.49$_{0.41}$ \\

Gemma-3-12B-IT & 28.27$_{1.58}$ & 25.85$_{2.45}$ & 20.99$_{1.57}$ & 22.95$_{0.53}$ & 14.09$_{2.03}$ & 31.09$_{1.98}$ & 16.74$_{2.94}$ & 18.49$_{1.29}$ & 22.83$_{2.25}$ & 21.59$_{0.55}$ \\

Gemma-3-27B-IT & 23.63$_{6.72}$ & 31.02$_{2.85}$ & 25.41$_{1.91}$ & 22.48$_{0.33}$ & 17.62$_{0.38}$ & 30.77$_{3.93}$ & 23.35$_{0.95}$ & 21.74$_{1.35}$ & 26.64$_{3.04}$ & 23.78$_{0.88}$ \\

Qwen3-VL-2B-IT & 24.05$_{1.03}$ & 22.88$_{1.33}$ & 19.55$_{0.52}$ & 13.75$_{0.65}$ & 8.67$_{2.03}$ & 23.08$_{2.08}$ & 8.52$_{1.36}$ & 9.61$_{0.60}$ & 15.75$_{0.85}$ & 14.24$_{0.34}$ \\

Qwen3-VL-4B-IT & 25.32$_{1.03}$ & 34.43$_{1.38}$ & 26.75$_{1.62}$ & 16.73$_{0.79}$ & 13.28$_{1.38}$ & 24.36$_{2.52}$ & 13.22$_{0.95}$ & 17.68$_{0.70}$ & 21.78$_{0.67}$ & 19.88$_{0.42}$ \\

Qwen3-VL-8B-IT & 25.74$_{1.58}$ & 37.95$_{1.23}$ & 29.63$_{1.97}$ & 20.86$_{1.64}$ & 13.82$_{1.76}$ & 33.01$_{2.40}$ & 22.61$_{1.16}$ & 22.63$_{0.62}$ & 29.53$_{0.96}$ & 24.56$_{0.77}$ \\

Qwen3-VL-30B-A3B-IT & 28.27$_{8.35}$ & 46.09$_{2.04}$ & 31.58$_{1.43}$ & 22.38$_{1.04}$ & 13.55$_{2.51}$ & 25.96$_{4.08}$ & 30.54$_{3.00}$ & 32.64$_{0.15}$ & 39.37$_{0.64}$ & 29.45$_{0.98}$ \\

Qwen3-VL-32B-IT & 27.43$_{1.19}$ & 52.15$_{1.64}$ & 36.93$_{0.29}$ & 28.64$_{0.75}$ & 22.76$_{1.76}$ & 32.69$_{2.08}$ & 43.91$_{1.45}$ & 36.29$_{0.66}$ & 48.95$_{1.86}$ & 35.65$_{0.48}$ \\

Qwen3-VL-235B-A22B-IT & 25.32$_{2.73}$ & \textbf{57.87$_{0.82}$} & 37.45$_{1.79}$ & 31.38$_{0.56}$ & \textbf{24.66$_{2.68}$} & 35.58$_{2.36}$ & \textbf{53.01$_{1.26}$} & \textbf{41.40$_{0.64}$} & \textbf{53.81$_{1.45}$} & \textbf{39.44$_{0.08}$} \\

Llama-4-Scout-17B-16E-IT & 26.58$_{1.03}$ & 51.05$_{1.09}$ & 36.01$_{1.54}$ & 30.81$_{1.06}$ & 23.85$_{2.33}$ & 33.65$_{1.36}$ & 34.07$_{0.91}$ & 30.86$_{0.06}$ & 39.63$_{1.65}$ & 33.67$_{0.33}$ \\

Llama-4-Maverick-17B-128E-IT & 27.43$_{2.60}$ & 55.89$_{2.06}$ & \textbf{39.81$_{1.82}$} & \textbf{34.09$_{1.04}$} & 20.87$_{2.03}$ & \textbf{38.46$_{0.79}$} & 43.61$_{0.95}$ & 39.29$_{0.46}$ & 48.56$_{0.81}$ & 38.95$_{0.47}$ \\
\midrule
\multicolumn{11}{l}{\textbf{Open-Weight Korean Non-Reasoning Models}} \\

\midrule

HyperCLOVAX-SEED-Vision-3B & 22.36$_{2.98}$ & 26.73$_{1.23}$ & 24.28$_{0.15}$ & 17.22$_{0.21}$ & 17.07$_{1.15}$ & 29.81$_{0.79}$ & 13.66$_{0.36}$ & 13.63$_{0.34}$ & 17.59$_{0.49}$ & 18.14$_{0.15}$ \\

VARCO-VISION-2.0-1.7B & 23.63$_{6.64}$ & 26.40$_{2.16}$ & 29.12$_{2.92}$ & 25.90$_{0.57}$ & 19.24$_{1.01}$ & 24.68$_{2.40}$ & 19.82$_{2.36}$ & 19.38$_{0.66}$ & 20.73$_{2.92}$ & 23.59$_{0.91}$ \\

VARCO-VISION-2.0-14B & \textbf{28.69$_{3.63}$} & \textbf{34.76$_{1.75}$} & \textbf{33.02$_{0.25}$} & \textbf{27.96$_{0.48}$} & \textbf{20.05$_{2.03}$} & \textbf{33.01$_{1.98}$} & \textbf{24.08$_{1.85}$} & \textbf{23.52$_{0.30}$} & \textbf{27.30$_{1.65}$} & \textbf{27.55$_{0.37}$} \\
\midrule
\multicolumn{11}{l}{\textbf{Open-weight Multilingual Reasoning Models}} \\
\midrule
Qwen3-VL-30B-A3B-Thinking & 24.68$_{0.63}$ & 52.31$_{0.83}$ & 35.96$_{2.62}$ & 27.54$_{0.77}$ & 21.54$_{1.22}$ & \textbf{34.13$_{2.40}$} & 46.92$_{1.10}$ & 38.20$_{0.12}$ & 44.88$_{0.39}$ & 35.47$_{0.68}$ \\
Qwen3-VL-32B-Thinking & \textbf{25.74$_{3.16}$} & 58.75$_{1.77}$ & \textbf{41.77$_{1.24}$} & 28.10$_{0.44}$ & 20.05$_{3.07}$ & 33.97$_{2.52}$ & 49.93$_{0.91}$ & 39.86$_{0.45}$ & 51.18$_{1.47}$ & 37.80$_{0.49}$ \\
Qwen3-VL-235B-A22B-Thinking & 22.36$_{4.30}$ & \textbf{62.38$_{0.47}$} & 40.33$_{1.27}$ & \textbf{32.79$_{0.98}$} & \textbf{30.08$_{2.89}$} & 32.05$_{2.76}$ & \textbf{56.09$_{0.55}$} & \textbf{45.30$_{0.80}$} & \textbf{57.87$_{0.85}$} & \textbf{42.05$_{0.48}$} \\
% KOREAson-G3-12B-1009 & 21.28$_{0.76}$ & 39.43$_{0.61}$ & 27.63$_{0.99}$ & 16.80$_{1.12}$ & 15.81$_{2.06}$ & 23.80$_{4.33}$ & 24.01$_{2.04}$ & 17.55$_{0.41}$ & 29.26$_{1.64}$ & 21.62$_{0.66}$ \\

\bottomrule
\end{tabular}
}
\caption{\footnotesize
    \textbf{Accuracy (\%) on the KMMMU by disciplines.}
    Overall accuracy is averaged across disciplines.
    Mean accuracy is reported in percentage, with standard deviation shown as a subscript.
    Best in each model group is shown in \textbf{bold}.
}
\label{tab:main-result-overall}
% \vspace{-2mm}
\end{table*}

\section{Experimental setup}
\subsection{Evaluated Models}
We evaluate a diverse set of multimodal models covering both open-weight and proprietary systems.
The models are organized according to whether they employ explicit reasoning during inference.

The \textbf{Open-weight Non-Reasoning} group includes \texttt{Gemma-3}~\citep{team2025gemma} (4B, 12B, 27B), \texttt{Qwen3-VL}~\citep{bai2025qwen3vltechnicalreport} (2B, 4B, 8B, 32B, 30B-A3B, 235B-A22B), and \texttt{Llama-4}~\citep{meta_llama4_scout_17b_16e_hf, meta_llama4_maverick_17b_128e_instruct_hf} (Scout and Maverick), along with the Korean models \texttt{VARCO-VISION-2.0}~\citep{cha2025varco} (1.7B, 14B) and \texttt{HyperCLOVAX-SEED-Vision-3B}~\citep{naver_hyperclovax_seed_vision_instruct_3b_hf}.

The \textbf{Open-Weight Reasoning} group includes \texttt{Qwen3-VL-Thinking}~\citep{bai2025qwen3vltechnicalreport} (30B-A3B, 32B, 235B-A22B).

Because of the cost constraints, we report the \textbf{Proprietary model} group \emph{only on the KMMMU-Adv}: \texttt{GPT-5} and \texttt{GPT-5-mini}~\citep{openai_introducing_gpt5_2025}, \texttt{Claude-Opus-4.5} and \texttt{Claude-Sonnet-4.5}~\citep{anthropic_introducing_claude_opus_45_2025}, \texttt{Grok-4} and \texttt{Grok-4.1-Fast}~\citep{xai_grok_models_docs}, \texttt{Gemini-3-Pro}~\citep{google_vertexai_gemini_3_pro_preview}, \texttt{Gemini-3-Flash}~\citep{google_vertexai_gemini_3_flash_preview}, and \texttt{Mistral-Large-3-675B-IT}~\citep{mistral_large_3_25_12_docs}.

\subsection{Evaluation Protocols}

All evaluations are conducted zero-shot using a shared prompt template, with provider-recommended decoding values throughout (see Appendix~\ref{app:inference_settings} for details).

For scalable scoring across both multiple-choice and free-form questions, we extract and
normalize the final answer from each response, and \texttt{Gemini-2.5-Flash} with temperature 0
then decides whether it matches the gold answer. We call this procedure LLM-based answer
scoring, since it checks answer-level correctness rather than overall response quality.
Section~\ref{sec:llm-judge} validates it against human correctness labels. 
Each model is evaluated over three trials, and we report mean accuracy and standard deviation.

\subsection{Is LLM-based Answer Scoring Reliable?}
\label{sec:llm-judge}
\begin{figure}[h]
    \centering
    \includegraphics[width=1\columnwidth]{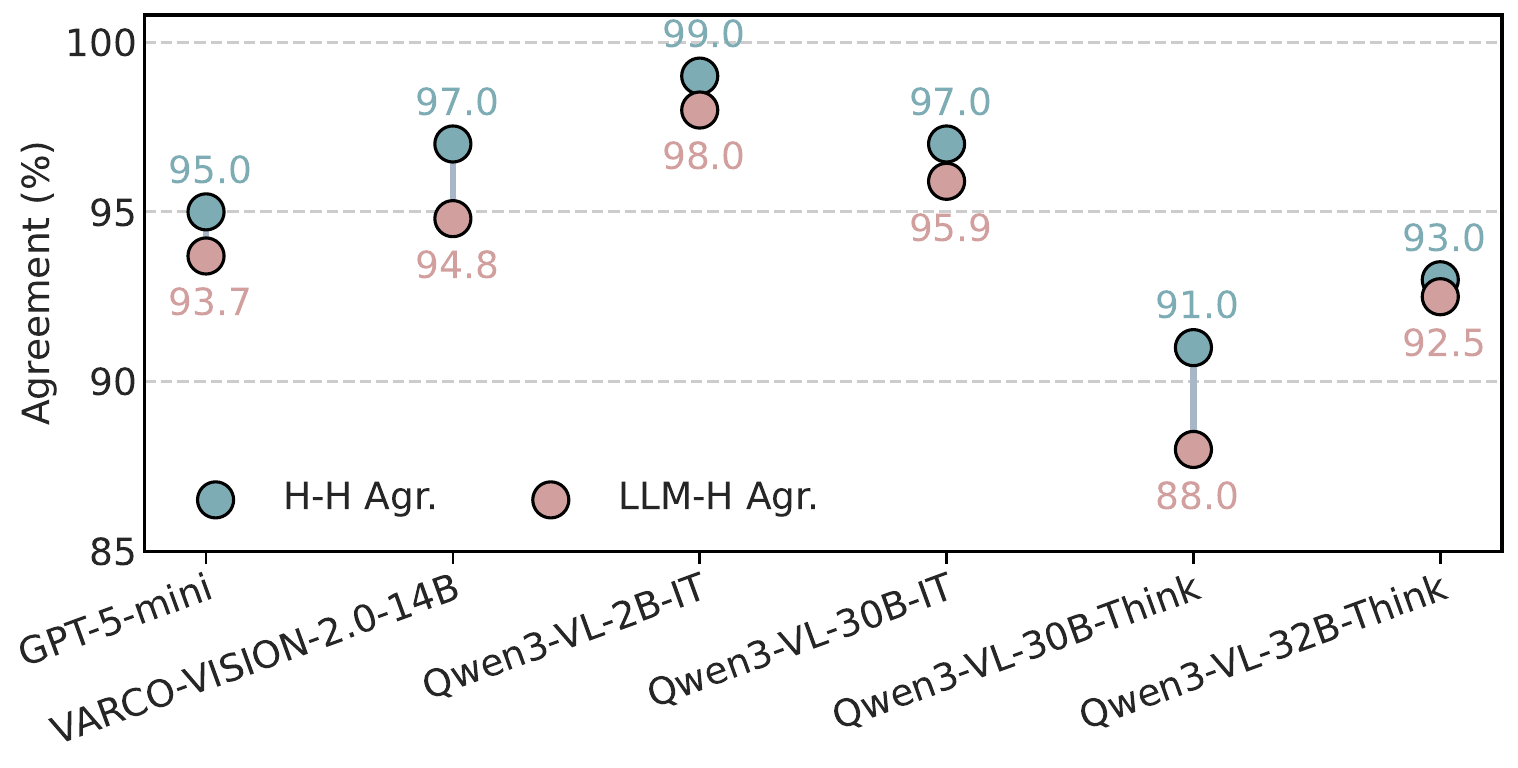}
    \caption{\footnotesize \textbf{Human alignment of the LLM scorer.}
    Each x-axis label indicates the source model of the sampled responses.
    We independently sample and re-annotate 100 outputs per model.
    H-H Agr. denotes inter-annotator agreement between human annotators, and LLM-H Agr. denotes agreement between human annotations and the LLM scorer.
    The LLM scorer is \texttt{Gemini-2.5-Flash}, the same judge used throughout our evaluation pipeline.}
    \label{fig:judge_validation}
\end{figure}

To validate the LLM scorer's evaluation capability, we conduct a human alignment study on 600 examples, sampled from six model runs (100 outputs each) and balanced across question formats.
Three annotators assign binary labels for model-response correctness and mark whether each response is complete or not (e.g., terminated mid-reasoning or degeneration).

As shown in Figure~\ref{fig:judge_validation}, inter-annotator agreement is consistently high, ranging from 0.91 to 0.99, which indicates that correctness labels are generally well defined.
LLM scorer also aligns well with human annotations, achieving agreement between 0.88 and 0.98 across models.
Although alignment varies by model, lower LLM--human agreement tends to coincide with lower human--human agreement, suggesting that these cases are better explained by outputs that are difficult for both humans and the LLM judge to interpret than by bias toward a particular model family.
Some annotation noise therefore remains inevitable, but we reduce its impact by evaluating each model over three independent runs and reporting mean performance and standard deviation. For more details on judging validation analysis, see Appendix~\ref{app:judge_alignment}.

\begin{table*}[h]
\centering
\setlength{\abovecaptionskip}{4pt}
\small
\resizebox{\textwidth}{!}{%
\begin{tabular}{lcccccccccc}
\toprule
\multicolumn{1}{l}{\textbf{Model}} &
\multicolumn{1}{c}{\shortstack{\textbf{Arts} \\ \textbf{\& Design}}} &
\multicolumn{1}{c}{\shortstack{\textbf{Business} \\ \textbf{\& Public}}} &
\multicolumn{1}{c}{\shortstack{\textbf{CS} \\ \textbf{\& IT}}} &
\multicolumn{1}{c}{\textbf{Engineering}} &
\multicolumn{1}{c}{\shortstack{\textbf{General}}} &
\multicolumn{1}{c}{\shortstack{\textbf{Law} \\ \textbf{\& Ethics}}} &
\multicolumn{1}{c}{\shortstack{\textbf{Math} \\ \textbf{\& Stats}}} &
\multicolumn{1}{c}{\shortstack{\textbf{Natural} \\ \textbf{Sciences}}} &
\multicolumn{1}{c}{\shortstack{\textbf{Social} \\ \textbf{Sciences}}} &
\multicolumn{1}{c}{\shortstack{\textbf{Overall} \\ \textbf{Acc.}}} \\
\midrule
Claude-Opus-4.5 & 28.00$_{6.53}$ & 32.38$_{3.56}$ & 25.73$_{2.19}$ & 23.12$_{2.31}$ & 21.93$_{3.28}$ & 24.07$_{6.93}$ & 19.54$_{3.25}$ & 22.94$_{1.22}$ & 35.19$_{4.72}$ & 24.51$_{0.54}$ \\

Claude-Sonnet-4.5 & 16.00$_{0.00}$ & 24.76$_{5.39}$ & 14.62$_{4.60}$ & 18.58$_{2.04}$ & 12.28$_{5.41}$ & 37.04$_{5.24}$ & 17.24$_{5.63}$ & 18.40$_{0.81}$ & 19.44$_{2.27}$ & 18.55$_{1.31}$ \\

Gemini-3-Flash & 48.00$_{0.00}$ & 42.86$_{7.00}$ & 48.54$_{1.65}$ & 46.52$_{1.32}$ & 21.93$_{3.28}$ & 46.30$_{9.44}$ & 37.93$_{0.00}$ & 46.10$_{1.84}$ & 56.48$_{3.46}$ & 45.14$_{0.98}$ \\

Gemini-3-Pro & \textbf{50.67$_{4.99}$} & \textbf{61.90$_{1.35}$} & \textbf{57.31$_{0.83}$} & \textbf{54.04$_{2.28}$} & \textbf{27.19$_{1.24}$} & \textbf{46.30$_{5.24}$} & \textbf{48.28$_{2.82}$} & \textbf{52.60$_{3.23}$} & \textbf{58.33$_{2.27}$} & \textbf{52.42$_{0.94}$} \\

Mistral-Large-3-675B-IT & 20.00$_{3.27}$ & 16.19$_{1.35}$ & 15.79$_{3.79}$ & 15.32$_{2.28}$ & 4.39$_{1.24}$ & 16.67$_{4.54}$ & 6.90$_{2.82}$ & 16.23$_{0.53}$ & 21.30$_{3.46}$ & 15.15$_{1.17}$ \\

GPT-5 & 20.00$_{3.27}$ & 26.67$_{3.56}$ & 26.90$_{2.19}$ & 30.35$_{1.40}$ & 24.56$_{3.28}$ & 18.52$_{2.62}$ & 34.48$_{2.82}$ & 33.55$_{2.51}$ & 45.37$_{3.46}$ & 30.57$_{0.40}$ \\

GPT-5-Mini & 14.67$_{6.80}$ & 25.71$_{2.33}$ & 14.04$_{2.48}$ & 18.58$_{1.40}$ & 17.54$_{4.96}$ & 24.07$_{11.42}$ & 29.89$_{3.25}$ & 24.89$_{1.70}$ & 31.48$_{1.31}$ & 21.32$_{1.04}$ \\

Grok-4 & 28.00$_{0.00}$ & 24.29$_{4.29}$ & 27.19$_{2.63}$ & 26.38$_{1.28}$ & 18.42$_{2.63}$ & 16.67$_{0.00}$ & 34.48$_{0.00}$ & 22.73$_{4.55}$ & 31.94$_{1.39}$ & 25.44$_{0.88}$ \\

Grok-4.1-Fast & 26.67$_{3.77}$ & 20.95$_{1.35}$ & 17.54$_{1.43}$ & 20.00$_{0.35}$ & 9.65$_{2.48}$ & 7.41$_{2.62}$ & 12.64$_{3.25}$ & 22.73$_{1.59}$ & 25.93$_{3.46}$ & 19.78$_{0.13}$ \\
\bottomrule
\end{tabular}
}
\caption{\footnotesize
\textbf{Accuracy (\%) on the KMMMU-Adv by disciplines.}
Overall accuracy is averaged across disciplines.
Mean accuracy is reported in percentage, with standard deviation shown as a subscript.
The best result is shown in \textbf{bold}.
}
\label{tab:hard-result-overall}
\end{table*}
%  ======================================
% MAIN TABLE 끝=================================================

\section{Results}

\subsection{Main Results}

\textbf{KMMMU remains challenging, even for strong multimodal models.}
As shown in Table~\ref{tab:main-result-overall}, on the full set, the strongest open-weight model reaches 42.05\% overall accuracy, while the best Korean-focused open-weight model, \texttt{VARCO-VISION-2.0-14B}, reaches 27.55\%.
This gap suggests that Korean-language specialization alone is insufficient for expert-level multimodal reasoning, and that strong performance still depends heavily on overall model capacity.

\textbf{Model scale consistently improves performance, but explicit test-time reasoning yields smaller and less consistent gains.}
Within the \texttt{Qwen3-VL} family, larger models generally outperform smaller ones, with especially large gains in disciplines such as \textit{Math \& Stats} and \textit{Social Sciences}.
This suggests that train-time scaling, including greater model capacity and broader knowledge acquisition, remains important for KMMMU.
By contrast, reasoning variants show only modest or uneven improvements over their non-reasoning counterparts, indicating that simply increasing test-time deliberation is not sufficient when errors stem from missing knowledge, weak grounding, or incorrect multimodal interpretation.

Performance also varies substantially across disciplines.
\textbf{While stronger models improve markedly in some areas, \textit{General} and \textit{Arts \& Design} remain persistent bottlenecks, with only limited gains even at larger scales.}
This pattern suggests that KMMMU requires more than surface-level recognition, demanding multimodal grounding, contextual interpretation, and discipline-specific knowledge.

A similar pattern appears on \textit{KMMMU-Adv} (Table~\ref{tab:hard-result-overall}).
\texttt{Gemini-3-Pro} achieves the best overall accuracy at 52.42\%, followed by \texttt{Gemini-3-Flash} at 45.14\%, while the remaining models perform substantially worse.
Discipline-level variation also remains strong: \textit{General} is again one of the weakest areas, with \texttt{Gemini-3-Pro} reaching only 27.19\%, far below its scores in other disciplines.
Taken together, these results show that \textbf{KMMMU-Adv sharply exposes residual weaknesses in reasoning, multimodal understanding, and discipline-specific interpretation.}

\begin{table}[h]
\centering
\setlength{\abovecaptionskip}{4pt}
\scriptsize
\resizebox{\columnwidth}{!}{%
\begin{tabular}{lrrrr}
\toprule
\textbf{Model} &
\multicolumn{1}{c}{\shortstack{\textbf{Non} \\ \textbf{Korean--spec}}} &
\multicolumn{1}{c}{\shortstack{\textbf{Korean} \\ \textbf{--spec}}} &
\multicolumn{1}{c}{\shortstack{\textbf{Raw} \\ \textbf{gap}}} &
\multicolumn{1}{c}{\shortstack{\textbf{Controlled} \\ \textbf{gap}}} \\
\midrule
\multicolumn{5}{l}{\textbf{Open-Weight Multilingual Models}} \\
\midrule
% Qwen3-VL-2B-IT & 14.20 & 14.78 & +0.58 & -1.53 \\
% Qwen3-VL-4B-IT & 19.87 & 20.00 & +0.13 & -1.04 \\
Qwen3-VL-8B-IT & 24.33 & 27.11 & +2.78 & +4.01 \\
Qwen3-VL-30B-A3B-IT & 30.16 & 22.00 & -8.16 & -5.97 \\
Qwen3-VL-30B-A3B-Thinking & 36.32 & 26.67 & -9.65 & -7.99 \\
Qwen3-VL-32B-IT & 36.56 & 26.22 & -10.33 & -8.09 \\
Qwen3-VL-32B-Thinking & 38.70 & 28.33 & -10.37 & -10.00 \\
Qwen3-VL-235B-A22B-IT & 40.62 & 27.11 &
\tightcell{red!12}{-13.51} &
\tightcell{red!12}{-13.43} \\
Qwen3-VL-235B-A22B-Thinking & 43.18 & 30.22 & -12.96 & -11.17 \\
Llama-4-Maverick-17B-IT & 39.96 & 28.44 & -11.51 & -12.35 \\
Llama-4-Scout-17B-IT & 34.27 & 27.44 & -6.83 & -7.97 \\
\midrule
\multicolumn{5}{l}{\textbf{Open-Weight Korean Models}} \\
\midrule
HyperCLOVAX-SEED-Vision-3B & 17.57 & 24.22 & 
\tightcell{green!12}{+6.66} &
\tightcell{green!12}{+5.15} \\
% \textbf{+6.66} & 
% \textbf{+5.15} \\
VARCO-VISION-2.0-14B & 27.70 & 26.11 & -1.59 & -4.18 \\
\bottomrule
\end{tabular}
}
\caption{\footnotesize
\textbf{Performance on KMMMU-Local subset.}
Raw gap is the accuracy difference between Korean-specific and non-Korean-specific questions; controlled gap is the discipline-controlled difference.
Negative values indicate worse performance on Korean-specific questions.
The largest positive gap is green shaded, and the largest negative gap is red shaded.
}
\label{tab:korean-specific-gap-full}
\end{table}

% \vspace{-2mm}
\subsection{What makes KMMMU-Local hard?}
\label{sec:xling_ablation}

We investigate whether the difficulty of \textit{KMMMU-Local} stems from Korean linguistic processing or from the localized knowledge the items embed.
Because \textit{KMMMU-Local} items are unevenly distributed across disciplines, particularly in \textit{Business \& Public} and \textit{Law \& Ethics}, we report both raw gaps and discipline-controlled gaps in Table~\ref{tab:korean-specific-gap-full}.

\paragraph{Multilingual scale does not close the gap.}
Most multilingual open-weight models show lower accuracy on \textit{KMMMU-Local} even after controlling for discipline composition.
The gap is especially pronounced among the strongest models: \texttt{Qwen3-VL-235B-A22B-IT}, \texttt{Qwen3-VL-235B-A22B-Thinking}, and \texttt{Llama-4-Maverick-17B-IT} show controlled gaps of $-13.43$, $-11.17$, and $-12.35$ percentage points, respectively.
For Qwen3-VL-235B-A22B-IT, the 95\% CI of this gap is $[-19.27, -7.44]$, and the deficit
is concentrated in Business \& Public and Law \& Ethics (Appendix~\ref{app:local_ci}).

\paragraph{Korean-focused training is not sufficient either.}
\texttt{HyperCLOVAX-SEED-Vision-3B} shows a positive controlled gap of $+5.15$ points, but \texttt{VARCO-VISION-2.0-14B} still shows a negative gap of $-4.18$ points. Being a Korean-focused model thus helps in some cases but does not reliably resolve the difficulty of Korean-specific institutional and terminological knowledge.

\paragraph{The deficit is localized knowledge, not Korean-language processing.}
To isolate the role of the input language, we conduct a controlled cross-lingual ablation on 200 Korean-specific paired items, translating the textual question and answer options into English while keeping the visual input, answer, and question type fixed (Table~\ref{tab:xling_ablation_main}).
Both \texttt{GPT-5} and \texttt{Gemini-3-Flash} perform \emph{worse} on the English translations: \texttt{GPT-5} drops from 42.5\% to 29.0\% and \texttt{Gemini-3-Flash} from 58.0\% to 40.0\%, with paired discordance statistically significant in both cases and substantially more Korean-only correct cases than English-only correct cases.
Translating the question text strips localized terminology, institutional references, and the grounding between Korean context and visual evidence; the difficulty therefore reflects the localized knowledge that current models do not yet encode, regardless of language coverage or Korean focus.
See Appendix~\ref{app:xling_ablation} for the full paired-outcome breakdown.

% 이런 말 써도 되려나..
% These two Korean-focused models also differ in their foundations: \texttt{HyperCLOVAX-SEED-Vision-3B} starts from NAVER's proprietary HyperCLOVAX-SEED-Text-Base-3B, while \texttt{VARCO-VISION-2.0-14B} uses \texttt{Qwen3-14B} as its language backbone and SigLIP2 as its vision encoder.
% This architectural difference may partly explain why Korean-focused models do not show a uniform pattern on Korean-specific items.

% \begin{table}[h]
% \centering
% \scriptsize
% \setlength{\tabcolsep}{4pt}
% \begin{tabular}{lccc}
% \toprule
% \textbf{Model} &
% \textbf{H-H Agr.} & \textbf{LLM-H Agr.} & \textbf{No Answer} \\
% \midrule
% GPT-5-mini            & 95.0 & 93.7 & 1 \\
% VARCO-VISION-2.0-14B        & 97.0 & 94.8 & 4 \\
% Qwen3-VL-2B-IT  & 99.0 & 98.0 & 51 \\
% Qwen3-VL-30B-A3B-IT & 97.0 & 95.9 & 18 \\
% Qwen3-VL-30B-A3B-Think & 91.0 & 88.0 & 22 \\
% Qwen3-VL-32B-Think & 93.0 & 92.5 & 31 \\
% \bottomrule
% \end{tabular}
% \caption{\footnotesize \textbf{Human alignment of LLM-Judge.}
% We report inter-annotator agreement (H-H Agr.), agreement between human annotations and LLM-Judge (LLM-H Agr.), and the no-answer rate on 100 sampled outputs per model.}
% \label{tab:judge_validation}
% \end{table}

\begin{table}[t]
\centering
\setlength{\abovecaptionskip}{4pt}
\scriptsize
\setlength{\tabcolsep}{3pt}
\begin{tabular}{lrrrrr}
\toprule
\textbf{Model} & \textbf{Ko Acc.} & \textbf{En Acc.} & \textbf{$\Delta$} & \textbf{Ko/En-only} & \textbf{$p$} \\
\midrule
GPT-5 & 42.5 & 29.0 & +13.5 & 33/6 & $1.43{\times}10^{-5}$ \\
Gemini-3-Flash & 58.0 & 40.0 & +18.0 & 38/2 & $1.49{\times}10^{-9}$ \\
\bottomrule
\end{tabular}
\caption{\footnotesize
\textbf{Cross-lingual ablation on 200 Korean-specific paired items.}
We translate the textual question and answer options into English.
Ko/En-only denotes paired discordant cases answered correctly only in Korean or only in English.
$p$ is from an exact McNemar test.
}
\label{tab:xling_ablation_main}
% \vspace{-3mm}
\end{table}

% For example, on multiple-choice questions, some models output the content of the correct option rather than its index, which can cause the judge to mark an otherwise correct response as incorrect.

% In this section, we conduct targeted manual error analysis by reading through selected model generations.
% Our analyses examine paired reversals between \texttt{Qwen3-VL-32B-IT} and \texttt{Qwen3-VL-32B-Thinking}, Korean-specific comparisons between \texttt{Qwen3-VL-235B-A22B-IT} and \texttt{HyperCLOVAX-SEED-Vision-3B}, and representative bottleneck cases from persistently difficult disciplines, with additional reference to corresponding \texttt{Qwen3-VL-235B-A22B-Thinking} outputs where relevant.

\section{Error Analysis}
In this section, we conduct a targeted manual error analysis over selected model generations. 
The full inspection protocol is provided in Appendix~\ref{app:error_analysis_details}. 
% We inspected 75 paired reversal cases between \texttt{Qwen3-VL-32B-IT} and \texttt{Qwen3-VL-32B-Thinking}, sampled as 25 cases each from \textit{Math \& Stats}, \textit{Engineering}, and \textit{Natural Sciences}. We also inspected 50 Korean-specific incorrect cases from the 300-item set, with 25 errors sampled from each of \texttt{Qwen3-VL-235B-A22B-IT} and \texttt{HyperCLOVAX-SEED-Vision-3B}. Finally, we examined representative failure cases from persistently difficult disciplines, especially \textit{Arts \& Design} and \textit{General}, with reference to corresponding \texttt{Qwen3-VL-235B-A22B-Thinking} outputs where relevant.

Across inspected cases, errors split into three patterns: incomplete answer composition (Section~\ref{subsec:reasoning_effects}), knowledge recall and category matching failures (Section~\ref{subsec:korean_specific_errors}), and weak rule induction on symbolic problems (Section~\ref{subsec:disciplinary_bottlenecks}).

\begin{figure*}
    \centering
    \includegraphics[width=0.9\linewidth]{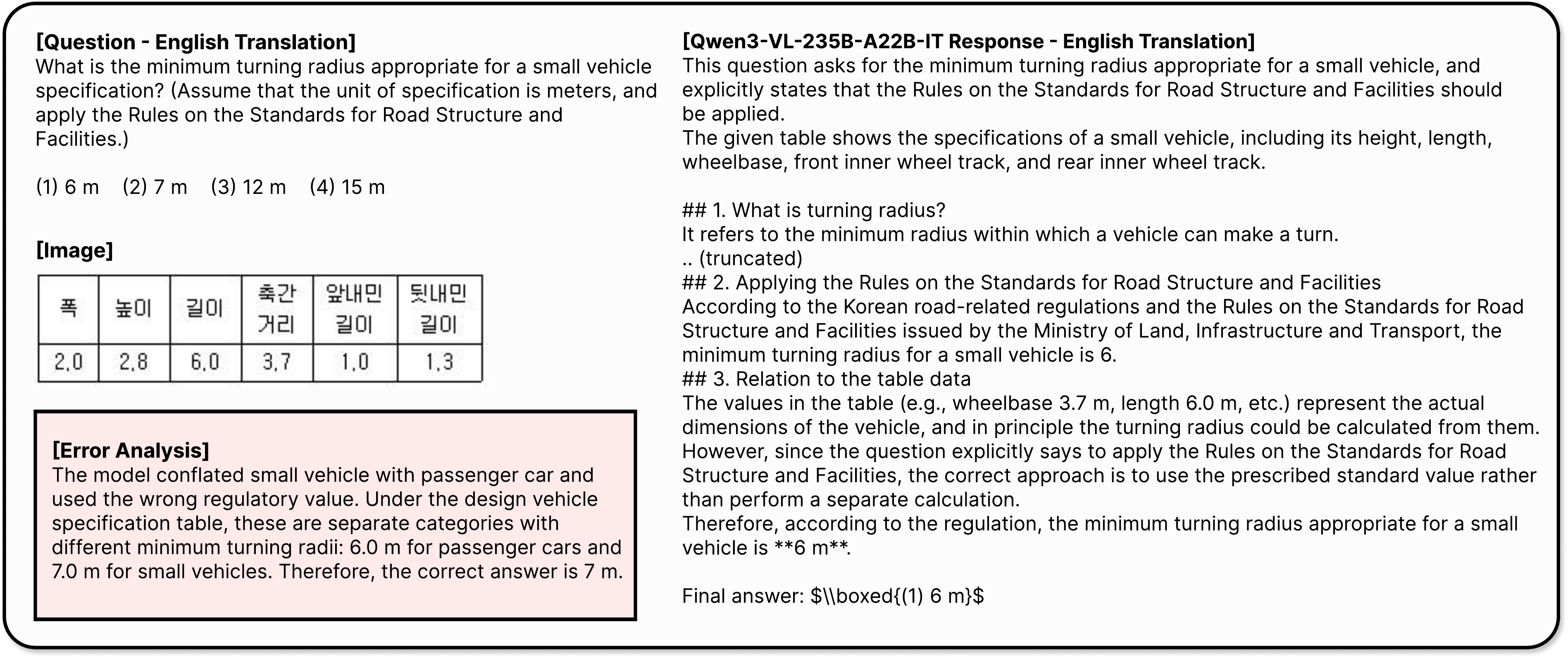}
\caption{\footnotesize \textbf{Example of a Korean-specific regulatory category mismatch.}
\texttt{Qwen3-VL-235B-A22B-IT} reads the table correctly, but maps \textit{small vehicle} to the wrong category and applies the wrong standard. This is a failure of institutional knowledge recall and lexical category matching, not OCR.}
    \label{fig:qwen235-korean-specific}
% \vspace{-2mm}
\end{figure*}

\subsection{Post-perceptual Effects of Reasoning}
\label{subsec:reasoning_effects}

\paragraph{Different failure patterns across quantitative domains.}
Although reasoning improves overall performance in some quantitative disciplines, especially \textit{Math \& Stats} (43.91$\rightarrow$49.93), its remaining failures follow different patterns across domains.
To examine this, we sampled 25 items each from \textit{Math \& Stats}, \textit{Engineering}, and \textit{Natural Sciences} among questions answered correctly by \texttt{Qwen3-VL-32B-IT} but incorrectly by \texttt{Qwen3-VL-32B-Thinking}, as these disciplines show contrasting reasoning effects.
The clearest pattern appears in \textit{Math \& Stats}. In our inspected sample, 72\% (18/25) of these reversals were not caused by obviously worse intermediate reasoning, but by \textit{answer finalization failure}.
The thinking model often developed a partially correct or plausible solution path, but stopped before producing a fully resolved final answer.
By contrast, reversals in \textit{Engineering} and \textit{Natural Sciences} more often reflected incorrect problem framing than incomplete finalization.
In these cases, the thinking model sometimes appears to map partial visual or textual cues onto a familiar \textit{device type, curve pattern, control category}, or \textit{physical scenario} too early, and then elaborate that interpretation into a coherent but incorrect solution (see Appendix~\ref{app:it_vs_thinking} for detailed examples).

\begin{figure}[h]
    \centering
    \includegraphics[width=1\columnwidth]{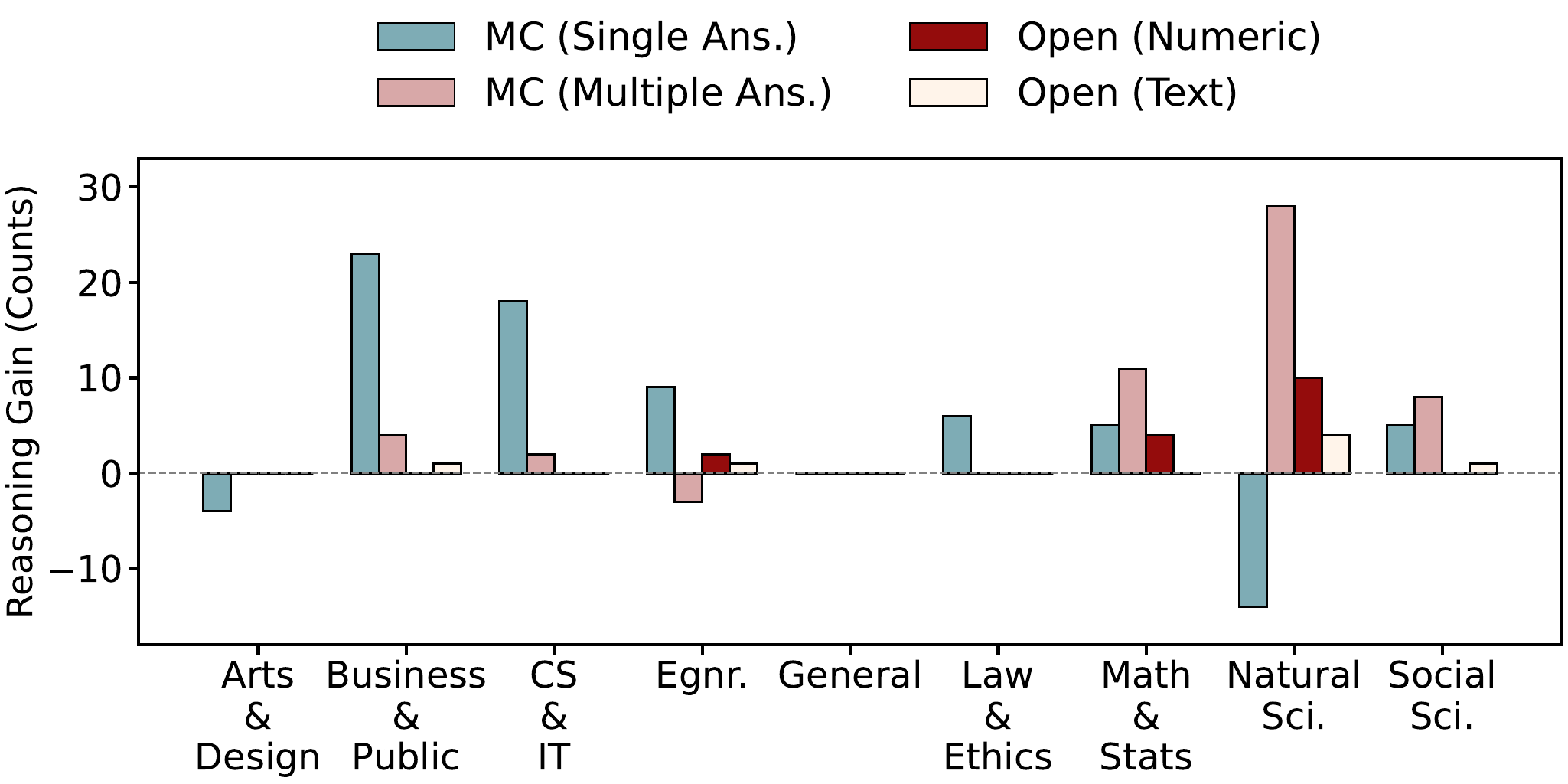}
    \caption{\footnotesize\textbf{Reasoning gain by discipline and question type in Qwen3-VL-32B (IT vs.\ Thinking).}}
    \label{fig:qwen_reasoning_gain}
% \vspace{-3mm}
\end{figure}

\paragraph{Reasoning gains in answer composition tasks.}
Among cases where \texttt{Qwen3-VL-32B-Thinking} succeeds and \texttt{Qwen3-VL-32B-IT} fails, the clearest gains appear on questions that require \textit{answer composition}.
This is especially visible for open-ended questions that ask for multiple requested outputs and for multiple-choice questions with multiple correct answers. In such cases, the \textit{Thinking} variant does a notably better job in formatting their responses for questions requiring multiple outputs, while the \textit{Instruct} variant often misses to do so, even after solving correctly. The non-reasoning model is  more likely to provide a subset of the requested components whereas the reasoning model is more likely to preserve the required answer structure and return all necessary components.

This tendency is also reflected in the aggregate pattern across question types (Figure~\ref{fig:qwen_reasoning_gain}).
Reasoning gains are large for \textit{multiple-answer} and \textit{open-form} questions,
% , including numerical and text items.
These findings suggest that explicit reasoning helps most with constraint tracking, structured decomposition, and complete answer assembly, so its benefits appear more in \textit{output completeness} than in knowledge recovery, helping explain why the gains in the main results are uneven.

Across visual modalities, the two variants show broadly similar performance, and inspected differences rarely come from one variant clearly reading or missing the image while the other does not (Appendix~\ref{app:it_vs_thinking}, Figure~\ref{fig:qwen_visual_acc_compare}).

\subsection{Knowledge recall and category matching failure in Korean-specific questions}
\label{subsec:korean_specific_errors}

Interestingly, although \texttt{Qwen3-VL-235B-A22B-IT} substantially outperforms \texttt{HyperCLOVAX-SEED-Vision-3B} on the full benchmark (39.44 vs.\ 18.14), the gap narrows on Korean-specific items.
In a 300-item comparison, the two models achieve relatively similar performance, scoring 83/300 and 72/300, respectively (Figure~\ref{tab:korean-specific-gap-full}). This reduced separation suggests that general reasoning ability provides limited advantage on Korean-specific items, many of which depend on regulation-specific knowledge or fine-grained administrative distinctions.

Figure~\ref{fig:qwen235-korean-specific} is a representative failure case concerning regulation-specific terminology. Korean law distinguishes \textit{소형차} (small vehicle), defined by an engine displacement of 1000 to 1600 cc, from \textit{승용차} (passenger vehicle), which refers to cars with up to 15 seats. However, \texttt{Qwen3-VL-235B-A22B-IT} appears to collapse both terms into the same English expression during intermediate reasoning, producing an incorrect answer. 
Similar patterns are reported by \citet{son2025pushing}, who show that multilingual models often translate inputs into a preferred language, introducing noise and reducing task performance.
Thus, the main bottleneck is not visual perception but the mapping between Korean legal terms and their institution-specific categories.

\subsection{Disciplinary Bottlenecks}
\label{subsec:disciplinary_bottlenecks}
Among the subject groups, \textit{Arts \& Design} and \textit{General} remain consistently difficult across models, suggesting bottlenecks that are not readily resolved by either scale or reasoning mode.
The inspected failures point to different sources of difficulty in the two categories.

In \textit{General}, many failures occur in items sourced from the KLO (Korea Linguistic Olympiad) exam.
These questions often combine several skills in a single problem, including linguistic puzzle solving, Korean orthography, semantic change, dictionary ordering, and symbol-to-sound or symbol-to-word rule induction from a small set of examples.
In many failures, models identify some local patterns but fail to recover the full rule system, leading to plausible but unsupported answers.
This points to a limitation in few-shot symbolic induction.

In \textit{Arts \& Design}, many failures instead involve exact technical labels for specialized visual conventions.
As shown in Appendix~\ref{app:disciplinary_examples}, some models recognize the relevant visual cues but choose a nearby standardized term rather than the correct one.
This is especially common when distinguishing related expert categories such as \textit{local} versus \textit{partial projection}, \textit{cutting line} versus \textit{revolved-section line}, or similar notation symbols.
These cases suggest that the bottleneck is not basic visual recognition, but precise mapping from visual conventions to domain-specific terms.
Taken together, these results suggest two complementary directions for improvement.
For \textit{Arts \& Design}, models may need broader exposure to specialized materials, and for \textit{General}, models may need training examples that require rule induction, linguistic reasoning, and knowledge retrieval within the single problem.

% Each of these problems require huge cognitive load to solve, mixing heterogeneous problem types such as linguistics and notation puzzles, Korean orthography and semantic change, dictionary ordering and some also requiring the model to infer a latent symbol-to-sound or symbol-to-word rule from a small set of examples.
% 원래문장
% Taken together, these results suggest two complementary directions for improvement. For \textit{Arts \& Design}, stronger performance may require pretraining on materials from niche domains, particularly sources that contain Korean-specific technical terminology and conventions. For \textit{General}, gains may depend more on post-training with instruction data that imposes higher cognitive load, requiring models to coordinate multiple abilities, such as pattern induction, linguistic reasoning, and knowledge retrieval, within a single problem.

\section{Conclusion}
We introduce \textbf{KMMMU}, a native Korean benchmark for expert-level multimodal understanding in culturally and institutionally grounded settings.
Across 3,466 carefully verified questions, KMMMU shows that current MLLMs remain far from robust on Korean real-world assessment materials.
Our findings suggest that key failures arise less from insufficient reasoning depth than from weak convention-to-label mapping, few-shot symbolic induction, localized knowledge recall, and familiarity with domain-specific standards and terminology.
These bottlenecks help explain the limited and uneven gains from reasoning, the persistent difficulty of Korean-specific content, and strong disciplinary variation in performance.
We hope KMMMU to serve as a rigorous benchmark for evaluating expert-level Korean multimodal understanding and a practical testbed for developing more culturally grounded and institutionally aware MLLMs.

\section*{Limitations}

\paragraph{Scope and factor separation.}
KMMMU targets expert-level multimodal understanding in native Korean exam-style settings.
Its results may not generalize to everyday perception, interactive image use, or non-exam multimodal tasks.
KMMMU-Local is a diagnostic subset for localized institutional, cultural, legal, and terminological knowledge, but Korean language processing, domain difficulty, visual complexity, and local knowledge often co-occur.
Thus, our controlled analyses help characterize these factors but do not fully separate their independent effects.

\paragraph{Filtering and subset bias.}
Model-based filtering is used as a difficulty-selection signal, while item validity is supported separately by Korean examination sources, answer keys, OCR checks, image--text verification, and human review.
Because the final set retains items missed by multiple models, KMMMU is not a random sample of Korean exam questions.
KMMMU-Adv is also model-dependent and should be interpreted as a baseline-relative stress test, not a model-neutral measure of intrinsic difficulty.

\paragraph{Annotation, evaluation, and contamination.}
KMMMU uses LLM-assisted metadata labels and LLM-based answer scoring for mixed-format answers.
Although we audit labels, manually verify Korean-specific positives, and validate judge outputs against human annotations, residual taxonomy and grading noise may remain.
Our source selection and prefix-completion analysis reduce contamination concerns, but they cannot rule out exposure to source documents or near-duplicates because model training data are not fully transparent.
Finally, results reflect the evaluated models and zero-shot protocol; newer models, prompts, decoding settings, retrieval, or tool use may change scores and rankings.

\section*{Acknowledgments}
This work was supported by the National Research Foundation of Korea (NRF) grant funded by the Ministry of Science and ICT (RS-2026-25477127).

\bibliography{custom}

\begin{thebibliography}{45}
\providecommand{\natexlab}[1]{#1}

\bibitem[{Abdin et~al.(2024)Abdin, Aneja, Behl, Bubeck, Eldan, Gunasekar, Harrison, Hewett, Javaheripi, Kauffmann et~al.}]{abdin2024phi}
Marah Abdin, Jyoti Aneja, Harkirat Behl, S{\'e}bastien Bubeck, Ronen Eldan, Suriya Gunasekar, Michael Harrison, Russell~J Hewett, Mojan Javaheripi, Piero Kauffmann, and 1 others. 2024.
\newblock Phi-4 technical report.
\newblock \emph{arXiv preprint arXiv:2412.08905}.

\bibitem[{Alayrac et~al.(2022)Alayrac, Donahue, Luc et~al.}]{alayrac2022flamingo}
Jean-Baptiste Alayrac, Jeff Donahue, Pauline Luc, and 1 others. 2022.
\newblock Flamingo: a visual language model for few-shot learning.
\newblock In \emph{NeurIPS}.

\bibitem[{{Anthropic}(2025)}]{anthropic_introducing_claude_opus_45_2025}
{Anthropic}. 2025.
\newblock \href {https://www.anthropic.com/news/claude-opus-4-5} {{Introducing Claude Opus 4.5}}.
\newblock Anthropic Newsroom.
\newblock Published: 2025-11-24. Accessed: 2026-01-05.

\bibitem[{Bai et~al.(2025)Bai, Cai, Chen, Chen, Chen, Cheng, Deng, Ding, Gao, Ge, Ge, Guo, Huang, Huang, Huang, Hui, Jiang, Li, Li, Li, Li, Lin, Lin, Liu, Liu, Liu, Liu, Liu, Liu, Lu, Luo, Lv, Men, Meng, Ren, Ren, Song, Sun, Tang, Tu, Wan, Wang, Wang, Wang, Wang, Xie, Xu, Xu, Xu, Yang, Yang, Yang, Yang, Yu, Zhang, Zhang, Zhang, Zheng, Zhong, Zhou, Zhou, Zhou, Zhu, and Zhu}]{bai2025qwen3vltechnicalreport}
Shuai Bai, Yuxuan Cai, Ruizhe Chen, Keqin Chen, Xionghui Chen, Zesen Cheng, Lianghao Deng, Wei Ding, Chang Gao, Chunjiang Ge, Wenbin Ge, Zhifang Guo, Qidong Huang, Jie Huang, Fei Huang, Binyuan Hui, Shutong Jiang, Zhaohai Li, Mingsheng Li, and 45 others. 2025.
\newblock \href {https://arxiv.org/abs/2511.21631} {Qwen3-vl technical report}.
\newblock \emph{Preprint}, arXiv:2511.21631.

\bibitem[{Cha et~al.(2025)Cha, Ju, Park, Lee, Yu, and Kim}]{cha2025varco}
Young-rok Cha, Jeongho Ju, SunYoung Park, Jong-Hyeon Lee, Younghyun Yu, and Youngjune Kim. 2025.
\newblock Varco-vision-2.0 technical report.
\newblock \emph{arXiv preprint arXiv:2509.10105}.

\bibitem[{Choi et~al.(2026{\natexlab{a}})Choi, Bae, Kweon, and Choi}]{choi2026kormedmcqa}
Byungjin Choi, Seongsu Bae, Sunjun Kweon, and Edward Choi. 2026{\natexlab{a}}.
\newblock Kormedmcqa-v: A multimodal benchmark for evaluating vision-language models on the korean medical licensing examination.
\newblock \emph{arXiv preprint arXiv:2602.13650}.

\bibitem[{Choi et~al.(2026{\natexlab{b}})Choi, Son, Lee, Kim, Ko, Lim, Eungyeol, Kim, Hong, and Song}]{choi2026users}
Dasol Choi, Guijin Son, Hanwool Lee, Minhyuk Kim, Hyunwoo Ko, Teabin Lim, Ahn Eungyeol, Jungwhan Kim, Seunghyeok Hong, and Youngsook Song. 2026{\natexlab{b}}.
\newblock What users leave unsaid: Under-specified queries limit vision-language models.
\newblock \emph{arXiv preprint arXiv:2601.06165}.

\bibitem[{Comanici et~al.(2025)Comanici, Bieber, Schaekermann, Pasupat, Sachdeva, Dhillon, Blistein, Ram, Zhang, Rosen et~al.}]{comanici2025gemini}
Gheorghe Comanici, Eric Bieber, Mike Schaekermann, Ice Pasupat, Noveen Sachdeva, Inderjit Dhillon, Marcel Blistein, Ori Ram, Dan Zhang, Evan Rosen, and 1 others. 2025.
\newblock Gemini 2.5: Pushing the frontier with advanced reasoning, multimodality, long context, and next generation agentic capabilities.
\newblock \emph{arXiv preprint arXiv:2507.06261}.

\bibitem[{{Google Cloud}(2025{\natexlab{a}})}]{google_vertexai_gemini_3_flash_preview}
{Google Cloud}. 2025{\natexlab{a}}.
\newblock \href {https://docs.cloud.google.com/vertex-ai/generative-ai/docs/models/gemini/3-flash} {{Gemini 3 Flash (Preview) | Generative AI on Vertex AI}}.
\newblock Google Cloud Documentation.
\newblock Release date: 2025-12-17. Accessed: 2026-01-05.

\bibitem[{{Google Cloud}(2025{\natexlab{b}})}]{google_vertexai_gemini_3_pro_preview}
{Google Cloud}. 2025{\natexlab{b}}.
\newblock \href {https://docs.cloud.google.com/vertex-ai/generative-ai/docs/models/gemini/3-pro} {{Gemini 3 Pro (Preview) | Generative AI on Vertex AI}}.
\newblock Google Cloud Documentation.
\newblock Release date: 2025-11-18. Accessed: 2026-01-05.

\bibitem[{Hendrycks et~al.(2020)Hendrycks, Burns, Basart, Zou, Mazeika, Song, and Steinhardt}]{hendrycks2020measuring}
Dan Hendrycks, Collin Burns, Steven Basart, Andy Zou, Mantas Mazeika, Dawn Song, and Jacob Steinhardt. 2020.
\newblock Measuring massive multitask language understanding.
\newblock \emph{arXiv preprint arXiv:2009.03300}.

\bibitem[{Hong et~al.(2025)Hong, Kim, Son, Kim, Hong, and Lee}]{hong2025kmmlu}
Seokhee Hong, Sunkyoung Kim, Guijin Son, Soyeon Kim, Yeonjung Hong, and Jinsik Lee. 2025.
\newblock From kmmlu-redux to kmmlu-pro: A professional korean benchmark suite for llm evaluation.
\newblock \emph{arXiv preprint arXiv:2507.08924}.

\bibitem[{Hwang et~al.(2025)Hwang, Kim, Lee, Kim, and Eun}]{hwang2025kreta}
Taebaek Hwang, Minseo Kim, Gisang Lee, Seonuk Kim, and Hyunjun Eun. 2025.
\newblock Kreta: A benchmark for korean reading and reasoning in text-rich vqa attuned to diverse visual contexts.
\newblock In \emph{Proceedings of the 2025 Conference on Empirical Methods in Natural Language Processing}, pages 33409--33420.

\bibitem[{Ju et~al.(2024)Ju, Kim, Park, and Kim}]{ju2024varcovisionexpandingfrontierskorean}
Jeongho Ju, Daeyoung Kim, SunYoung Park, and Youngjune Kim. 2024.
\newblock \href {https://arxiv.org/abs/2411.19103} {Varco-vision: Expanding frontiers in korean vision-language models}.
\newblock \emph{Preprint}, arXiv:2411.19103.

\bibitem[{Kim and Jung(2025)}]{kim2025koffvqa}
Yoonshik Kim and Jaeyoon Jung. 2025.
\newblock Koffvqa: An objectively evaluated free-form vqa benchmark for large vision-language models in the korean language.
\newblock In \emph{Proceedings of the Computer Vision and Pattern Recognition Conference}, pages 575--585.

\bibitem[{Le~Bras et~al.(2020)Le~Bras, Swayamdipta, Bhagavatula, Zellers, Peters, Sabharwal, and Choi}]{le2020adversarial}
Ronan Le~Bras, Swabha Swayamdipta, Chandra Bhagavatula, Rowan Zellers, Matthew Peters, Ashish Sabharwal, and Yejin Choi. 2020.
\newblock Adversarial filters of dataset biases.
\newblock In \emph{International conference on machine learning}, pages 1078--1088. Pmlr.

\bibitem[{Li et~al.(2023{\natexlab{a}})Li, Li, Savarese, and Hoi}]{li2023blip2}
Junnan Li, Dongxu Li, Silvio Savarese, and Steven Hoi. 2023{\natexlab{a}}.
\newblock Blip-2: Bootstrapping language-image pre-training with frozen image encoders and large language models.
\newblock In \emph{ICML}.

\bibitem[{Li et~al.(2023{\natexlab{b}})Li, Du, Zhou, Wang, Zhao, and Wen}]{li2023evaluating}
Yifan Li, Yifan Du, Kun Zhou, Jinpeng Wang, Wayne~Xin Zhao, and Ji-Rong Wen. 2023{\natexlab{b}}.
\newblock Evaluating object hallucination in large vision-language models.
\newblock \emph{arXiv preprint arXiv:2305.10355}.

\bibitem[{Liu et~al.(2023)Liu, Li, Wu, and Lee}]{liu2023llava}
Haotian Liu, Chunyuan Li, Qingyang Wu, and Yong~Jae Lee. 2023.
\newblock Visual instruction tuning.
\newblock In \emph{NeurIPS}.

\bibitem[{{Meta}(2025{\natexlab{a}})}]{meta_llama4_maverick_17b_128e_instruct_hf}
{Meta}. 2025{\natexlab{a}}.
\newblock \href {https://huggingface.co/meta-llama/Llama-4-Maverick-17B-128E-Instruct} {{meta-llama/Llama-4-Maverick-17B-128E-Instruct}}.
\newblock Hugging Face model card and weights.
\newblock Model release date: 2025-04-05. Accessed: 2026-01-05.

\bibitem[{{Meta}(2025{\natexlab{b}})}]{meta_llama4_scout_17b_16e_hf}
{Meta}. 2025{\natexlab{b}}.
\newblock \href {https://huggingface.co/meta-llama/Llama-4-Scout-17B-16E} {{meta-llama/Llama-4-Scout-17B-16E}}.
\newblock Hugging Face model card and weights.
\newblock Model release date: 2025-04-05. Accessed: 2026-01-05.

\bibitem[{{Mistral AI}(2025)}]{mistral_large_3_25_12_docs}
{Mistral AI}. 2025.
\newblock \href {https://docs.mistral.ai/models/mistral-large-3-25-12} {{Mistral Large 3 (v25.12) | Mistral Docs}}.
\newblock Mistral Documentation.
\newblock Dated: 2025-12-02. Accessed: 2026-01-05.

\bibitem[{{NAVER HyperCLOVAX}(2025)}]{naver_hyperclovax_seed_vision_instruct_3b_hf}
{NAVER HyperCLOVAX}. 2025.
\newblock \href {https://huggingface.co/naver-hyperclovax/HyperCLOVAX-SEED-Vision-Instruct-3B} {{HyperCLOVAX-SEED-Vision-Instruct-3B}}.
\newblock Hugging Face model card and weights.
\newblock Model release date: 2025-04-24 (as stated in repository license). Accessed: 2026-01-05.

\bibitem[{Niu et~al.(2025)Niu, Liu, Gu, Wang, Ouyang, Zhao, Chu, He, Wu, Zhang, Jin, Liang, Zhang, Zhang, Qu, Ren, Sun, Zheng, Ma, Tang, Niu, Miao, Dong, Qian, Zhang, Chen, Wang, Zhao, Wei, Li, Wang, Xu, Cao, Chen, Wu, Gu, Lu, Wang, Lin, Shen, Zhou, Zhang, Zang, Dong, Wang, Zhang, Bai, Chu, Li, Wu, Wu, Li, Wang, Tu, Xu, Chen, Qiao, Zhou, Lin, Zhang, and He}]{niu2025mineru25decoupledvisionlanguagemodel}
Junbo Niu, Zheng Liu, Zhuangcheng Gu, Bin Wang, Linke Ouyang, Zhiyuan Zhao, Tao Chu, Tianyao He, Fan Wu, Qintong Zhang, Zhenjiang Jin, Guang Liang, Rui Zhang, Wenzheng Zhang, Yuan Qu, Zhifei Ren, Yuefeng Sun, Yuanhong Zheng, Dongsheng Ma, and 42 others. 2025.
\newblock \href {https://arxiv.org/abs/2509.22186} {Mineru2.5: A decoupled vision-language model for efficient high-resolution document parsing}.
\newblock \emph{Preprint}, arXiv:2509.22186.

\bibitem[{Onohara et~al.(2025)Onohara, Miyai, Imajuku, Egashira, Baek, Yue, Neubig, and Aizawa}]{onohara2025jmmmu}
Shota Onohara, Atsuyuki Miyai, Yuki Imajuku, Kazuki Egashira, Jeonghun Baek, Xiang Yue, Graham Neubig, and Kiyoharu Aizawa. 2025.
\newblock Jmmmu: A japanese massive multi-discipline multimodal understanding benchmark for culture-aware evaluation.
\newblock In \emph{Proceedings of the 2025 Conference of the Nations of the Americas Chapter of the Association for Computational Linguistics: Human Language Technologies (Volume 1: Long Papers)}, pages 932--950.

\bibitem[{OpenAI(2024)}]{openai2024mmmlu}
OpenAI. 2024.
\newblock Multilingual massive multitask language understanding (mmmlu).
\newblock \url{https://huggingface.co/datasets/openai/MMMLU}.

\bibitem[{{OpenAI}(2025)}]{openai_introducing_gpt5_2025}
{OpenAI}. 2025.
\newblock \href {https://openai.com/index/introducing-gpt-5/} {{Introducing GPT-5}}.
\newblock OpenAI.
\newblock Published: 2025-08-07. Accessed: 2026-01-05.

\bibitem[{Park et~al.(2025)Park, Baek, Kim, Heo, Chang, and Choo}]{park2025evaluating}
ChaeHun Park, Yujin Baek, Jaeseok Kim, Yu-Jung Heo, Du-Seong Chang, and Jaegul Choo. 2025.
\newblock Evaluating visual and cultural interpretation: The k-viscuit benchmark with human-vlm collaboration.
\newblock In \emph{Proceedings of the 63rd Annual Meeting of the Association for Computational Linguistics (Volume 1: Long Papers)}, pages 21960--21974.

\bibitem[{Park and Kim(2025)}]{park2025konet}
Sanghee Park and Geewook Kim. 2025.
\newblock Evaluating multimodal generative ai with korean educational standards.
\newblock In \emph{Proceedings of the 2025 Conference of the Nations of the Americas Chapter of the Association for Computational Linguistics: Human Language Technologies (Volume 2: Short Papers)}, pages 671--688.

\bibitem[{Phan et~al.(2025)Phan, Gatti, Han, Li, Hu, Zhang, Zhang, Shaaban, Ling, Shi et~al.}]{phan2025humanity}
Long Phan, Alice Gatti, Ziwen Han, Nathaniel Li, Josephina Hu, Hugh Zhang, Chen Bo~Calvin Zhang, Mohamed Shaaban, John Ling, Sean Shi, and 1 others. 2025.
\newblock Humanity's last exam.
\newblock \emph{arXiv preprint arXiv:2501.14249}.

\bibitem[{Son et~al.(2025{\natexlab{a}})Son, Lee, Kim, Kim, Muennighoff, Choi, Park, Yoo, and Biderman}]{son2025kmmlu}
Guijin Son, Hanwool Lee, Sungdong Kim, Seungone Kim, Niklas Muennighoff, Taekyoon Choi, Cheonbok Park, Kang~Min Yoo, and Stella Biderman. 2025{\natexlab{a}}.
\newblock Kmmlu: Measuring massive multitask language understanding in korean.
\newblock In \emph{Proceedings of the 2025 Conference of the Nations of the Americas Chapter of the Association for Computational Linguistics: Human Language Technologies (Volume 1: Long Papers)}, pages 4076--4104.

\bibitem[{Son et~al.(2025{\natexlab{b}})Son, Yang, Patel, Agarwal, Ko, Lim, Panda, Kim, Drolia, Choi et~al.}]{son2025pushing}
Guijin Son, Donghun Yang, Hitesh~Laxmichand Patel, Amit Agarwal, Hyunwoo Ko, Chanuk Lim, Srikant Panda, Minhyuk Kim, Nikunj Drolia, Dasol Choi, and 1 others. 2025{\natexlab{b}}.
\newblock Pushing on multilingual reasoning models with language-mixed chain-of-thought.
\newblock \emph{arXiv preprint arXiv:2510.04230}.

\bibitem[{Team and Google(2023)}]{team2023gemini}
Gemini Team and Google. 2023.
\newblock Gemini: A family of highly capable multimodal models.
\newblock \emph{arXiv preprint arXiv:2312.11805}.

\bibitem[{Team et~al.(2025)Team, Kamath, Ferret, Pathak, Vieillard, Merhej, Perrin, Matejovicova, Ram{\'e}, Rivi{\`e}re et~al.}]{team2025gemma}
Gemma Team, Aishwarya Kamath, Johan Ferret, Shreya Pathak, Nino Vieillard, Ramona Merhej, Sarah Perrin, Tatiana Matejovicova, Alexandre Ram{\'e}, Morgane Rivi{\`e}re, and 1 others. 2025.
\newblock Gemma 3 technical report.
\newblock \emph{arXiv preprint arXiv:2503.19786}.

\bibitem[{Wang et~al.(2025)Wang, Gao, Gu, Pu, Cui, Wei, Liu, Jing, Ye, Shao et~al.}]{wang2025internvl3_5}
Weiyun Wang, Zhangwei Gao, Lixin Gu, Hengjun Pu, Long Cui, Xingguang Wei, Zhaoyang Liu, Linglin Jing, Shenglong Ye, Jie Shao, and 1 others. 2025.
\newblock Internvl3.5: Advancing open-source multimodal models in versatility, reasoning, and efficiency.
\newblock \emph{arXiv preprint arXiv:2508.18265}.

\bibitem[{Wang et~al.(2024)Wang, Ma, Zhang, Ni, Chandra, Guo, Ren, Arulraj, He, Jiang et~al.}]{wang2024mmlu}
Yubo Wang, Xueguang Ma, Ge~Zhang, Yuansheng Ni, Abhranil Chandra, Shiguang Guo, Weiming Ren, Aaran Arulraj, Xuan He, Ziyan Jiang, and 1 others. 2024.
\newblock Mmlu-pro: A more robust and challenging multi-task language understanding benchmark.
\newblock \emph{Advances in Neural Information Processing Systems}, 37:95266--95290.

\bibitem[{Wei et~al.(2024)Wei, Karina, Chung, Jiao, Papay, Glaese, Schulman, and Fedus}]{wei2024measuring}
Jason Wei, Nguyen Karina, Hyung~Won Chung, Yunxin~Joy Jiao, Spencer Papay, Amelia Glaese, John Schulman, and William Fedus. 2024.
\newblock Measuring short-form factuality in large language models.
\newblock \emph{arXiv preprint arXiv:2411.04368}.

\bibitem[{{xAI}(2026)}]{xai_grok_models_docs}
{xAI}. 2026.
\newblock \href {https://docs.x.ai/docs/models} {{Models and Pricing (xAI API Documentation)}}.
\newblock xAI Developer Docs.
\newblock Accessed: 2026-01-05.

\bibitem[{Yue et~al.(2024)Yue, Ni, Zhang, Zheng, Liu, Zhang, Stevens, Jiang, Ren, Sun et~al.}]{yue2024mmmu}
Xiang Yue, Yuansheng Ni, Kai Zhang, Tianyu Zheng, Ruoqi Liu, Ge~Zhang, Samuel Stevens, Dongfu Jiang, Weiming Ren, Yuxuan Sun, and 1 others. 2024.
\newblock Mmmu: A massive multi-discipline multimodal understanding and reasoning benchmark for expert agi.
\newblock In \emph{Proceedings of the IEEE/CVF Conference on Computer Vision and Pattern Recognition}, pages 9556--9567.

\bibitem[{Yue et~al.(2025)Yue, Zheng, Ni, Wang, Zhang, Tong, Sun, Yu, Zhang, Sun et~al.}]{yue2025mmmu}
Xiang Yue, Tianyu Zheng, Yuansheng Ni, Yubo Wang, Kai Zhang, Shengbang Tong, Yuxuan Sun, Botao Yu, Ge~Zhang, Huan Sun, and 1 others. 2025.
\newblock Mmmu-pro: A more robust multi-discipline multimodal understanding benchmark.
\newblock In \emph{Proceedings of the 63rd Annual Meeting of the Association for Computational Linguistics (Volume 1: Long Papers)}, pages 15134--15186.

\bibitem[{Zellers et~al.(2018)Zellers, Bisk, Schwartz, and Choi}]{zellers2018swag}
Rowan Zellers, Yonatan Bisk, Roy Schwartz, and Yejin Choi. 2018.
\newblock Swag: A large-scale adversarial dataset for grounded commonsense inference.
\newblock In \emph{Proceedings of the 2018 conference on empirical methods in natural language processing}, pages 93--104.

\bibitem[{Zellers et~al.(2019)Zellers, Holtzman, Bisk, Farhadi, and Choi}]{zellers2019hellaswag}
Rowan Zellers, Ari Holtzman, Yonatan Bisk, Ali Farhadi, and Yejin Choi. 2019.
\newblock Hellaswag: Can a machine really finish your sentence?
\newblock In \emph{Proceedings of the 57th annual meeting of the association for computational linguistics}, pages 4791--4800.

\bibitem[{Zhang et~al.(2024)Zhang, Du, Chen, Liang, Luo, Zheng, Zhu, Cheng, Xu, Guo et~al.}]{zhang2024cmmmu}
Ge~Zhang, Xinrun Du, Bei Chen, Yiming Liang, Tongxu Luo, Tianyu Zheng, Kang Zhu, Yuyang Cheng, Chunpu Xu, Shuyue Guo, and 1 others. 2024.
\newblock Cmmmu: A chinese massive multi-discipline multimodal understanding benchmark.
\newblock \emph{arXiv preprint arXiv:2401.11944}.

\bibitem[{Zhang et~al.(2023)Zhang, Aljunied, Gao, Chia, and Bing}]{zhang2023m3exam}
Wenxuan Zhang, Mahani Aljunied, Chang Gao, Yew~Ken Chia, and Lidong Bing. 2023.
\newblock M3exam: A multilingual, multimodal, multilevel benchmark for examining large language models.
\newblock \emph{Advances in Neural Information Processing Systems}, 36:5484--5505.

\bibitem[{Zhong et~al.(2024)Zhong, Cui, Guo, Liang, Lu, Wang, Saied, Chen, and Duan}]{zhong2024agieval}
Wanjun Zhong, Ruixiang Cui, Yiduo Guo, Yaobo Liang, Shuai Lu, Yanlin Wang, Amin Saied, Weizhu Chen, and Nan Duan. 2024.
\newblock Agieval: A human-centric benchmark for evaluating foundation models.
\newblock In \emph{Findings of the association for computational linguistics: NAACL 2024}, pages 2299--2314.

\end{thebibliography}
\clearpage
\appendix
\begin{figure*}[h]
    \centering
    \includegraphics[width=0.9\textwidth]{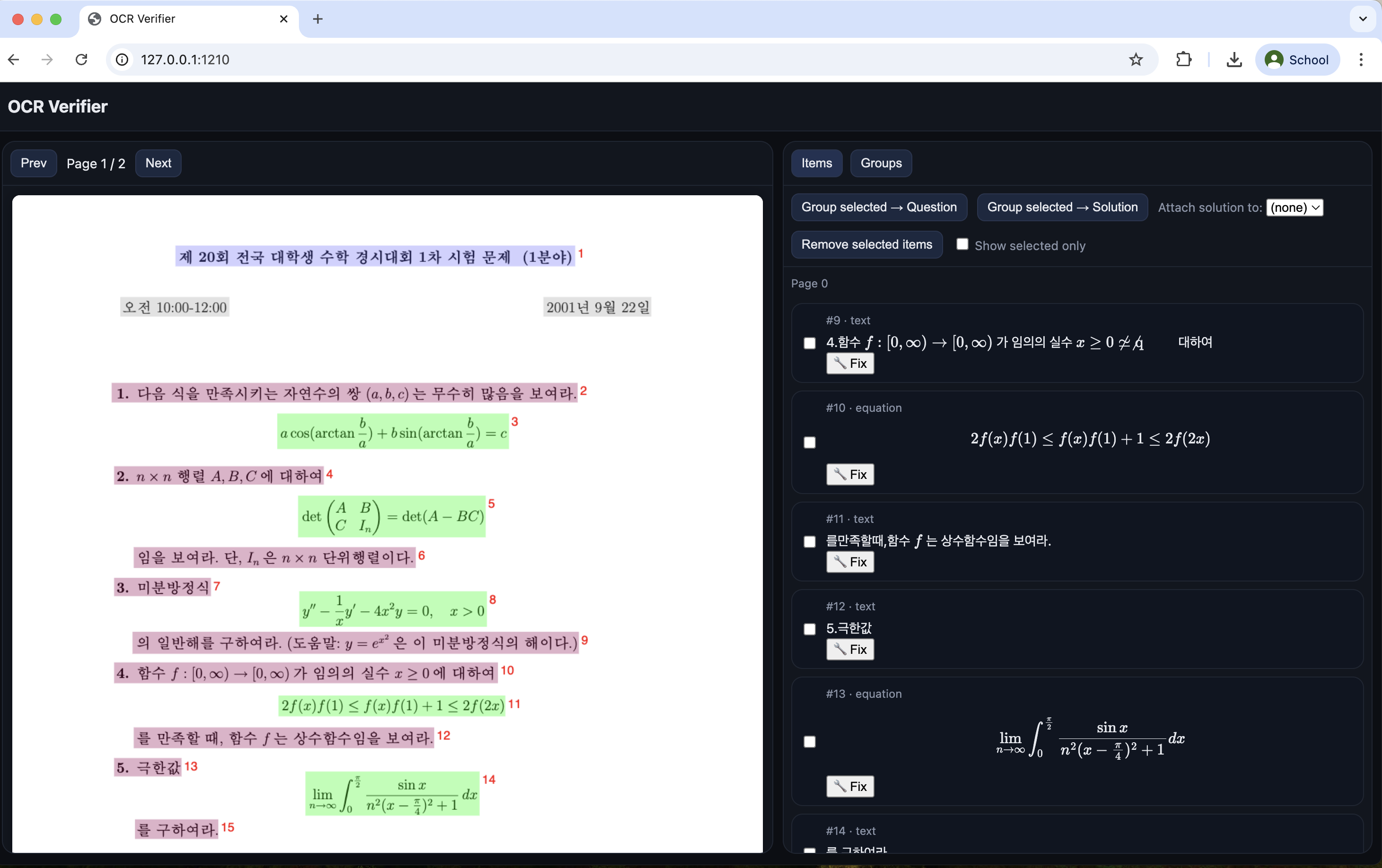}
    \caption{\footnotesize
    \textbf{Annotation tool interface used for OCR verification.} The tool displays the original PDF page on the left and the parsed text and images on the right, allowing annotators to correct OCR errors and validate image cropping in real time.}
    \label{fig:annotation_tool}
\end{figure*}

\section{Data Sources and Collection Scope}
\label{sec:appendix_sources}

KMMMU is collected from four high stakes sources in South Korea.
We summarize the collection scope for each source below.

\subsection{PSAT}
We annotate ten years of past examinations from civil service recruitment tracks.
The PSAT includes Language Logic, Data Interpretation, and Situational Judgment sections that assess logical reasoning and information integration.

\subsection{National Technical Qualifications}
We collect fifteen years of questions from 252 distinct certification exams, including Information Processing Engineer, Electric Engineer, and Fire Safety Manager.
These exams cover a wide range of technical domains across industrial and engineering fields.

\subsection{Olympiads}
To incorporate academically challenging reasoning problems, we gather ten years of Olympiad questions spanning middle school, high school, and university levels.
The collected problems focus primarily on mathematics and science.

\subsection{NCS}
We include three years of National Competency Standards examinations covering all ten competency areas, such as Communication, Numeracy, and Problem Solving.
These exams are used in recruitment for public sector organizations.

\section{Annotation and Quality Control Details}
\label{sec:appendix_annotation}

The construction of KMMMU uses a rigorous pipeline that combines automated processing with human verification to ensure high data fidelity.

\subsection{Human Verification Interface}
We utilized a custom built annotation tool to verify and correct the output of the OCR pipeline.
Raw data digitized by MinerU-2.5~\cite{niu2025mineru25decoupledvisionlanguagemodel} often contained artifacts and formula errors.
Figure~\ref{fig:annotation_tool} shows the interface where five Korean annotators reviewed the parsed content against the original PDF source.
Annotators were instructed to
\begin{itemize}
    \item Correct LaTeX formatting for mathematical formulas
    \item Verify that image references in the text matched the cropped images
    \item Discard questions where essential visual information was illegible or missing.
\end{itemize}
All five annotators are native Korean speakers with at least a bachelor’s degree and prior experience in annotation or dataset curation.
They are also familiar with AI-related workflows, which helped them reliably identify OCR artifacts, formula corruption, and image--text mismatches during verification.

\subsection{Automatic Labeling and Taxonomy Consolidation}
\label{sec:appendix_labeling}

% \begin{table}[t]
% \centering
% \scriptsize
% \setlength{\tabcolsep}{5pt}
% \begin{tabular}{llcc}
% \toprule
% \textbf{Label type} & \textbf{$n$} & 
% \textbf{H--H agr.} & \textbf{H--L agr.} \\
% \midrule
% Visual modality & 300 & 99.0\% & 98.0\% \\
% Question format & 300 & 95.0\% & 96.0\% \\
% Korean-specificity & All positive cases (300) & -- & 100.0\% \\
% \bottomrule
% \end{tabular}
% \caption{\footnotesize
% \textbf{Human validation of metadata labels.}
% Human--human agreement is computed as average pairwise agreement among three annotators.
% Metadata validation denotes agreement between the initial LLM-assigned metadata label and the human-majority label for visual modality and question format. 
% For Korean-specificity, we manually verify all positive cases because false positives could inflate localization analyses.
% Most remaining mismatches involve ambiguous boundaries, such as diagram/table distinctions for visual modality and open-numeric versus open-text distinctions for question format.
% }
% \label{tab:taxonomy_audit}
% \end{table}

We annotate several auxiliary attributes to support analysis and stratified reporting, including discipline, visual modality type, question format and a Korean-specific flag.
All taxonomy labels are assigned using \texttt{Gemini-2.5-Flash}.
For each labeling job, the model is given the question text and its associated image, and outputs the most appropriate label.

We use an open labeling step that does not constrain predictions to a fixed label set.
This reduces forced assignments when an instance does not cleanly match a predefined taxonomy.
All label types are generated independently.

\paragraph{Manual audit and consolidation}
We conduct a manual audit by randomly sampling around 300 instances and reviewing the assigned labels.
Based on the audited outputs, we consolidate the discipline taxonomy through human curation into 45 sub-discipline categories and 9 macro discipline categories.

\paragraph{Verification of Korean-specific cases}
Because false positives can inflate localization analyses, we manually verify all instances labeled as Korean-specific.
We confirm that each positive case requires Korean-specific knowledge or context rather than general world knowledge expressed in Korean.

% \subsection{Adversarial Filtering Protocol}

% To ensure benchmark difficulty, we apply a multi-stage adversarial filtering pipeline~\citep{zellers2018swag, le2020adversarial} that removes instances solvable by current multimodal models without advanced reasoning.
% Starting from a manually verified pool of approximately 68{,}000 questions, we apply the following procedure.

% \begin{enumerate}
%     \item \textbf{Data cleaning and de-duplication.}
%     We first remove samples with invalid image links and de-duplicate near-duplicate questions across exam years using image and text similarity checks.

%     \item \textbf{Model-based adversarial filtering.}
%     We then sequentially filter the remaining candidate pool using four multimodal models:
%     \textsc{Phi-3.5-Vision-Instruct}~\citep{abdin2024phi},
%     \textsc{InternVL-3.5-38B}~\citep{wang2025internvl3_5},
%     \textsc{Gemini-2.5-Flash-Lite}, and
%     \textsc{Gemini-2.5-Flash}~\citep{comanici2025gemini}.
%     Each model is evaluated in a zero-shot setting, and questions answered correctly at each stage are removed from the candidate pool.

%     \item \textbf{Final retention.}
%     Only questions that remain unsolved after all four filtering stages are retained in the final benchmark.
% \end{enumerate}

% The resulting KMMMU benchmark contains \textbf{3,466} curated questions.

\subsection{Adversarial Filtering Protocol}
\label{app:adversarial_filtering}

To construct a challenging diagnostic benchmark, we apply a multi-stage adversarial filtering pipeline~\citep{zellers2018swag, le2020adversarial} as a difficulty-oriented selection mechanism.
Model-based filtering is used only to select challenging items; item validity is established separately through source-grounded answer keys and human verification.
Starting from a cleaned and manually verified pool of 68{,}214 questions, we apply the following procedure.

\begin{enumerate}
    \item \textbf{Data cleaning and de-duplication.}
    We remove samples with invalid image links and de-duplicate near-duplicate questions across exam years using image and text similarity checks.

    \item \textbf{Model-based adversarial filtering.}
    We sequentially filter the remaining candidate pool using four multimodal models:
    \texttt{Phi-3.5-Vision-Instruct}~\citep{abdin2024phi},
    \texttt{InternVL-3.5-38B}~\citep{wang2025internvl3_5},
    \texttt{Gemini-2.5-Flash-Lite}, and
    \texttt{Gemini-2.5-Flash}~\citep{comanici2025gemini}.
    Each model is evaluated in a zero-shot setting, and questions answered correctly at each stage are removed from the candidate pool.

    \item \textbf{Final retention.}
    Questions that remain unsolved after all four filtering stages are retained in the final benchmark.
\end{enumerate}

Table~\ref{tab:filtering_retention} reports the number of items retained after each filtering stage.
The resulting KMMMU benchmark contains \textbf{3,466} curated questions.

\begin{table}[t]
\centering
\scriptsize
\begin{tabular}{llrrr}
\toprule
Stage & Filtering model & Input & Removed & Remaining \\
\midrule
Initial & -- & -- & -- & 68{,}214 \\
1 & Phi-3.5-Vision-Instruct & 68{,}214 & 18{,}309 & 49{,}905 \\
2 & InternVL-3.5-38B & 49{,}905 & 8{,}390 & 41{,}515 \\
3 & Gemini-2.5-Flash-Lite & 41{,}515 & 18{,}368 & 23{,}147 \\
4 & Gemini-2.5-Flash & 23{,}147 & 19{,}681 & 3{,}466 \\
\bottomrule
\end{tabular}
\caption{\footnotesize
\textbf{Stage-wise retention during adversarial filtering.}
At each stage, questions answered correctly by the corresponding filtering model are removed from the candidate pool.
}
\label{tab:filtering_retention}
\end{table}

\section{Korean-Specific Context}
\label{sec:appendix_examples}

\begin{figure*}[h]
    \centering
    \includegraphics[width=0.95\textwidth]{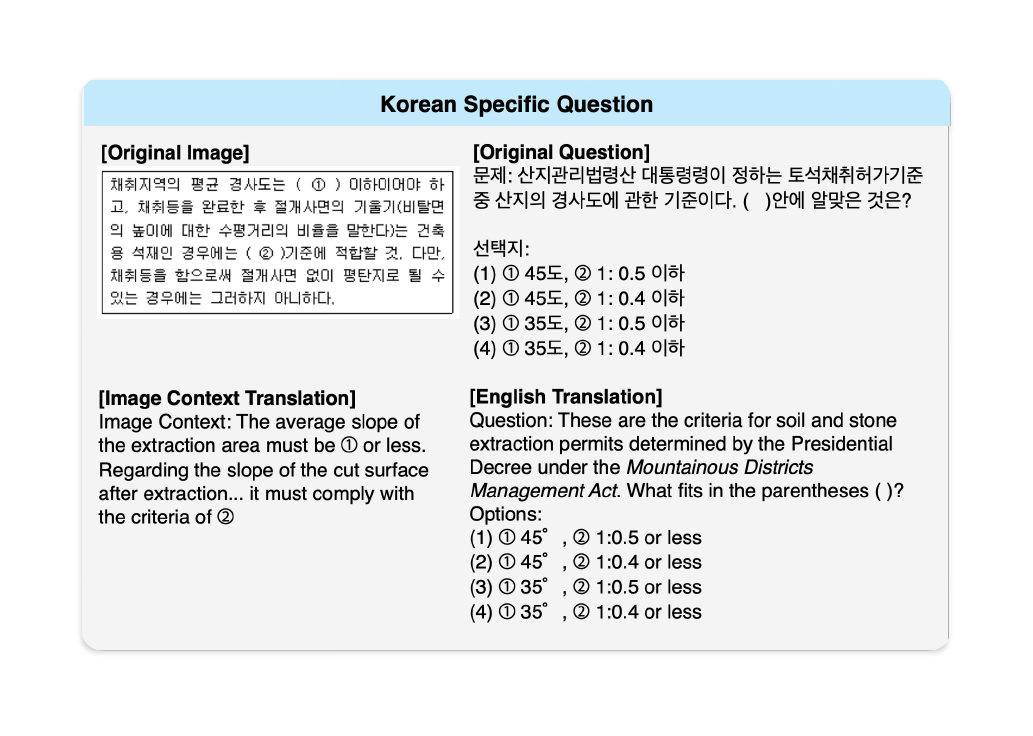}
    \captionsetup{skip=0pt}
    \caption{\footnotesize
    \textbf{Data Card for a Korean-Specific Question.} 
    The figure aggregates the raw inputs and their translations. 
    \textbf{[Original Image]} The original visual input containing a text-rich regulation box. 
    \textbf{[Original Question]} The original question text in Korean. 
    \textbf{[Translation]} English translations for both the visual context and the question. 
    Correctly answering this question requires retrieving specific legal provisions regarding slope limits for soil extraction permits in South Korea, demonstrating the benchmark's focus on localized expert knowledge.}
    \label{fig:ko_specific_example}
\end{figure*}

To provide a concrete illustration of KMMMU, Figure~\ref{fig:ko_specific_example} presents a \textit{Korean-Specific} instance from the benchmark.
Unlike standard multimodal benchmarks, which often emphasize culturally invariant knowledge such as Physics or Mathematics, KMMMU includes a dedicated subset of questions that require localized knowledge grounded in Korean institutional and legal contexts.
In this example, the input consists of an image containing regulation text and a corresponding question, and the model must interpret the visual text referring to the ``extraction area slope criteria'' in the specific context of South Korea's \textit{Mountainous Districts Management Act} to identify the correct legal standard (Option 3).
This example shows that solving such questions requires not only optical character recognition, but also grounded knowledge of Korean administrative law.

\section{Detailed Dataset Statistics}
\label{appendix:dataset_stat}

In this section, we provide a granular breakdown of the dataset composition.
Beyond the overview (Table~\ref{tab:dataset_overview}), we report (i) the distribution of fine-grained discipline categories (Table~\ref{tab:subject_category_dist}),
(ii) the question format distribution (Table~\ref{tab:question-type}),
and (iii) the examples of visual modalities (sub-visual modalities, Figure~\ref{fig:visual_modality_appendix})

% =========================
% Dataset Overview
% =========================
\begin{table}[h]
\centering
\scriptsize
\setlength{\tabcolsep}{6pt}
\renewcommand{\arraystretch}{1.06}
\begin{tabular}{lr}
\toprule
\textbf{Statistic} & \textbf{Count} \\
\midrule
Total Questions & 3,466 \\
KMMMU-Adv & 627 \\
Discipline Categories & 9 \\
Sub-discipline Categories & 45 \\
Visual Modality Types & 9 \\
% Task Types & 9 \\
Question Types & 5 \\
\midrule
Questions with in-image texts & 2,383 (68.75\%) \\
% Questions w/o in-image texts & 1,083 (31.25\%) \\
Korean-specific questions & 300 (8.65\%) \\
\bottomrule
\end{tabular}
\caption{\footnotesize
\textbf{Dataset distribution outline.} We report counts and percentages for key attributes such as in-image text and Korean-specific content.}
\label{tab:dataset_overview}
\end{table}

\begin{table*}[t]
\centering
\scriptsize
\begin{tabular}{l r l r l r}
\toprule
\textbf{Subject} & \textbf{Cnt} & \textbf{Subject} & \textbf{Cnt} & \textbf{Subject} & \textbf{Cnt} \\
\midrule
Physics & 474 & General Knowledge \& Interdisciplinary & 67 & Economics & 33 \\
Civil \& Structural Engineering & 332 & Industrial \& Systems Engineering & 64 & Earth \& Geological Sciences & 33 \\
Mechanical Engineering & 223 & Architecture \& Urban Studies & 57 & Library, Archival \& Information Science & 30 \\
Electrical \& Electronics Engineering & 221 & Human Resources \& Organizational Studies & 56 & Transportation \& Logistics & 29 \\
Computer Science & 164 & Linguistics \& Language Studies & 54 & Agriculture \& Life Sciences & 27 \\
Statistics \& Probability & 149 & Biology & 43 & Software \& Programming & 27 \\
Mathematics & 121 & Safety, Risk \& Reliability Engineering & 42 & Astronomy \& Space Science & 26 \\
Business Administration \& Management & 104 & Data Science \& Analytics & 41 & Psychology & 22 \\
Environmental Science \& Engineering & 99 & Finance \& Accounting & 40 & Cognitive \& Behavioral Sciences & 20 \\
Geography \& Spatial Studies & 95 & Sociology \& Social Sciences & 36 & Biomedical \& Health Sciences & 20 \\
Manufacturing \& Production Engineering & 90 & Chemical Engineering & 76 & Communication \& Media Studies & 17 \\
Public Administration \& Policy & 89 & Chemistry & 75 & Education \& Pedagogy & 16 \\
Design \& Visual Arts & 88 & Law \& Legal Studies & 74 & Marketing \& Consumer Studies & 15 \\
Chemical Engineering & 76 & Materials Science \& Metallurgy & 69 & Ethics \& Philosophy & 12 \\
Chemistry & 75 & Information Technology \& Systems & 69 & Artificial Intelligence \& Machine Learning & 12 \\
\bottomrule
\end{tabular}
\caption{\footnotesize
\textbf{Distribution of sub-discipline categories in KMMMU}.}
\label{tab:subject_category_dist}
\end{table*}

\begin{table}[h]
\centering
\scriptsize
\renewcommand{\arraystretch}{0.95}
\begin{tabular}{lrr}
\toprule
\textbf{Question Type} & \textbf{Count} & \textbf{\%} \\
\midrule
Multiple Choice (Single Answer) & 2{,}831 & 81.68  \\
Multiple Choice (Multiple Answers) & 346 & 9.98 \\
Open Format (Numerical / Calculation) & 207 & 5.98 \\
Open Format (Text) & 67 & 1.93 \\
Essay (Descriptive) & 15 & 0.43 \\
\midrule
\textbf{Total} & 3466 & 100 \\
\bottomrule
\end{tabular}
\caption{\footnotesize
\textbf{Question type distribution.}}
\label{tab:question-type}
\end{table}

\subsection{Discipline Category Distribution}
Table~\ref{tab:subject_category_dist} details the frequency of questions across 45 fine-grained Discipline categories.
The distribution reflects the emphasis on STEM (Science, Technology, Engineering, and Mathematics) fields, with \textit{Physics}, \textit{Civil Engineering}, and \textit{Mechanical Engineering} constituting the largest portions.
This heavy tail in engineering disciplines ensures that KMMMU serves as a robust benchmark for technical domain expertise.

\subsection{Question Format Distribution}
Because KMMMU contains both multiple-choice and free-form items, the answer format affects evaluation difficulty and failure modes.
Table~\ref{tab:question-type} reports the distribution of question formats in the benchmark.

\begin{figure}[h]
    \centering
    \includegraphics[width=1.0\columnwidth]{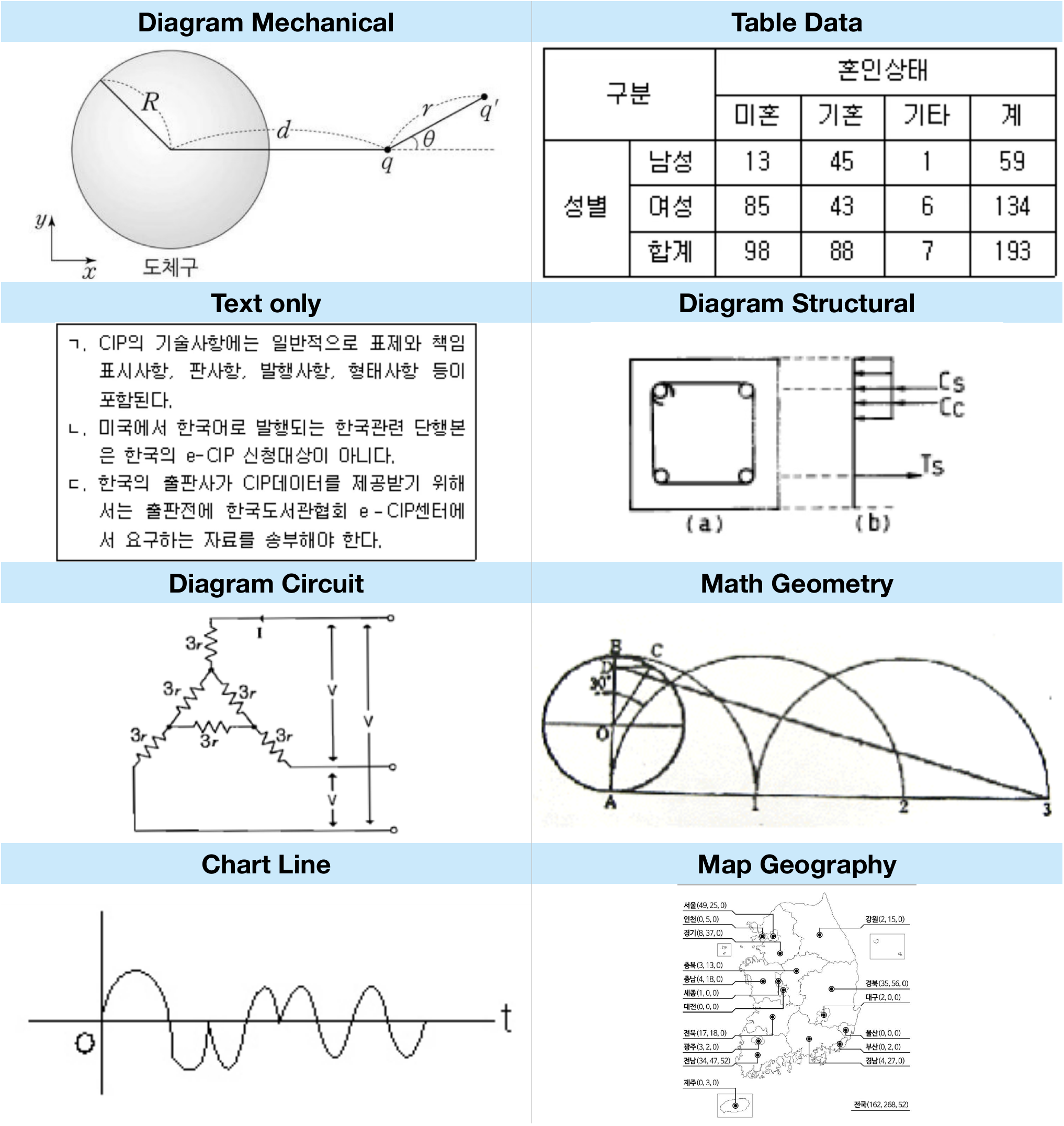}
    \caption{\footnotesize
    \textbf{Representative examples of fine-grained visual types in KMMMU.} 
    Before consolidation into the final macro-level visual modality categories, the dataset included diverse fine-grained visual types, such as specialized engineering diagrams, document-style text images, and South Korean geographic maps.}
    \label{fig:visual_modality_appendix}
\end{figure}

\subsection{Visual Modality Taxonomy}
\label{app:visual_modality_taxonomy}
KMMMU includes a wide range of fine-grained visual types, including circuit, mechanical, and structural diagrams, document-style text images, tables, mathematical figures, charts, maps, symbols, and photographs.
For analysis, we consolidate these fine-grained types into 9 macro-level visual modality categories.
Technical diagrams constitute a particularly large portion of the dataset, reflecting KMMMU's emphasis on professional and schematic visual reasoning.
Figure~\ref{fig:visual_modality_appendix} presents representative examples of these fine-grained visual types before consolidation.

\subsection{Question Type Taxonomy}
\label{sec:appendix_answer_type}

Table~\ref{tab:dataset_attributes} summarizes the distribution of answer formats within each macro subject.
This table clarifies which subjects are dominated by multiple choice items versus numerical or descriptive responses.
It also provides context for interpreting task-wise performance, since answer format affects both evaluation difficulty and failure modes.

\begin{table*}[t]
\centering
\scriptsize
% \resizebox{\textwidth}{!}{%
\begin{tabular}{lrrrrr}
\toprule
\textbf{Discipline} &
\textbf{\shortstack[l]{Multiple Choice\\(Single Answer)}} &
\textbf{\shortstack[l]{Multiple Choice\\(Multiple Answers)}} &
\textbf{\shortstack[l]{Open Format\\(Numerical)}} &
\textbf{\shortstack[l]{Open Format\\(Text)}} &
\textbf{\shortstack[l]{Essay\\(Descriptive)}} \\
\midrule
Engineering                & 1182 & 25  & 20  & 2  & 0  \\
Natural Sciences           & 450  & 180 & 154 & 27 & 11 \\
CS \& IT                   & 311  & 11  & 2   & 0  & 0  \\
Business \& Public         & 263  & 37  & 2   & 2  & 0  \\
Social Sciences            & 193  & 55  & 2   & 4  & 0  \\
Math \& Stats              & 161  & 36  & 27  & 3  & 0  \\
Law \& Ethics              & 103  & 1   & 0   & 0  & 0  \\
General/Interdisciplinary  & 89   & 0   & 0   & 30 & 4  \\
Arts \& Design             & 79   & 0   & 0   & 0  & 0  \\
\bottomrule
\end{tabular}
\caption{\footnotesize
\textbf{Dataset attributes by disciplines and answer type.}}
\label{tab:dataset_attributes}
\end{table*}

% =============================

\begin{figure*}
    \centering
    \includegraphics[width=1.0\linewidth]{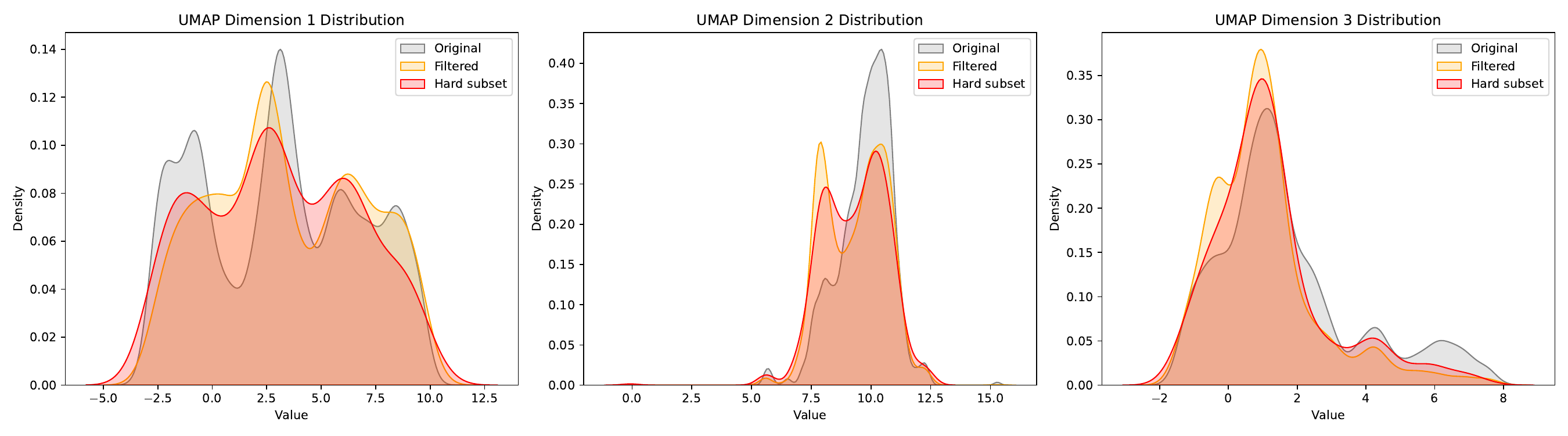}
    \caption{\footnotesize
    \textbf{Per-dimension density comparison after adversarial filtering.}
Kernel density estimates over the three UMAP dimensions for the original 68k corpus, the \textit{KMMMU}, and the \textit{KMMMU-Adv}. The filtered subsets broadly retain the major density peaks and multimodal trends of the original distribution, although the \textit{KMMMU-Adv} shows a somewhat larger deviation in Dimension 3.}
    \label{fig:embedding_kde}
\end{figure*}

\section{Additional Distributional Analysis}
\label{sec:appendix_distribution}

Figure~\ref{fig:embedding_kde} provides per-dimension density comparisons for the original 68k corpus, the KMMMU \textit{Full set}, and the \textit{KMMMU-Adv} in the 3D UMAP space. Across all three dimensions, the filtered subsets broadly preserve the major density peaks and overall multimodal structure of the original distribution. The \textit{KMMMU} remains especially close to the original corpus, while the \textit{KMMMU-Adv} shows a somewhat larger shift in parts of the latent space.

To quantify these differences, we compute the Kullback--Leibler (KL) divergence between the original distribution and each filtered subset along each UMAP dimension. For the \textit{Full set}, the divergence remains low across all three dimensions ($D_{KL}=0.1184$, $0.1459$, and $0.1437$ for Dimensions 1--3, respectively). The \textit{KMMMU-Adv} shows similarly low divergence on Dimensions 1 and 2 ($0.1081$ and $0.1699$), but a larger deviation on Dimension 3 ($0.3747$). Overall, these results are consistent with the main-text UMAP visualization: adversarial filtering increases difficulty while largely preserving the broader distributional structure of the original corpus.

\begin{table}[h]
\centering
\scriptsize
\begin{tabular}{lccc}
\toprule
\textbf{Model family} & \textbf{Temperature} & \textbf{Top-p} & \textbf{Top-k} \\
\midrule
Gemma-3 & 1.0 & 0.95 & 64 \\
Qwen3-VL-Instruct & 0.7 & 0.8 & 20 \\
Qwen3-VL-Thinking & 1.0 & 0.95 & 20 \\
Llama-4 & 0.6 & 0.9 & -- \\
HyperCLOVAX-SEED-Vision-3B & 1.0 & 1.0 & 50 \\
VARCO-VISION-2.0 & 1.0 & 1.0 & -- \\
\bottomrule
\end{tabular}
\caption{\footnotesize
\textbf{Decoding settings used for model evaluation.} Dashes indicate parameters that were not explicitly specified or not exposed by the corresponding provider interface.}
\label{tab:inference_settings}
\end{table}

\section{Inference Settings}
\label{app:inference_settings}

\subsection{Decoding Settings}
Table~\ref{tab:inference_settings} summarizes the main decoding settings used in our experiments.
For open-weight models, we follow the Hugging Face generation configuration or the official model-card recommendation when available.
If no model-specific setting is specified, we use the vLLM default sampling configuration.
For proprietary models, we use the default API settings provided by each vendor.

\subsection{Proprietary Model Provenance}
All evaluated proprietary models are publicly available models released in 2025. We access
the Gemini family through the Google Gemini Developer API, the GPT family through the OpenAI
API, and all other proprietary models through OpenRouter, using the default decoding settings
exposed by each endpoint. Table~\ref{tab:provenance} reports the exact API identifiers,
release dates, and access dates.

\begin{table}[h]
\centering
\small
\resizebox{\columnwidth}{!}{%
\begin{tabular}{lllll}
\toprule
Model & API identifier & Access & Released & Accessed \\
\midrule
GPT-5             & \texttt{gpt-5}                        & OpenAI API & 2025-08-07 & 2025-12-29 \\
GPT-5-mini        & \texttt{gpt-5-mini}                   & OpenAI API & 2025-08-07 & 2025-12-31 \\
Claude-Opus-4.5   & \texttt{anthropic/claude-opus-4.5}    & OpenRouter & 2025-11-24 & 2026-01-01 \\
Claude-Sonnet-4.5 & \texttt{anthropic/claude-sonnet-4.5}  & OpenRouter & 2025-09-29 & 2025-12-30 \\
Gemini-3-Pro      & \texttt{gemini-3-pro-preview}         & Gemini API & 2025-11-18 & 2025-12-31 \\
Gemini-3-Flash    & \texttt{gemini-3-flash-preview}       & Gemini API & 2025-12-17 & 2025-12-31 \\
Grok-4            & \texttt{x-ai/grok-4}                  & OpenRouter & 2025-07-09 & 2025-12-31 \\
Grok-4.1-Fast     & \texttt{x-ai/grok-4.1-fast}           & OpenRouter & 2025-11-19 & 2025-12-30 \\
Mistral-Large-3   & \texttt{mistralai/mistral-large-2512} & OpenRouter & 2025-12-02 & 2025-12-30 \\
\bottomrule
\end{tabular}}
\caption{\textbf{API identifiers and access information for proprietary models.}}
\label{tab:provenance}
\end{table}

\subsection{Prompt Set}
We use the same prompts for all evaluated models. The inference prompt asks the model to
place its final answer in \texttt{\textbackslash boxed\{\}}, and the scoring prompt asks
Gemini-2.5-Flash to decide whether the extracted answer matches the gold answer. An empty
extracted answer is scored FALSE.

\begin{promptbox}{Zero-shot Inference Prompt}
Solve the given question.
You should refer to question sentences and an image.
You can print your reasoning and solving process.
You may answer in any language that helps you solve the problem.
At the end, you MUST present the final answer enclosed in \boxed{ }.

The final line MUST strictly follow this format:
\boxed{your_final_answer}

Do not put anything outside the \boxed{ } in the final line.
\end{promptbox}

\begin{promptbox}{LLM-based Answer-Scoring Prompt}
You will be given a question, the gold answer, and a model response.
Your task is to judge whether the model response correctly answers the question.

Return ONLY one of the following tokens on the LAST line.

TRUE
FALSE

Do not explain.
Do not add any other text.
\end{promptbox}

% =======================

% 이건 새로 들어간 플롯임
\begin{figure}[h]
    \centering
    \includegraphics[width=1.0\linewidth]{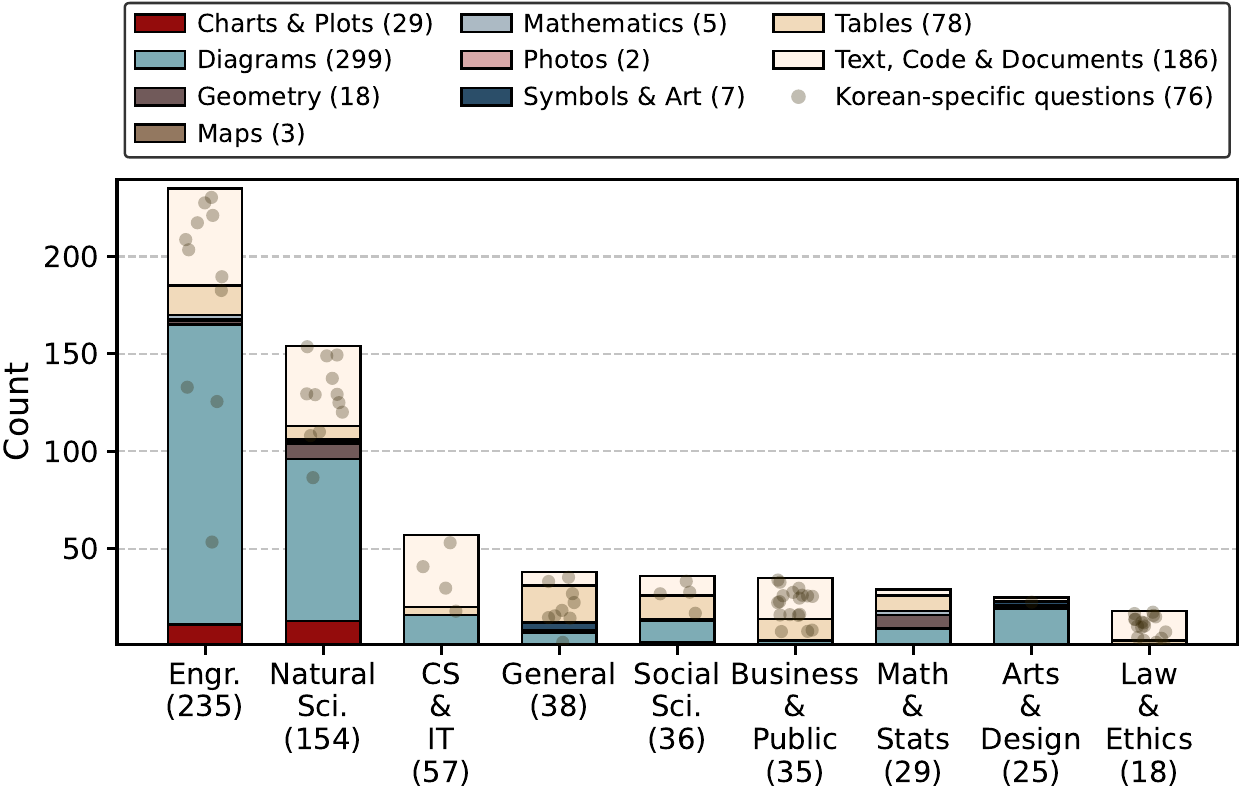}
    \caption{
    \footnotesize
\textbf{Discipline-wise visual modality composition
of \textit{KMMMU-Adv} subset.}
Stacked bars show the number of ques-
tions for each visual modality in each discipline, with
total counts shown beneath the labels. Scatter points
indicate Korean-specific (\textit{KMMMU-Local}) items overlaid on the corresponding discipline--modality segments.
The \textit{KMMMU-Adv} is concentrated in \textit{Engineering} and \textit{Natural Sciences}, similar to Full set.
}
    \label{fig:discipline_hard_visual_cross}
\end{figure}

%==============================

\begin{table}[h]
\centering
\scriptsize
\begin{tabular}{lrrr}
\toprule
\textbf{Split} & \textbf{Korean Specific} & \textbf{Not Korean Specific} & \textbf{Ratio (\%)} \\
\midrule
\textbf{KMMMU} & 300 & 3166 & 8.65 \\
\textbf{KMMMU-Adv} & 76 & 551  & 12.12 \\
\bottomrule
\end{tabular}
\caption{\footnotesize
\textbf{Number of Korean-specific questions each in Full set and \textit{KMMMU-Adv}.}}
\label{tab:korean_specific_distribution}
\end{table}

\section{Analysis of the KMMMU-Adv}
\label{app:hard_subset_analysis}

\subsection{Distributional Characteristics of the KMMMU-Adv}

We analyze the structural composition of the \textit{KMMMU-Adv} to better understand the types of instances that contribute to systematic model failures.

We first examine the prevalence of Korean-specific items (Table~\ref{tab:korean_specific_distribution}).
Korean-specific questions account for 8.65\% of the KMMMU set (300/3,466), but 12.12\% of the KMMMU-Adv (76/627).
This increase suggests that localized Korean content is somewhat overrepresented among harder examples.
However, because the majority of KMMMU-Adv subset questions are still not Korean-specific, localization alone does not fully explain the difficulty of the subset.

% ==================== 이 표는 업데이트 완료
\subsection{Model Performance on KMMMU-Adv}

\begin{table*}[h]
\centering
\small
\resizebox{\textwidth}{!}{%
\begin{tabular}{lcccccccccc}
\toprule
\multicolumn{1}{l}{\textbf{Model}} &
\multicolumn{1}{c}{\shortstack{\textbf{Arts} \\ \textbf{\& Design}}} &
\multicolumn{1}{c}{\shortstack{\textbf{Business} \\ \textbf{\& Public}}} &
\multicolumn{1}{c}{\shortstack{\textbf{CS} \\ \textbf{\& IT}}} &
\multicolumn{1}{c}{\textbf{Engineering}} &
\multicolumn{1}{c}{\shortstack{\textbf{General}}} &
\multicolumn{1}{c}{\shortstack{\textbf{Law} \\ \textbf{\& Ethics}}} &
\multicolumn{1}{c}{\shortstack{\textbf{Math} \\ \textbf{\& Stats}}} &
\multicolumn{1}{c}{\shortstack{\textbf{Natural} \\ \textbf{Sciences}}} &
\multicolumn{1}{c}{\shortstack{\textbf{Social} \\ \textbf{Sciences}}} &
\multicolumn{1}{c}{\shortstack{\textbf{Overall} \\ \textbf{Acc.}}} \\
\midrule
\multicolumn{11}{l}{\textbf{Open-weight Multilingual Non-Reasoning Models}} \\
\midrule
Gemma-3-4B-IT & 22.67$_{1.89}$ & 21.90$_{1.35}$ & 9.94$_{4.14}$ & 19.01$_{2.95}$ & 14.04$_{1.24}$ & 18.52$_{6.93}$ & 10.34$_{2.82}$ & 10.82$_{1.33}$ & 24.07$_{5.24}$ & 16.06$_{1.43}$ \\

Gemma-3-12B-IT & 14.67$_{3.77}$ & 8.57$_{4.04}$ & 11.70$_{4.60}$ & 15.32$_{1.84}$ & 6.14$_{3.28}$ & 22.22$_{0.00}$ & 5.75$_{4.30}$ & 8.87$_{1.70}$ & 12.04$_{7.29}$ & 12.01$_{0.72}$ \\

Gemma-3-27B-IT & 16.44$_{8.46}$ & 11.63$_{10.45}$ & 15.11$_{7.09}$ & 14.04$_{4.78}$ & 14.04$_{12.59}$ & 16.20$_{15.06}$ & 9.51$_{8.57}$ & 12.61$_{6.48}$ & 16.24$_{0.60}$ & 13.72$_{5.88}$ \\

Llama-4-Scout-17B-16E-IT & 13.33$_{3.77}$ & 9.52$_{1.35}$ & 9.36$_{2.19}$ & 17.73$_{0.53}$ & 10.53$_{2.15}$ & 20.37$_{2.62}$ & 6.90$_{2.82}$ & 11.26$_{2.72}$ & 19.44$_{4.54}$ & 13.98$_{0.53}$ \\

Llama-4-Maverick-17B-128E-IT & 17.33$_{3.77}$ & 13.33$_{1.35}$ & 20.47$_{1.65}$ & 16.74$_{1.06}$ & 10.53$_{2.15}$ & 22.22$_{4.54}$ & 17.24$_{4.88}$ & 14.29$_{1.91}$ & 21.30$_{3.46}$ & 16.37$_{1.08}$ \\

Qwen3-VL-2B-IT & 25.33$_{3.77}$ & 9.52$_{4.86}$ & 12.87$_{2.98}$ & 11.49$_{1.84}$ & 3.51$_{2.48}$ & 12.96$_{2.62}$ & 4.60$_{3.25}$ & 6.28$_{0.31}$ & 6.48$_{1.31}$ & 9.73$_{1.14}$ \\

Qwen3-VL-4B-IT & 18.67$_{1.89}$ & 10.48$_{3.56}$ & 11.70$_{0.83}$ & 12.34$_{0.35}$ & 3.51$_{1.24}$ & 3.70$_{2.62}$ & 3.45$_{2.82}$ & 4.98$_{0.61}$ & 8.33$_{2.27}$ & 9.20$_{0.08}$ \\

Qwen3-VL-8B-IT & 20.00$_{3.27}$ & 14.29$_{2.33}$ & 16.37$_{3.60}$ & 14.04$_{0.35}$ & 5.26$_{0.00}$ & 5.56$_{0.00}$ & 6.90$_{2.82}$ & 8.01$_{0.81}$ & 14.81$_{2.62}$ & 11.96$_{0.13}$ \\

Qwen3-VL-30B-A3B-IT & 21.33$_{8.22}$ & 14.29$_{4.04}$ & 6.43$_{2.19}$ & 13.62$_{0.92}$ & 4.39$_{2.48}$ & 9.26$_{6.93}$ & 8.05$_{1.63}$ & 10.17$_{0.61}$ & 16.67$_{6.00}$ & 11.70$_{1.21}$ \\

Qwen3-VL-32B-IT & 14.67$_{4.99}$ & 12.38$_{4.86}$ & 14.62$_{3.60}$ & 13.90$_{1.45}$ & 7.02$_{3.28}$ & 11.11$_{4.54}$ & 9.20$_{1.63}$ & 11.04$_{1.40}$ & 21.30$_{3.46}$ & 12.92$_{0.78}$ \\

Qwen3-VL-235B-A22B-IT & 9.33$_{4.99}$ & 12.38$_{2.69}$ & 13.45$_{1.65}$ & 17.02$_{0.92}$ & 7.89$_{4.30}$ & 7.41$_{5.24}$ & 13.79$_{2.82}$ & 14.94$_{1.06}$ & 24.07$_{2.62}$ & 15.05$_{0.87}$ \\

\midrule
\multicolumn{11}{l}{\textbf{Open-weight Korean Non-Reasoning Models}} \\
\midrule

HyperCLOVAX-SEED-Vision-3B & 9.33$_{1.89}$ & 11.43$_{0.00}$ & 16.37$_{2.19}$ & 15.46$_{0.53}$ & 4.39$_{1.24}$ & 11.11$_{0.00}$ & 8.05$_{1.63}$ & 8.66$_{0.61}$ & 14.81$_{2.62}$ & 12.23$_{0.27}$ \\

VARCO-VISION-2.0-1.7B & 17.33$_{1.89}$ & 21.90$_{4.86}$ & 19.88$_{0.83}$ & 20.43$_{1.81}$ & 7.02$_{1.24}$ & 27.78$_{12.00}$ & 12.64$_{4.30}$ & 11.47$_{0.61}$ & 16.67$_{3.93}$ & 16.96$_{1.54}$ \\

VARCO-VISION-2.0-14B & 20.00$_{3.27}$ & 23.81$_{4.86}$ & 22.81$_{7.58}$ & 24.40$_{1.12}$ & 3.51$_{1.24}$ & 24.07$_{6.93}$ & 13.79$_{4.88}$ & 15.15$_{2.14}$ & 16.67$_{0.00}$ & 19.56$_{0.72}$ \\

\midrule
\multicolumn{11}{l}{\textbf{Open-weight Multilingual Reasoning Models}} \\
\midrule

Qwen3-VL-30B-A3B-Thinking & 18.00$_{2.00}$ & 11.43$_{0.00}$ & 4.39$_{2.63}$ & 12.55$_{0.21}$ & 6.58$_{1.32}$ & 16.67$_{0.00}$ & 15.52$_{1.72}$ & 11.69$_{0.65}$ & 15.28$_{4.17}$ & 11.80$_{0.48}$ \\

Qwen3-VL-32B-Thinking & 22.67$_{1.89}$ & 6.67$_{4.86}$ & 16.37$_{1.65}$ & 12.34$_{1.04}$ & 6.14$_{1.24}$ & 16.67$_{4.54}$ & 14.94$_{4.30}$ & 11.47$_{3.84}$ & 13.89$_{6.00}$ & 12.55$_{0.53}$ \\

Qwen3-VL-235B-A22B-Thinking & 4.00$_{3.27}$ & 6.67$_{5.87}$ & 3.51$_{2.48}$ & 7.80$_{5.54}$ & 4.39$_{6.20}$ & 7.41$_{5.24}$ & 6.90$_{7.45}$ & 9.31$_{6.59}$ & 13.89$_{10.39}$ & 7.66$_{5.44}$ \\
\bottomrule
\end{tabular}
}
\caption{\footnotesize
    \textbf{Accuracy (\%) on the \textit{KMMMU-Adv} subset by disciplines in Open-Weight models.}
    This table reports results recomputed by restricting full-set evaluation outputs to the adversarially filtered \textit{Adv} subset.
    Overall accuracy is averaged across all disciplines.
    Mean accuracy is reported in percentage, with standard deviation shown as a subscript.
    The best result for each discipline and overall accuracy is shown in bold.
}
\label{app:adv_open_table}
\end{table*}

Table~\ref{app:adv_open_table} reports KMMMU-Adv results for open-weight models only.
Unlike the main KMMMU-Adv table, which reports proprietary models evaluated directly on KMMMU-Adv, this appendix table is obtained by restricting open-weight models' full-set evaluation outputs to the retained KMMMU-Adv instances.
Performance drops substantially relative to their full-set results across nearly all models, indicating that KMMMU-Adv is meaningfully more difficult for the evaluated open-weight models.

Even the strongest model remains below 20\% overall accuracy, with \texttt{VARCO-VISION-2.0-14B} achieving the highest overall score at 19.56\%.
% This result suggests that filtering step successfully removes many easier instances while preserving questions that remain challenging even for relatively strong multimodal systems.

Performance also varies considerably across disciplines.
For example, \textit{Law \& Ethics} and \textit{Arts \& Design} remain difficult for most models, while \textit{Engineering} and \textit{Natural Sciences} still show modest separation among stronger systems.
At the same time, reasoning models do not exhibit a consistent advantage over non-reasoning models on this subset.
This pattern suggests that many KMMMU-Adv subset failures arise not simply from insufficient chain-of-thought depth, but from more persistent limitations in knowledge, grounding, visual interpretation, and answer execution.
%==============================================

\section{Additional Analyses of Korean-specific Items}
\label{app:local_analysis}

\subsection{Paired Outcomes of the Cross-lingual Ablation}
\label{app:xling_ablation}

\begin{table*}[t]
\centering
\scriptsize
\setlength{\tabcolsep}{5pt}
\begin{tabular}{lrrrrrrrr}
\toprule
\textbf{Model} & \textbf{$n$} &
\textbf{Ko Acc.} & \textbf{En Acc.} &
\textbf{$\Delta$} &
\textbf{Both Correct} &
\textbf{Ko-only} &
\textbf{En-only} &
\textbf{Both Wrong} \\
\midrule
GPT-5 & 200 & 42.5 & 29.0 & +13.5 & 52 & 33 & 6 & 109 \\
Gemini-3-Flash & 200 & 58.0 & 40.0 & +18.0 & 78 & 38 & 2 & 82 \\
\bottomrule
\end{tabular}
\caption{\footnotesize
\textbf{Detailed paired outcomes for the cross-lingual ablation on Korean-specific items.}
We evaluate 200 Korean-specific paired items in Korean and English.
The English condition translates the textual question and answer options while keeping the other conditions are fixed.
Accuracies are percentages. $\Delta$ denotes Korean accuracy minus English accuracy.
Both Correct and Both Wrong indicate paired cases where the model is correct or incorrect in both language conditions.
Ko-only and En-only indicate paired discordant cases where the model answers correctly only in Korean or only in English, respectively.
Exact McNemar tests over the paired predictions show significant paired differences for both GPT-5 ($p=1.43{\times}10^{-5}$) and Gemini-3-Flash ($p=1.49{\times}10^{-9}$).
}
\label{tab:xling_ablation_appendix}
\end{table*}

Table~\ref{tab:xling_ablation_appendix} provides a detailed paired-outcome breakdown of the controlled cross-lingual ablation discussed in Section~\ref{sec:xling_ablation}.
All 200 items in this analysis are Korean-specific, after removing non-Korean-specific questions from the cross-lingual subset.
The English condition translates only the textual question and answer options, while preserving the original visual input, answer, and task structure.

Across both models, accuracy is lower in the English condition than in the original Korean condition.
GPT-5 decreases from 42.5\% to 29.0\%, and Gemini-3-Flash decreases from 58.0\% to 40.0\%.
The paired outcomes show the same direction.
For GPT-5, 33 items are answered correctly only in Korean, compared to 6 items answered correctly only in English.
For Gemini-3-Flash, 38 items are answered correctly only in Korean, compared to only 2 items answered correctly only in English.
Exact McNemar tests over the paired predictions indicate that these differences are statistically significant for both models.

These results suggest that translating Korean-specific questions into English does not remove the main source of difficulty.
Instead, translation may disrupt localized legal, administrative, technical, or institutional terminology, as well as the alignment between the translated question and Korean visual evidence.
% Since the visual input is kept unchanged, Korean text embedded inside images may remain present in the English condition.
% Therefore, this experiment should be interpreted as a question-language ablation under fixed visual evidence, rather than as a fully English-localized benchmark.

\subsection{Discipline-level Korean-specific Gaps}
\label{app:local_ci}
Table~\ref{tab:local_ci} decomposes the Korean-specific gap of Qwen3-VL-235B-A22B-IT by
discipline. The gap is Korean-specific accuracy minus non-Korean-specific accuracy, and
negative values indicate lower accuracy on Korean-specific items. The discipline-standardized
gap is the average of within-discipline gaps weighted by the discipline distribution of the
full benchmark. We estimate 95\% confidence intervals with 100,000 bootstrap resamples over
the 3,466 items with three complete trials. Both Korean-specific and non-Korean-specific
items are presented in Korean under the same zero-shot protocol with the same English
instruction prompt, so this gap is computed independently of the cross-lingual ablation in
Appendix~\ref{app:xling_ablation}.

The deficit is concentrated in Business \& Public and Law \& Ethics. The confidence intervals
for the other sufficiently represented disciplines except Natural Sciences and CS \& IT
include zero, so the aggregate gap is not uniform across disciplines. The Arts \& Design
estimate is descriptive because it contains only one Korean-specific item, and we report no
confidence interval for it. The Math \& Stats estimate should also be read with caution
because it contains only six Korean-specific items.

\begin{table}[h]
\centering
\small
\setlength{\tabcolsep}{3pt}
\resizebox{\columnwidth}{!}{%
\begin{tabular}{lrrrrrc}
\toprule
Discipline & $n_{\text{KS}}$ & $n_{\text{Non}}$ & Acc.$_{\text{KS}}$ & Acc.$_{\text{Non}}$ & Gap & 95\% CI \\
\midrule
Business \& Public          & 76  & 227   & 23.2 & 69.5 & $-46.2$ & $[-55.4, -36.6]$ \\
Law \& Ethics               & 82  & 22    & 27.6 & 65.2 & $-37.5$ & $[-46.7, -28.3]$ \\
Arts \& Design$^{\dagger}$  & 1   & 78    & 0.0  & 25.6 & $-25.6$ & -- \\
Natural Sciences            & 31  & 792   & 17.2 & 42.4 & $-25.1$ & $[-34.6, -14.1]$ \\
CS \& IT                    & 16  & 308   & 14.6 & 38.6 & $-24.1$ & $[-38.4, -6.6]$ \\
General                     & 28  & 95    & 21.4 & 25.6 & $-4.2$  & $[-17.9, 10.5]$ \\
Math \& Stats$^{\dagger}$   & 6   & 221   & 50.0 & 53.1 & $-3.1$  & $[-17.0, 10.8]$ \\
Engineering                 & 44  & 1,185 & 31.1 & 31.4 & $-0.3$  & $[-12.2, 11.9]$ \\
Social Sciences             & 16  & 238   & 66.7 & 52.9 & $+13.7$ & $[-7.0, 32.8]$ \\
\midrule
Raw gap                     & 300 & 3,166 & --   & --   & $-13.51$ & $[-17.94, -9.02]$ \\
Discipline-standardized gap & 300 & 3,166 & --   & --   & $-13.43$ & $[-19.27, -7.44]$ \\
\bottomrule
\end{tabular}}
\caption{\textbf{Discipline-level Korean-specific gaps of Qwen3-VL-235B-A22B-IT.} KS and Non
denote Korean-specific and non-Korean-specific items. Accuracy and gaps are in percentage
points. $^{\dagger}$ denotes disciplines with fewer than ten Korean-specific items.}
\label{tab:local_ci}
\end{table}

%==============================================

\begin{figure*}[h]
    \centering
    \includegraphics[width=0.9\textwidth]{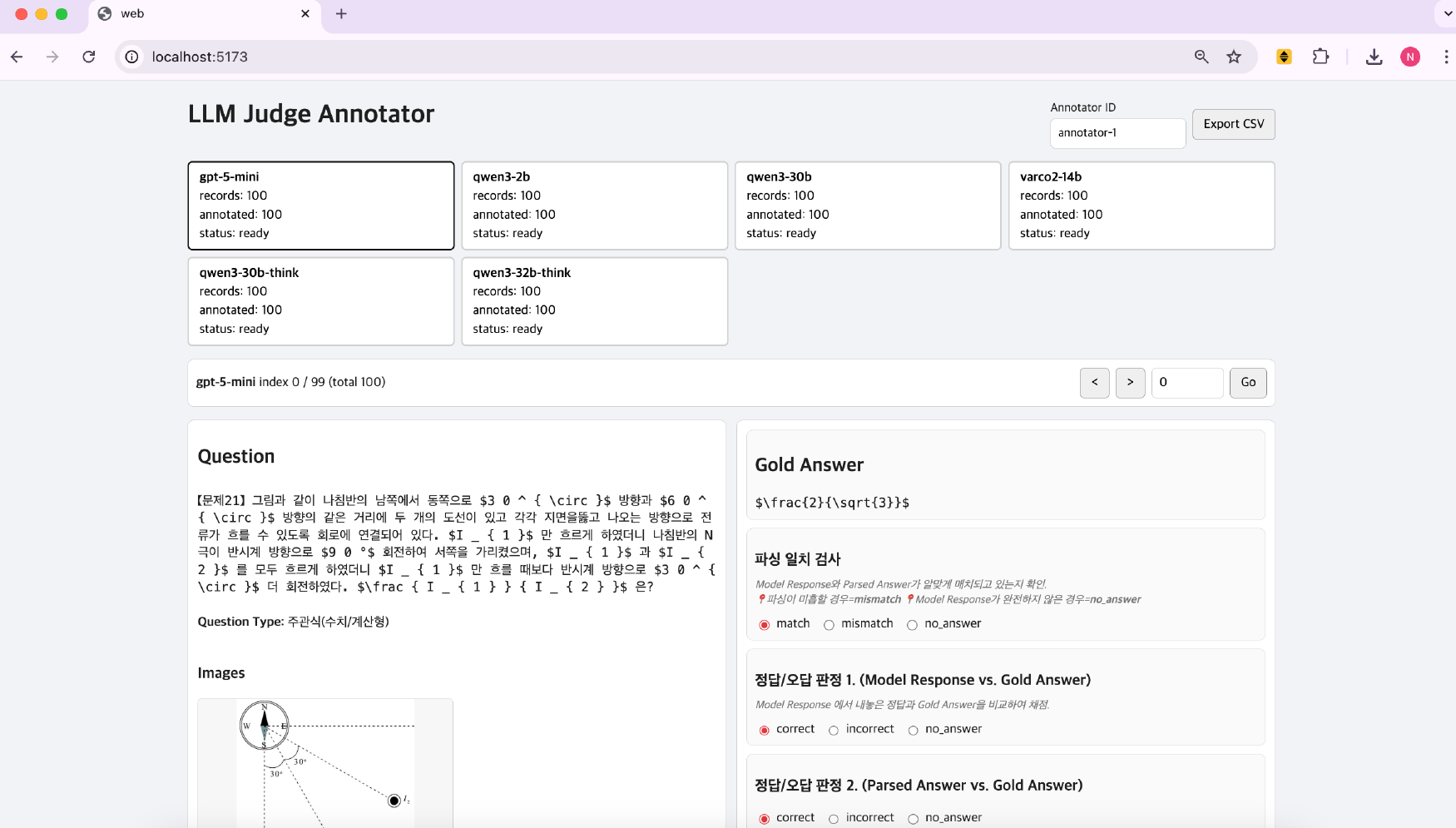}
    \caption{
    \textbf{Annotation interface for manual validation of LLM scorer outputs.}
    For each sample, annotators review the question, image, gold answer, model response, and parsed answer, and record parsing consistency, correctness judgments, metadata consistency, and optional comments.
    }
    \label{fig:llm-as-judge-annotator}
\end{figure*}

% \begin{table*}[t]
% \centering
% \scriptsize
% \resizebox{\textwidth}{!}{%
% \begin{tabular}{lcccc}
% \toprule
% Model 
% & Parsed acc. (\textit{no answer}) 
% & Response acc. (\textit{no answer})
% & Parsed acc. (answered) 
% & Response acc. (answered) \\
% \midrule
% Qwen3-VL-2B-IT          & 0.96 & 0.96 & 1.000 & 1.000 \\
% Qwen3-VL-30B-A3B-IT         & 0.94 & 0.72 & 0.962 & 0.975 \\
% VARCO-VISION-2.0-14B           & 1.00 & 1.00 & 0.946 & 0.926 \\
% GPT-5-Mini           & 1.00 & 1.00 & 0.936 & 0.891 \\
% Qwen3-VL-32B-Thinking   & 1.00 & 0.20 & 0.889 & 0.734 \\
% Qwen3-VL-30B-A3B-Thinking   & 0.95 & 0.15 & 0.729 & 0.347 \\
% \midrule
% Average              & 0.98 & 0.67 & 0.910 & 0.796 \\
% \bottomrule
% \end{tabular}
% }
% \caption{
% \textbf{Judge--human alignment broken down by response completeness.}
% Parsed-answer-based judging remains more robust on incomplete responses and generally aligns better with human labels on answered cases as well.
% For the \textit{no\_answer} subset, we report accuracy because the labels are highly imbalanced.
% }
% \label{tab:judge_alignment_breakdown}
% \end{table*}

\begin{table*}[t]
\centering
\scriptsize
\setlength{\tabcolsep}{3.0pt}
\resizebox{\textwidth}{!}{%
\begin{tabular}{lcccccccc}
\toprule
\multirow{2}{*}{\textbf{Model}} 
& \multirow{2}{*}{\textbf{Human agr.}}
& \multirow{2}{*}{\textbf{No-answer rate}}
& \multicolumn{2}{c}{\textbf{Overall judge--human agr.}}
& \multicolumn{2}{c}{\textbf{No-answer acc.}}
& \multicolumn{2}{c}{\textbf{Answered acc.}} \\
\cmidrule(lr){4-5}
\cmidrule(lr){6-7}
\cmidrule(lr){8-9}
& & & \textbf{Parsed} & \textbf{Response}
& \textbf{Parsed} & \textbf{Response}
& \textbf{Parsed} & \textbf{Response} \\
\midrule
Qwen3-2B-IT          & 0.99 & 0.51 & 0.980 & 0.980 & 0.96 & 0.96 & 1.000 & 1.000 \\
Qwen3-30B-IT         & 0.97 & 0.18 & 0.959 & 0.928 & 0.94 & 0.72 & 0.962 & 0.975 \\
Qwen3-32B-Thinking   & 0.93 & 0.31 & 0.925 & 0.495 & 1.00 & 0.20 & 0.889 & 0.734 \\
Qwen3-30B-Thinking   & 0.91 & 0.22 & 0.880 & 0.326 & 0.95 & 0.15 & 0.729 & 0.347 \\
VARCO2-14B           & 0.97 & 0.04 & 0.948 & 0.929 & 1.00 & 1.00 & 0.946 & 0.926 \\
GPT-5-Mini           & 0.95 & 0.01 & 0.937 & 0.892 & 1.00 & 1.00 & 0.936 & 0.891 \\
\midrule
Average              & 0.95 & 0.21 & 0.921 & 0.742 & 0.98 & 0.67 & 0.910 & 0.796 \\
\bottomrule
\end{tabular}
}
\caption{\footnotesize
\textbf{Human agreement and judge--human alignment across six model runs.}
``Human agr.'' denotes pairwise human agreement.
``Parsed'' and ``Response'' report judge--human alignment under parsed-answer-based and full-response-based evaluation, respectively.
The no-answer rate denotes the proportion of outputs without a complete final answer.
The no-answer and answered columns further break down judge--human accuracy by response completeness.
}
\label{tab:judge_alignment}
\end{table*}

\section{Reliability of LLM-based Answer Scoring}
\subsection{Annotation Protocol}

To validate the reliability of our evaluation pipeline, we conducted a manual annotation study using a custom annotation interface (Figure~\ref{fig:llm-as-judge-annotator}).
Three annotators independently reviewed each sample with access to the question, associated image, gold answer, model response, and parsed answer.

For each sample, annotators first evaluated whether the parsed answer faithfully reflected the answer expressed in the original model response, labeling it as \textit{match}, \textit{mismatch}, or \textit{no\_answer}.
Here, \textit{mismatch} indicates that the parser failed to preserve the intended answer, while \textit{no\_answer} indicates that the model response itself did not contain a complete answer.

Annotators then assessed correctness with respect to the gold answer in two ways: once based on the full model response and once based on the parsed answer, each labeled as \textit{correct}, \textit{incorrect}, or \textit{no\_answer}.
This design allowed us to distinguish parsing failures from genuine model errors.
In addition, annotators verified the consistency of the recorded question type and image type using \textit{match}, \textit{mismatch}, or \textit{unsure}, and could provide free-form comments for ambiguous cases.

\subsection{Human Alignment Results}
\label{app:judge_alignment}

The alignment study contains 600 examples drawn from six model runs, with 100 outputs sampled from each run and balanced across question formats.
Table~\ref{tab:judge_alignment} reports pairwise human agreement and judge--human alignment under both parsed-answer-based and full-response-based evaluation.
Human agreement is consistently high, with pairwise agreement ranging from 0.91 to 0.99, indicating that the annotation task is generally well defined.
Agreement is slightly lower for some reasoning-heavy outputs, where responses are longer and often contain partially correct intermediate reasoning without a clearly finalized answer.

Overall, parsed-answer-based judging aligns more closely with human labels than full-response judging.
Averaged across the six runs, parsed-answer-based judging achieves 0.921 overall agreement, compared with 0.742 agreement for full-response judging.
The gap is especially large for reasoning models, where full responses can contain ambiguous or incomplete reasoning traces even when the parsed final answer is easier to evaluate consistently.

Table~\ref{tab:judge_alignment} also breaks down judge--human accuracy by response completeness.
On \textit{no\_answer} cases, parsed-answer-based judging reaches 0.98 accuracy on average, whereas full-response judging drops to 0.67.
On answered cases, parsed-answer-based judging also remains stronger, achieving 0.910 accuracy compared with 0.796 for full-response judging.
These results indicate that the benefit of parsed-answer-based evaluation is not limited to incomplete outputs; it also improves alignment when the model produces a final answer.

Manual inspection of disagreement cases suggests that many residual mismatches arise from answer-formatting and completion issues.
In multiple-choice questions, some models produce the content of the correct option rather than its explicit index, which can cause an otherwise correct response to be judged as incorrect.
More broadly, disagreement is concentrated in cases where the response contains extended or partially correct reasoning but fails to end with a clearly finalized answer.
Taken together, these results support our use of parsed-answer-based judging as the primary evaluation protocol, especially for long or reasoning-heavy model outputs.

\section{Error Analysis Details}
\label{app:error_analysis_details}

\subsection{Error Inspection Methodology}
To investigate the mechanisms underlying the patterns in Tables~\ref{tab:main-result-overall}--\ref{tab:korean-specific-gap-full}, we conducted targeted manual error inspection over three focused subsets.

First, for paired reasoning comparison, we examined reversal cases between \texttt{Qwen3-VL-32B-IT} and \texttt{Qwen3-VL-32B-Thinking}.
To analyze domain-specific reasoning effects, we sampled 25 items each from \textit{Math \& Stats}, \textit{Engineering}, and \textit{Natural Sciences} among questions answered correctly by \texttt{Qwen3-VL-32B-IT} but incorrectly by \texttt{Qwen3-VL-32B-Thinking}, yielding 75 inspected reversals in total.

Second, for Korean-specific failures, we analyzed incorrect outputs from a 300-item comparison set between \texttt{Qwen3-VL-235B-A22B-IT} and \texttt{HyperCLOVAX-SEED-Vision-3B}.
We randomly sampled 25 incorrect cases from each model for qualitative inspection, focusing on recurring patterns of localized knowledge failure, regulatory category mismatch, and terminology grounding errors.

Third, to characterize persistent disciplinary bottlenecks, we additionally inspected representative failure cases from \textit{Arts \& Design} and \textit{General}, focusing on \texttt{Qwen3-VL-235B-A22B-IT} with reference to corresponding \texttt{Qwen3-VL-235B-A22B-Thinking} outputs where relevant.

Each inspected case was reviewed by two authors, who examined the image, question, model output, and ground-truth answer.
Disagreements were resolved through discussion.

\subsection{Additional Qualitative Examples for Post-perceptual Reasoning Effects}
\label{app:it_vs_thinking}

\begin{figure}
    \centering
    \includegraphics[width=1.0\linewidth]{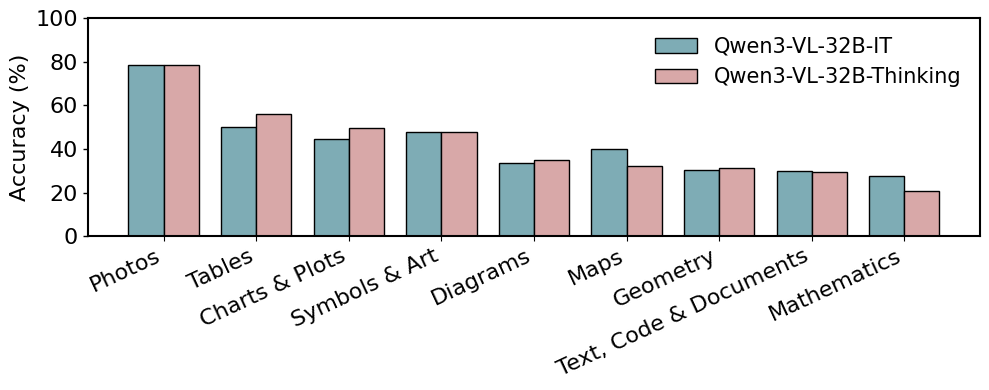}
    \caption{\textbf{Accuracy by visual modality for Qwen3-VL-32B-IT and Qwen3-VL-32B-Thinking.}
Performance remains broadly similar across visual modality categories, suggesting that explicit reasoning does not systematically change raw visual evidence extraction.
The main differences appear to arise after evidence extraction, such as in task framing, constraint tracking, and answer finalization.}
    \label{fig:qwen_visual_acc_compare}
\end{figure}

Figure~\ref{fig:qwen_visual_acc_compare} reports performance by visual modality for
\texttt{Qwen3-VL-32B-IT} and \texttt{Qwen3-VL-32B-Thinking}.
Across most modality categories, the two variants remain broadly similar, with no consistent pattern indicating that explicit reasoning systematically improves raw visual evidence extraction.

\begin{figure*}
    \centering
    \includegraphics[width=0.9\linewidth]{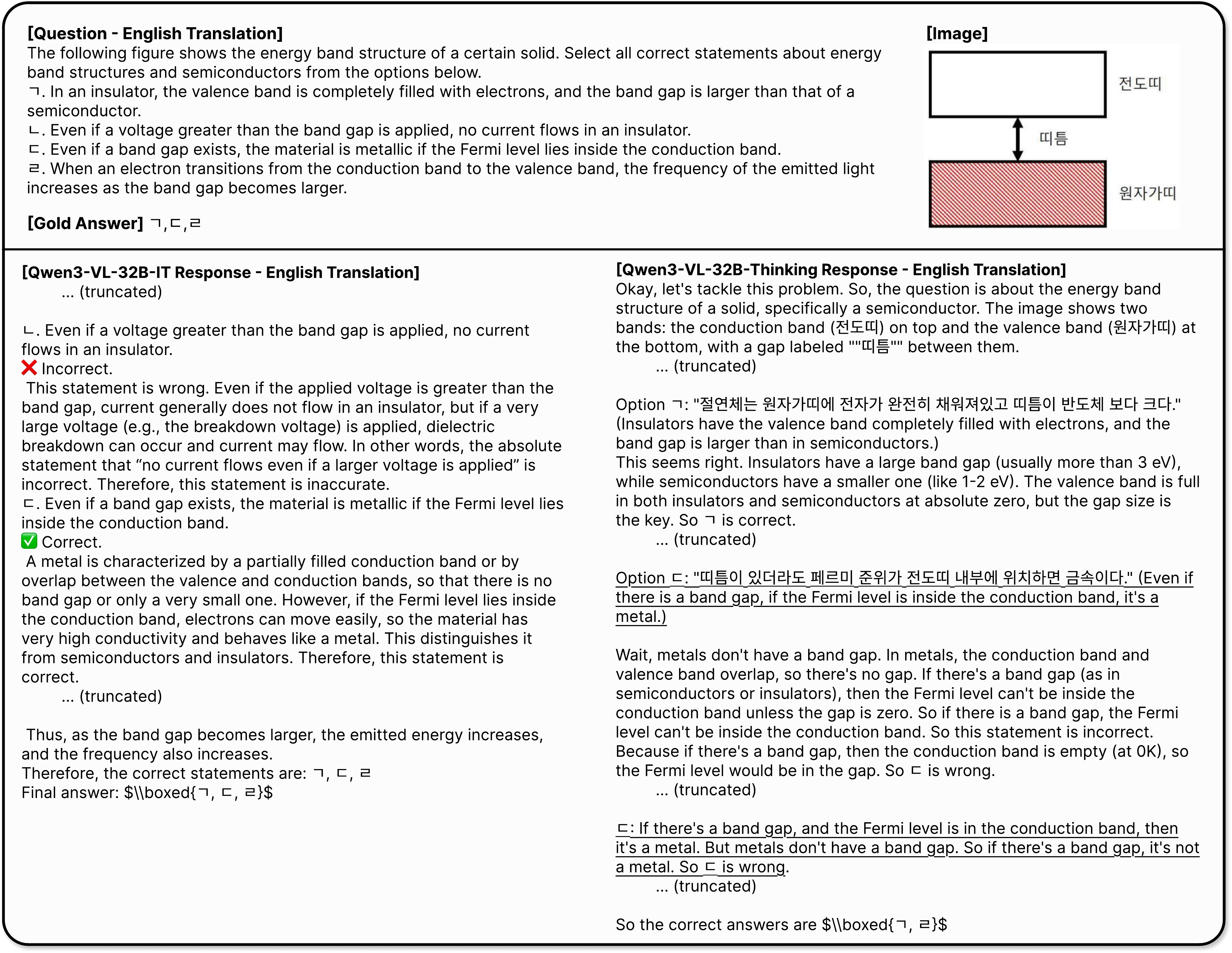}
\caption{\textbf{Rigid conceptual framing in a \textit{Natural Sciences} reversal.}
\texttt{Qwen3-VL-32B-IT} correctly applies the relevant conductivity criterion, whereas \texttt{Qwen3-VL-32B-Thinking} overcommits to an overly rigid band-gap-based schema and rejects the crucial statement about the Fermi level in the conduction band.}
    \label{fig:qwen_it_think_example_1}
\end{figure*}

To complement the aggregate modality comparison, we include a representative reversal case from \textit{Natural Sciences} where \texttt{Qwen3-VL-32B-IT} answers correctly but \texttt{Qwen3-VL-32B-Thinking} fails in Figure~\ref{fig:qwen_it_think_example_1}.
This example illustrates the broader pattern discussed in the main text: the difference does not arise from one variant clearly seeing the image while the other does not, but from how the extracted evidence is framed and translated into a final judgment.

\begin{figure*}
    \centering
    \includegraphics[width=0.9\linewidth]{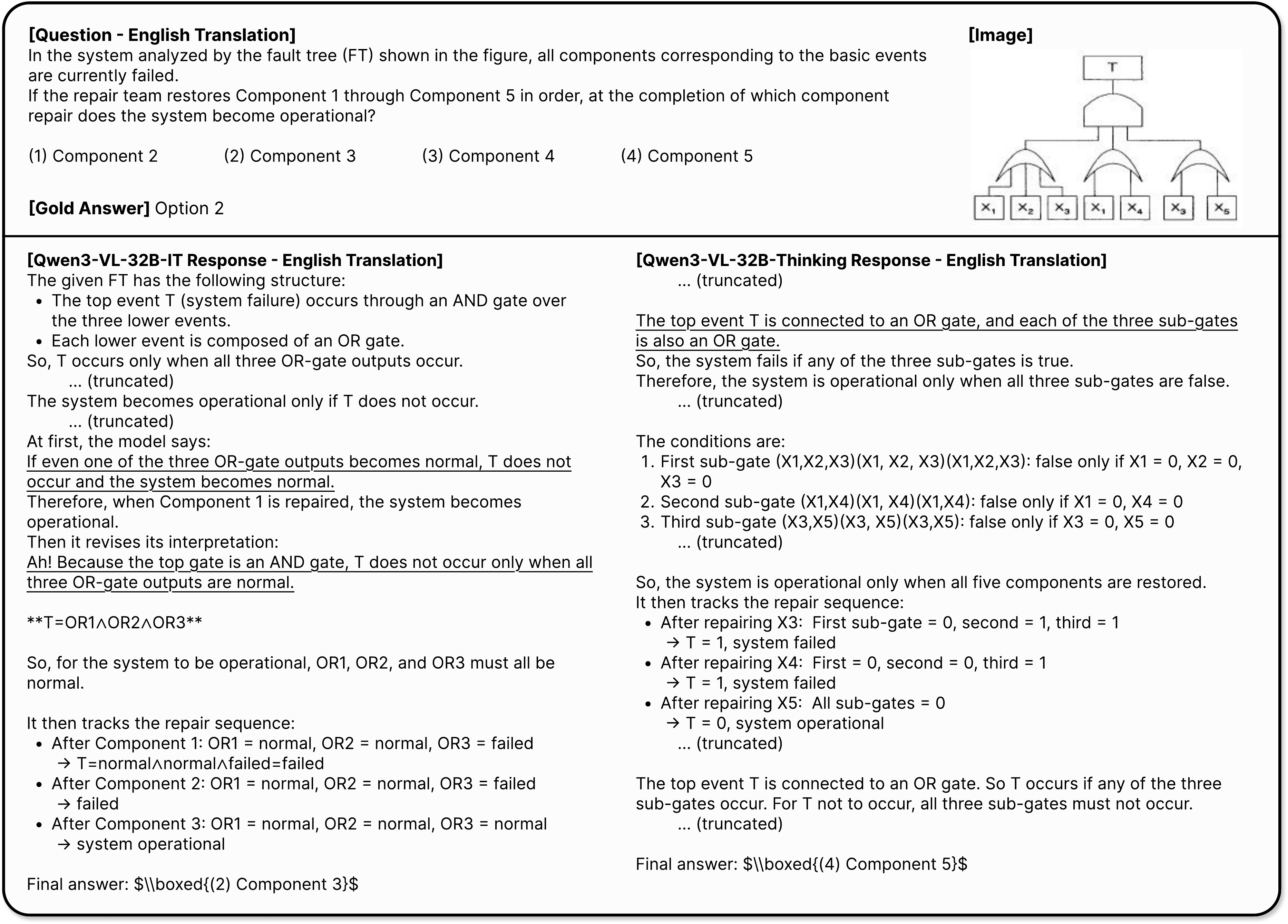}
    \caption{\textbf{Structural misinterpretation in an \textit{Engineering} reversal (Fault Tree analysis).}
\texttt{Qwen3-VL-32B-IT} partially corrects an early gate-level misinterpretation, whereas \texttt{Qwen3-VL-32B-Thinking} persists with an incorrect top-level gate reading and derives the wrong recovery point.}
    \label{fig:qwen_it_think_example_2}
\end{figure*}

\begin{figure*}
    \centering
    \includegraphics[width=0.9\linewidth]{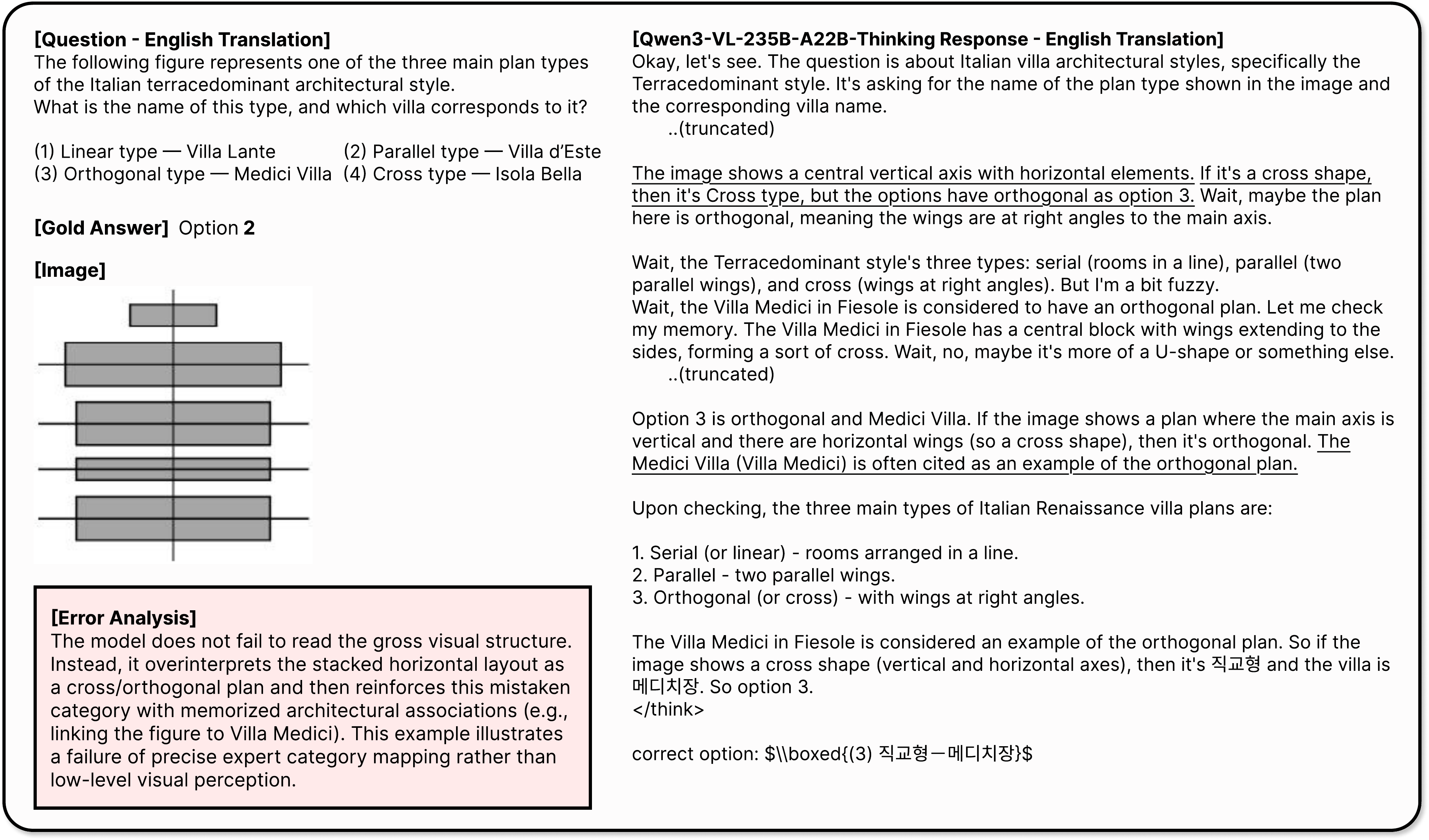}
    \caption{\textbf{Exact architectural category misclassification in \textit{Arts \& Design}.}
    The model recognizes the overall structure of the plan, but fails to map it to the correct standardized architectural category. Instead, it overcommits to an orthogonal-plan interpretation and supports it with a plausible but incorrect villa association.}
    \label{fig:art_example_1}
\end{figure*}

\begin{figure*}
    \centering
    \includegraphics[width=0.9\linewidth]{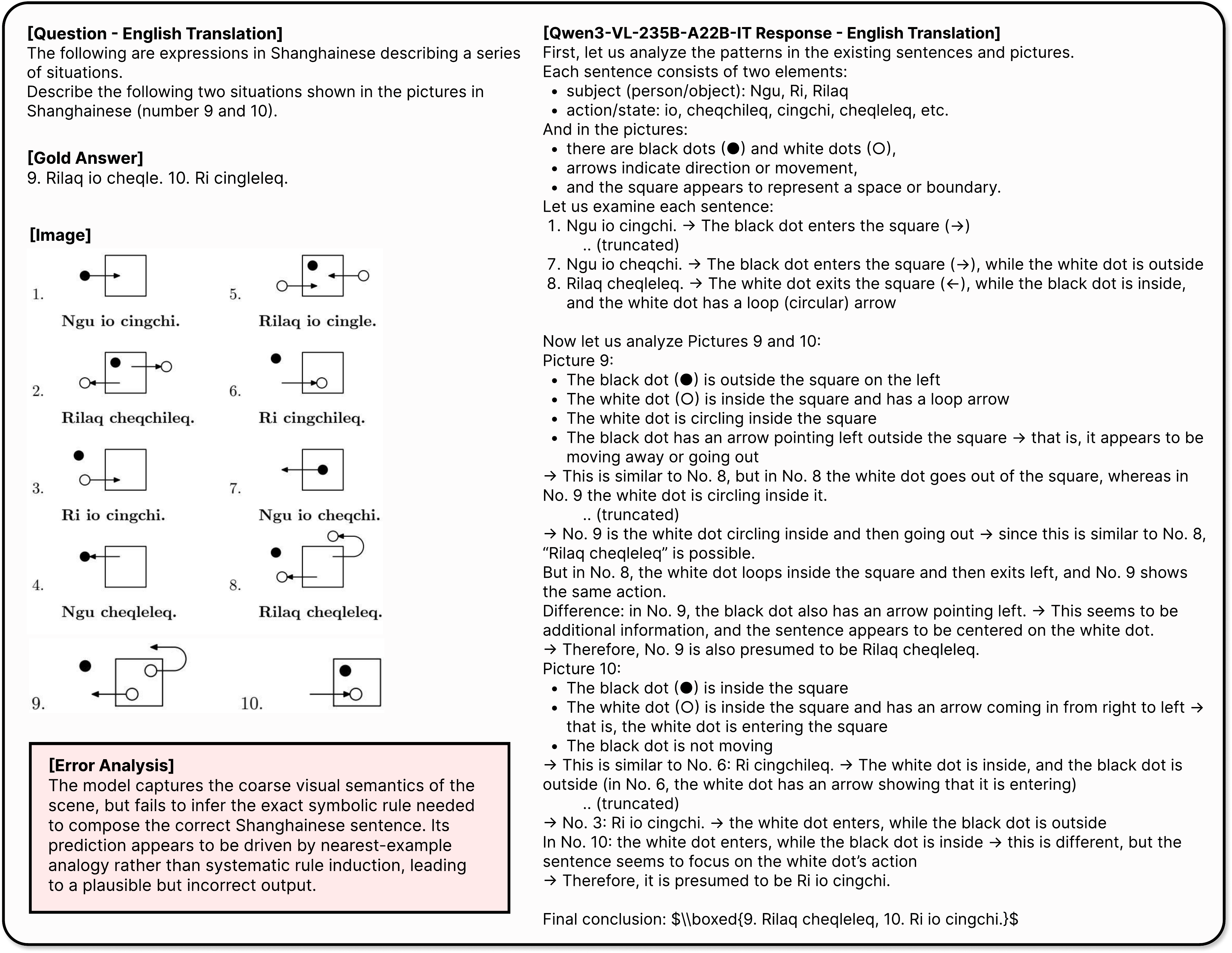}
    \caption{\textbf{Few-shot symbolic induction failure in \textit{General}.}
    A Shanghainese-language item requiring the model to infer a latent mapping from a small set of diagram--expression pairs and apply it to new cases.
    Although \texttt{Qwen3-VL-235B-A22B-IT} produces a detailed step-by-step analysis, it fails to recover the full correspondence system and instead relies on partial surface analogies, leading to a plausible but incorrect answer.}
    \label{fig:general_example_1}
\end{figure*}

\begin{figure*}
    \centering
    \includegraphics[width=0.9\linewidth]{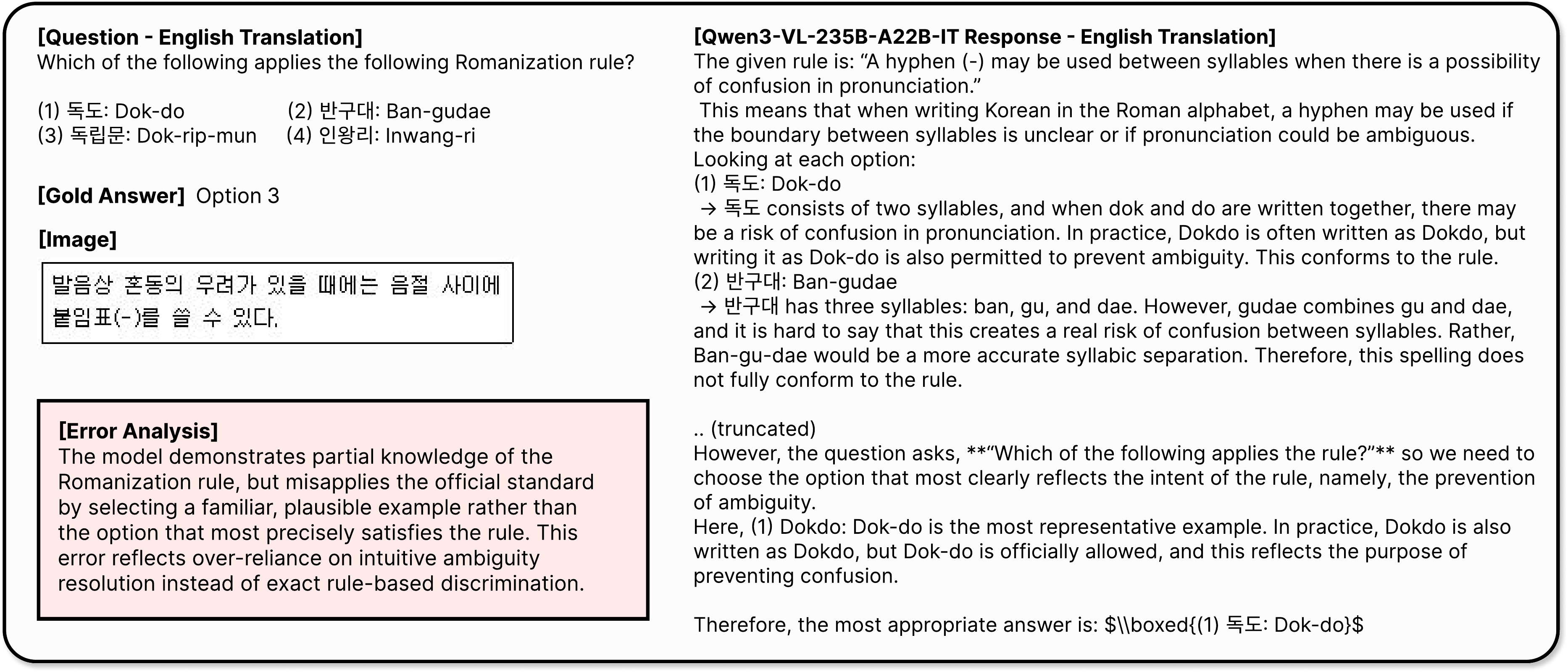}
    \caption{\textbf{Exact standards and rule-criterion misapplication in \textit{General}.}
    An item testing the official Romanization rule for hyphen use.
    The model gives a broadly reasonable explanation of the rule, but selects the wrong option because it applies an approximate plausibility-based criterion rather than the exact condition required by the formal standard.}
    \label{fig:general_example_2}
\end{figure*}

In this case, the \textit{Thinking} variant does not fail because it misses the basic visual structure or the relevant physical relation.
Instead, it becomes anchored on an overly rigid conceptual rule and evaluates the option through that internal schema rather than the condition stated in the question itself.
This qualitatively matches the pattern in our reversal inspection for \textit{Natural Sciences}, where errors often stem from premature commitment to an incorrect problem frame rather than from missing visual evidence.

Figure~\ref{fig:qwen_it_think_example_2} shows a representative \textit{Engineering} reversal in which both variants initially misread the same FT diagram, but differ in whether they recover from that early structural error.
The underlined spans mark the decision points that anchor each model's subsequent reasoning.
For \texttt{Qwen3-VL-32B-IT}, the underlined text shows an initial misinterpretation followed by an explicit self-correction once the model recognizes that the top gate is an AND gate, allowing it to recover the correct answer.
By contrast, \texttt{Qwen3-VL-32B-Thinking} remains committed to the mistaken assumption that the top event is connected to an OR gate.
That early structural error then propagates through the entire derivation, leading the model to produce a fully consistent but fundamentally incorrect recovery analysis.
This example matches the broader pattern we observed in \textit{Engineering}: failures often arise from incorrect diagram-level structure interpretation, after which the model elaborates a coherent solution under the wrong logical frame.

\subsection{Additional Qualitative Examples for Disciplinary Bottlenecks}
\label{app:disciplinary_examples}

For disciplinary bottlenecks, we use \texttt{Qwen3-VL-235B-A22B-IT} as a consistent reference point for representative qualitative examples.
In additional inspected cases, we also examined corresponding outputs from \texttt{Qwen3-VL-235B-A22B-Thinking}, and observed qualitatively similar failure patterns.
The examples in this appendix illustrate recurring errors in exact convention-to-label mapping in \textit{Arts \& Design} and few-shot symbolic induction or terminology grounding in \textit{General}.

\paragraph{Example 1: Expert category mismatch in \textit{Arts \& Design}.}
Figure~\ref{fig:art_example_1} shows a representative \textit{Arts \& Design} failure in which the model captures the coarse spatial organization of an architectural plan but fails to assign the correct standardized category.
Rather than identifying the plan type required by the question, the model overinterprets the stacked horizontal layout as evidence for an orthogonal or cross-shaped structure, and then reinforces this mistaken frame with a plausible but incorrect villa association.
This is not a low-level perception failure: the model recognizes salient geometric structure, but fails at precise convention-to-label mapping among closely related expert categories.
The case therefore illustrates a recurring bottleneck in \textit{Arts \& Design}, where errors arise not from missing the visual content altogether, but from overconfident misclassification of specialized visual conventions into the wrong technical label.

\paragraph{Example 2: Few-shot symbolic induction failure in \textit{General}.}
As shown in Figure~\ref{fig:general_example_1}, this item requires the model to infer a latent mapping between diagram configurations and Shanghainese expressions from a small set of paired examples, and then apply that rule to unseen cases.
The model produces a long, locally plausible analysis, but fails to recover the full underlying correspondence system.
Instead, it partially matches surface patterns and then drifts into self-invented regularities, yielding answers that are structurally plausible but incorrect.
This example illustrates a recurring \textit{General} bottleneck in KMMMU: some items require few-shot rule induction from sparse symbolic evidence, not just fluent explanation or broad world knowledge.

\paragraph{Example 3: Exact standards and rule-criterion misapplication in \textit{General}.}
Figure~\ref{fig:general_example_2} shows a case that depends on precise application of an official Romanization rule, rather than broad linguistic plausibility alone.
The model produces a reasonable explanation and discusses related principles, but selects the wrong option because it applies an approximate criterion instead of the exact standard required by the question.
This pattern appears in multiple \textit{General} items that test official terminology, orthographic conventions, or certification-style definitions: the model often gives a broadly sensible account, but misses the precise condition that determines correctness.

\section{Ablation Study}
\label{app:ablation}

\subsection{Evaluation of Image Dependency}

\begin{table}[h]
\scriptsize
\centering
\setlength{\tabcolsep}{8pt}
\renewcommand{\arraystretch}{1.1}
\begin{tabular}{lcc}
\toprule
\textbf{Model} & \textbf{Original} & \textbf{Text-only} \\
\midrule
Gemini-3-Flash & \pmstd{45.15}{0.98} & \pmstd{20.43}{0.48} \\
GPT-5-Mini     & \pmstd{21.32}{0.33} & \pmstd{9.75}{0.85} \\
\bottomrule
\end{tabular}
\caption{\textbf{Ablation results for image dependency.} We compare the average accuracy and standard deviation ($\pmstd{Acc}{std}$) of models on the original multimodal dataset versus the text-only baseline.}
\label{tab:image-dependency}
\end{table}

To verify that KMMMU functions as a genuinely multimodal benchmark, we measure how strongly performance depends on access to visual information.
We conduct a text-only ablation in which models receive the textual question and answer options, but the associated image is removed.

Table~\ref{tab:image-dependency} shows substantial performance drops for both \texttt{Gemini-3-Flash} and \texttt{GPT-5-Mini} under this setting.
For \texttt{Gemini-3-Flash}, accuracy declines from 45.15\% to 20.43\%, a drop of 24.72 percentage points.
For \texttt{GPT-5-Mini}, accuracy declines from 21.32\% to 9.75\%, a drop of 11.57 percentage points.
These results indicate that many KMMMU questions cannot be solved reliably from text alone.

We also manually inspected the 60 cases answered correctly by \texttt{GPT-5-Mini} in the text-only setting.
A non-trivial subset of visually accompanied items remains solvable without direct image access, but these cases often do not reflect genuine visual understanding.
Instead, they typically fall into several recurring patterns:
(i) the textual prompt already specifies most of the decisive constraints, making the image largely auxiliary;
(ii) the answer can be inferred from strong domain priors or option elimination rather than visual grounding;
(iii) the option structure, numerical form, or canonical diagram schema enables answer reconstruction without actual image reading; and
(iv) in some quantitative science items, the core reasoning is already determined by symbolic conditions in the text, with the image serving mainly as contextual support.

Overall, these findings support the multimodal validity of KMMMU while also clarifying that visual accompaniment and strict image-essentiality are not identical.
Although removing images causes large performance drops, some items remain text-solvable because they contain sufficient textual, structural, or prior-driven cues to permit correct answering without direct image use.

\begin{table*}[t]
\centering
\scriptsize
\resizebox{\textwidth}{!}{
\begin{tabular}{llcccccc}
\toprule
\textbf{Source} & \textbf{Model} & \textbf{Exactness$_{NH}$} & \textbf{Exactness$_H$} & \textbf{Refusal$_{NH}$} & \textbf{Refusal$_H$} & \textbf{Hallucination$_{NH}$} & \textbf{Hallucination$_H$} \\
\midrule
PSAT & Gemini-3-Flash & 14.20 & 13.14 & 17.53 & 24.70 & 76.10 & 68.13 \\
     & Gemini-3-Pro   & 0.34  & 0.46  & 95.60 & 94.78 & 3.20  & 3.21 \\
     & GPT-5-Mini     & 0.00  & 0.00  & 99.60 & 100.00 & 0.00  & 0.00 \\
\midrule
NCS  & Gemini-3-Flash & 10.05 & 6.54  & 29.08 & 50.20 & 50.20 & 35.57 \\
     & Gemini-3-Pro   & 0.18  & 0.04  & 99.21 & 100.00 & 0.40  & 0.00 \\
     & GPT-5-Mini     & 0.65  & 0.45  & 95.65 & 100.00 & 1.19  & 1.19 \\
\midrule
NTQ 
     & Gemini-3-Flash & 15.70 & 18.58 & 19.66 & 15.61 & 61.54 & 76.84 \\
     & Gemini-3-Pro   & 0.33  & 2.03  & 93.86 & 79.15 & 4.44  & 16.92 \\
     & GPT-5-Mini     & 0.45  & 0.25  & 97.96 & 99.83 & 0.51  & 0.17 \\
\midrule
Olympiads
     & Gemini-3-Flash & 10.48 & 10.91 & 23.12 & 37.56 & 55.58 & 49.48 \\
     & Gemini-3-Pro   & 0.13  & 0.38  & 97.92 & 97.41 & 1.56  & 2.33 \\
     & GPT-5-Mini     & 0.52  & 0.18  & 94.56 & 98.96 & 2.85  & 1.04 \\
\bottomrule
\end{tabular}
}
\caption{\textbf{Prefix-completion analysis for potential data contamination.}
Models are given the first 35\% of each question together with the associated image, and asked to generate the remaining continuation.
\textit{Exactness} denotes a judge-assigned 0--100 faithfulness rating with respect to the reference continuation.
\textit{Refusal} and \textit{Hallucination} denote judge-labeled failure modes, reported as percentages, under the No-hint ($NH$) and Hinted ($H$) settings.}
\label{tab:contamination_results}
\end{table*}

\subsection{Data Contamination Analysis}

To probe possible memorization, we run a prefix-completion test in which models receive the first 35\% of a question together with its image and are asked to generate the remaining continuation.
We restrict this analysis to questions longer than 150 tokens, since shorter exam-style items often begin with generic instructions that provide too little question-specific content for a meaningful reconstruction test.

We evaluate three frontier models under two settings.
In the first, no additional metadata is provided.
In the second, we provide the exam name and year as a potential memorization trigger.

We evaluate generated continuations using \texttt{Gemini-2.5-Flash} as a judge.
For each continuation, the judge assigns (i) an exactness rating from 0 to 100 based on overlap with the reference continuation and preservation of key details, and (ii) categorical labels indicating refusal or hallucination.
Thus, exactness reflects judge-rated reconstruction fidelity, whereas refusal and hallucination characterize distinct failure behaviors rather than the same measurement scale.

As shown in Table~\ref{tab:contamination_results}, \texttt{GPT-5-Mini} and \texttt{Gemini-3-Pro} exhibit very high refusal rates across all four sources, typically accompanied by near-zero exactness ratings.
This suggests that these models usually do not attempt faithful continuation under the prefix-completion setup.

\texttt{Gemini-3-Flash} attempts continuation more often, yielding lower refusal rates and somewhat higher exactness ratings than the other two models.
However, its exactness remains low overall, and the hinted setting does not produce a consistent increase across sources.
Moreover, hallucination rates remain high, indicating that many attempted continuations are low-fidelity generations rather than faithful reconstructions.

Taken together, these results do not provide strong evidence that benchmark performance is driven by simple memorization of question continuations.
If contamination were a major driver under this setup, we would expect more consistently faithful reconstruction and clearer improvement when exam metadata is provided as a hint.
Instead, the dominant pattern is either refusal or low-fidelity continuation.

\end{document}